\documentclass[epsfig,a4paper,12pt,titlepage]{book}
\usepackage{epsfig}
\usepackage{fancyhdr}
\usepackage{plain}
\usepackage{mathtools}
\usepackage{amsmath}
\usepackage{multirow}
\usepackage{multicol}
\usepackage{url}
\usepackage[table,xcdraw]{xcolor}
\usepackage{listings}
\usepackage{algorithm}
\usepackage{algpseudocode}
\usepackage{comment}
\usepackage{caption}
\usepackage[multiple]{footmisc}
\usepackage{graphicx}
\usepackage{color}
\usepackage[T1]{fontenc}
\usepackage[utf8]{inputenc}

\algnewcommand{\LineComment}[1]{\State  \(\triangleright\) #1}

\makeatletter
\renewcommand\part{%
  \if@openright
    \cleardoublepage
  \else
    \clearpage
  \fi
  \thispagestyle{empty}%
  \if@twocolumn
    \onecolumn
    \@tempswatrue
  \else
    \@tempswafalse
  \fi
  \null\vfil
  \secdef\@part\@spart}
\makeatother

\newcommand{\clearemptydoublepage}{\newpage{\pagestyle{empty}\cleardoublepage}}

\makeindex
\begin{document}
\pagestyle{plain}

\newpage
\clearemptydoublepage
\thispagestyle{empty}
\begin{center}

\begin{figure}[h!]
  \centerline{\psfig{file=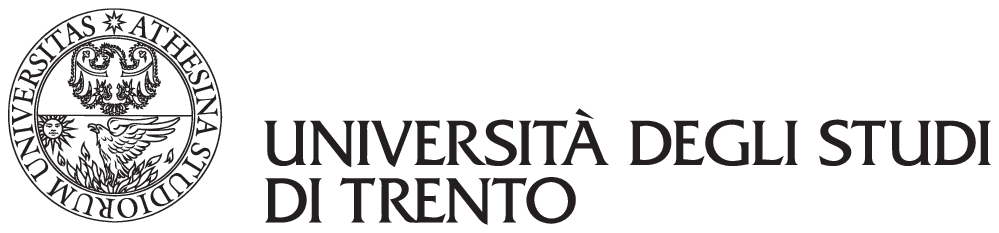,width=0.4\textwidth}}
\end{figure}

\hrulefill

DEPARTMENT OF INFORMATION ENGINEERING AND COMPUTER SCIENCE\\
\textbf{ICT International Doctoral School}\\

\vspace{1 cm} 
\Huge\textsc{Automatic Post-Editing for Machine Translation\\} 

\vspace{0.3 cm}

\begin{center}
\begin{tabular}{l}
\huge{Rajen Chatterjee}\\
\end{tabular}
\end{center}
\vspace{1 cm} 
\begin{flushleft}
\begin{tabular}{ll}
\multicolumn{2}{l}{\large Advisor}\\
 & \large Dr. Marco Turchi\\
 & \large Fondazione Bruno Kessler\\
\end{tabular}
\end{flushleft}

\begin{flushleft}
\begin{tabular}{ll}
\multicolumn{2}{l}{\large Co-Advisor}\\
 & \large Matteo Negri\\
 & \large Fondazione Bruno Kessler\\
\end{tabular}
\end{flushleft}

\hrulefill

\normalsize
October $2019$
\end{center}

\newpage
\clearemptydoublepage
\thispagestyle{empty}
\large
{\bf \Huge Abstract}
\vspace{0.2cm}
\noindent

Automatic Post-Editing (APE) aims to correct systematic errors in a machine translated text.
This is primarily useful when the machine translation (MT) system is not accessible for improvement, leaving APE as a viable option to improve translation quality as a downstream task - which is the focus of this thesis.
This field has received less attention compared to MT due to several reasons, which include: the limited availability of data to perform a sound research, contrasting views reported by different researchers about the effectiveness of APE, and limited attention from the industry to use APE in current production pipelines.

In this thesis, we perform a thorough investigation of APE as a downstream task in order to: \textit{i)} understand its potential to improve translation quality; \textit{ii)} advance the 
core technology - starting from classical methods to recent deep-learning based solutions; \textit{iii)} cope with limited and sparse data; \textit{iv)} better leverage multiple input sources; \textit{v)} mitigate the task-specific problem of over-correction; \textit{vi)} enhance neural decoding to leverage external knowledge; and \textit{vii)} establish an online learning framework to handle data diversity in real-time.

All the above contributions are discussed across several chapters, and most of them are evaluated in the APE shared task organized each year at the Conference on Machine Translation.
Our efforts in improving the technology resulted in the best system at the 2017 APE shared task, and our work on online learning received a distinguished paper award at the Italian Conference on Computational Linguistics.
Overall, outcomes and findings of our work have boost interest among researchers and attracted industries to examine this technology to solve real-word problems.

\vspace{0.5cm}
\noindent

{\bf Keywords}
\noindent
Automatic Post-Editing, Machine Translation, Deep Learning, Natural Language Processing, Neural Network

\clearemptydoublepage
{\bf \Huge Acknowledgement}
\vspace{0.5cm}
\noindent

First and foremost, I want to thank my advisor Dr. Marco Turchi. 
It has been an honor to be his PhD student.
I truly appreciate his guidance and moral support throughout my PhD pursuit.
It has always been fun to discuss technical topics and brainstorm research ideas.
He was always there to support my victory and share my failure.
His guidance, suggestions, and feedback helped me to be a better person in my professional and personal life.
Without his support this journey of PhD would have been incomplete.
I also want to thank my co-advisor Matteo Negri to be my inspiration of creativity.
His guidance helped me a lot to improve my skill in writing research papers, making presentation slides, and speaking in front of large audience about my work.
He has always been the source of motivation to achieve my goals.

I would like to thank all the members of the machine translation group at Fondazione Bruno Kessler for their valuable time to discuss research during meetings, presentations, and lunch break.
Thanks to all the fellow Interns, Masters, and PhD students for making my journey joyful.
A special thanks to all the people involved in providing infrastructure support and resolving any issue at their earliest.
I also want to thank all the staff of the University of Trento and the Welcome Office to help me with all the bureaucratic work.

Lastly, I want to thank my parents and all family members for their love, encouragement, and sacrifices.
A special thanks to my sister Moushmi Banerjee for guiding me through each step of my life and helping me to pursue the right career for which I was able to begin this exiting journey of PhD.
Most of all thanks to my loving, caring, and encouraging wife Kaustri Bhattacharyya whose support and patient during the final stages of my Ph.D. is much appreciated. 

\pagenumbering{roman}
\tableofcontents
\clearemptydoublepage
\listoftables
\clearemptydoublepage
\listoffigures
\clearemptydoublepage 
\pagestyle{fancy}
\renewcommand{\chaptermark}[1]{}
\pagenumbering{arabic}

\chapter{Introduction}
\label{chap:intro}
Translation that involves many languages is a growing requirement in today's world.
The need exists in various sectors, including tourism, education, the legal system, the media, and health care.
In a multilingual continent like Europe, this demand poses a challenge to language service providers (LSPs), who must deliver quality translation at speed.
To cope with the increasing demand, LSPs have shifted human translation from a completely manual process to a semi-automated one, using computer-assisted translation (CAT) tools to generate translation suggestions.
CAT is becoming a widespread and standard tool for LSPs, who daily face a trade-off between quality versus productivity.
Machines, however, are not perfect.
Machine translation (MT) cannot provide translations that match the quality of work done by humans.
High quality is paramount for translating critical data, such as medical transcriptions, legal documents, and patent files.
Therefore, human supervision is needed to verify machine-translated (MT) text and correct any errors.
This manual post-processing step, which transforms a ``raw'' MT text into a ``publishable'' text, is referred to as \textbf{post-editing} (PE).

The task of editing MT text is characterized by repetitiveness.
MT systems, irrespective of the underlying technology, usually provide the same translation for the same input.
Hence, if the MT system generates an erroneous translation, the same error is generated repeatedly, which the editor must then correct repeatedly.
To reduce the manual overhead of repeated corrections, in the last decade a branch of MT research has focused on automating PE.
This field is known as \textbf{automatic post-editing} (APE).
The aim is to detect and correct systematic errors in an MT text before it is seen by humans.

\section{Motivation}
As mentioned above, MT systems are not perfect. 
They are prone to making systematic errors that a human editor must fix before publication.
This process of translation is also known as ``translation as post-editing''
A by-product of the process is that it generates parallel data, consisting of the MT output and its corrected version.
These data can be leveraged to develop APE systems that can identify recurring MT errors and correct them.
The corrections range from fixing typographic errors to adapting the terminology  for a specific domain, or even modeling the personal style of an individual translator.
These capacities are crucial when the MT system used to produce the translation is a ``black-box'', whose inner workings are inaccessible and thus cannot be fine-tuned or re-trained.
The ``black-box'' condition frequently occurs for small LSPs who rely on third-party translation engines, such as Google or Bing to obtain the initial translations which are later corrected by the PE.
Regardless of the MT access rights, downstream processing of MT output always benefits from external resources.
Such resources include a target language parser to improve the syntax or a quality estimator to identify the errors in the MT output.

In general, APE serves various purposes, such as customizing translation to a domain, modeling the personal style of an individual translator, or meeting customer-specific needs.
These multi-purpose APE applications are of growing interest to industries.
Large e-commerce companies like eBay use APE tools to improve their title generation in different languages for the products displayed on their e-commerce websites \cite{mathur-ueffing-leusch:2017:INLG2017}.
They have obtained substantial improvement in translation quality by using APE rather than direct translation through an MT engine. 
The social networking entity Facebook is experimenting with the QuickEdit system to improve translations; text is automatically post-edited based on human feedback \cite{grangier-EtAl:2017:arxiv}.
This system could be helpful for improving the translation of user contents on social network.
One of the largest translation service providers, Systran, uses an APE engine in its production unit to improve the translations by its rule-based MT system \cite{DBLP:journals/corr/CregoKKRYSABCDE16}.
These real-world use cases make APE appealing and warrant further investigation.
The next section describes several problems in APE that are addressed in this thesis.

\section{Problems}
\label{chap:intro-problems}
This section provides an overview of several challenges and problems in APE.
The problems are grouped according to three aspects: data, technology, and application.

\subsection{Data}
From a data perspective, there have been limited efforts to create APE datasets for research purposes. 
APE data mainly consist of source text and MT output, which is used to learn the error patterns, as well as human PEs, which are used to learn the correction rules.
Although abundant data are generated by translation companies in their daily workflow, the information is usually confidential and not publicly available for research purposes.
Public datasets are limited and often noisy, covering few languages and containing only a few tens-thousand training samples.
The limited availability of training data often results in data sparsity.
Especially when different versions of MT systems are used to generate the translation, error patterns become more complex, which increases the challenge of identifying them.
The complexity further increases when corrections are provided by several post-editors, who may correct the same error in different ways.
The provision of different solutions for the same error increases the ambiguity of correction rules and their application can result in incoherent output.
The training data provide only a fraction of all possible error-correction examples, and not all of the learned error-correction rules are universally applicable.
Applying rules in the wrong context can damage the MT output instead of improving it.
This situation is likely to occur in a real-time CAT environment, where the APE tools should be robust enough to cope with the continuous stream of diverse data from different domains or genres.
In this scenario, the system should have the capacity to evolve over time through continuous learning from the post-editor's feedback.
To overcome the challenges of data scarcity, new APE datasets are needed to conduct sound and reliable research.
Also, novel system architectures are needed to inherently address the issue of learning from limited, sparse, and diverse data.

\subsection{Technology}
From a technological perspective, most research in APE to date has been based on classical MT approaches.
Researchers have approached the problem as a monolingual translation task to transform the raw MT output into a more correct version using classical phrase-based MT technology.
Recent advances in neural network, especially deep learning, have achieved state-of-the-art performance in many machine learning (ML) tasks.
The power of neural nets, which are capable of learning complex non-linear functions, can be leveraged along with the classical methods which are relatively efficient at learning from small datasets.
These two paradigms complement each other to provide better performance.
The strong learning ability of neural nets means that many ML tasks are transitioning completely from classical methods to end-to-end deep-learning-based solutions.
Integrating and evaluating the latest neural technology in APE involves several challenges.
It is known that deep-learning-based solutions require vast training data, which is a serious bottleneck for APE (as discussed above).
Training neural networks with limited and sparse data often leads to the problem of overfitting; as a result, the system may not generalize well using unseen datasets.
The challenge escalates if the network must learn the relation between words in the source and the MT sentence to generate a better translation.
Joint modelling of both inputs - source and MT - can help, but this approach increases the data sparsity.

APE methods follow a similar trend to MT approaches, namely generating a new sequence of target words given an input.
However, re-translating an existing MT output poses the risk of replacing a correct MT word with a synonym.
In such cases, the final output, although an alternate correct translation, is penalized by an evaluation metric that compares the translation with a human PE version.
Re-translation might also modify a correct MT word with an incorrect one, which will eventually increase the work of post-editors.
In table \ref{tab:over-correction}, an example is provided from our experiments to illustrate this phenomenon of over-correction by APE.
The source sentence was \textit{``All values, even primitive values, are objects.''}.
It was first translated in German by an MT system to \textit{``alle Werte, auch Grundwerte, handelt es sich um Objekte.''} which is almost the same as the reference \textit{``Bei allen Werten, auch Grundwerten, handelt es sich um Objekte.''}.
When re-translated by an APE system, the result was \textit{``Alle Werte, auch Grundwerte, sind Objekte.''} which introduced more errors in the final output.
\begin{table*}[h]
\begin{tabular}{p{15.5cm}}
\hline  \hline
\textbf{Src:} All values, even primitive values, are objects.\\
\textbf{MT:} alle Werte, auch Grundwerte, handelt es sich um Objekte.\\
\textbf{APE:} Alle Werte, auch Grundwerte, sind Objekte.\\
\textbf{Ref:} Bei allen Werten, auch Grundwerten, handelt es sich um Objekte.\\
\hline \hline
\end{tabular}
\caption{Examples illustrating the problem of over-correction in APE task.}
\label{tab:over-correction}
\end{table*}
We address the problem of \textbf{over-correction} in several chapters, either implicitly by focusing on system design or explicitly by leveraging external resources.
In short, a thorough investigation of recent technology for the APE task is needed.
New mechanisms must be developed to deal with data scarcity and to learn efficiently from multiple input (source and MT) while mitigating the problem of over-correction.

\subsection{Application}
From an application perspective, MT output post-processing using an APE tool can be helpful for several purposes.
These include customizing a translation for a domain, modelling the personal style of a translator, and adding customer-specific constraints to the translation.
A typical application for LSPs includes translating documents that must comply with specific terminology and style guidelines.
In such situations it is generally necessary to consider external resources, such as terminology dictionaries, to constrain the system to generate pre-defined translations.
Typically, APE operates at downstream level by re-translating the entire MT output; therefore, it might modify parts of the translation that are already correct - that is, it might ``over-correct''.
This would lead to an overall deterioration in the translation quality.
To address this problem, external resources can be leveraged that provide quality judgments for the MT words so that APE fixes only erroneous parts of the MT segment.
Efficient APE techniques should be investigated to assist in integrating various external resources within the APE system to leverage prior knowledge, and customize the system to specific target applications.

\section{Outline}
This thesis captures the revolution in technology for APE, starting with classical techniques and gradually shifting toward neural networks.
The final scenario is state-of-the-art end-to-end methods based on deep learning.
This revolution is a theme running through several chapters as the study addresses the problems mentioned in Section \ref{chap:intro-problems}.

Chapter 2 begins with a brief introduction to the translation market and the need for translation services.
The chapter discusses several challenges associated with manual translation.
It explains the overall translation workflow within a CAT framework, in which machines are used to assist humans by providing translation suggestions.
Different MT technologies are discussed broadly, covering rule-based, phrase-based, and neural-based paradigms - and their pros and cons.
The MT evaluation process is then described through a brief explanation of the human and automated metrics that are used to measure a system's performance.
Finally, the chapter provides an overview of various satellite tasks, such as APE and quality estimation (QE), that can be used together to improve the translation quality through downstream processing of the MT output.

Chapter 3 introduces the APE task, which is the main topic of this thesis.
The chapter briefly describes different working scenarios for using an APE system, such as a ``glass-box'', in which the MT system is accessible, and the ``black-box'', in which the MT system is not accessible.
The ``black-box'' is the focus of this thesis.
Several challenges in APE are discussed regarding the data, technology, and application.
The chapter also provides a literature review with a focus on various aspects, such as: \textit{i)} performance variations when APE is applied to correct the output of  \textit{rule-based versus statistical MT}; \textit{ii)} the use of APE  for \textit{error correction versus domain adaptation};  \textit{iii)} the difference between training in \textit{general domains versus domain-specific  data}; and \textit{iv)} performance variations when learning from  \textit{reference translations versus human post-edits}.
The chapter concludes with an overview of the evolution of APE during the first three rounds of the APE shared task that was organized at the conference on machine translation (WMT).

Chapter 4 summarizes our initial contribution when the state-of-the-art systems were based on classical APE technology.
The chapter provides a first systematic analysis of the two variants of the classical approach, namely the monolingual and context-aware approaches.
Their potential effect on translation quality is explained.
The analysis was performed across six language pairs to study the relationship between the performance gain from the APE system versus the original MT output quality.
The chapter also presents our proposed method to combine the two variants to leverage their possible complementarity.
In general, these classical methods often learn noisy and unreliable PE rules because of the weak statistical evidence that results from limited and sparse training data.
This problem is addressed in the final section of the chapter, through discussing task-specific dense features that help to filter out unreliable rules.

Chapter 5 examines the shift in technology from pure classical APE techniques to a hybrid approach that leverages the power of neural nets.
First, the chapter presents a novel approach to combining the above two classical APE variants in an elegant factored MT framework.
Second, we propose a mechanism to integrate a neural network with classical methods using this framework.
This enables us to explore the complementarity of neural nets and classical models.
This chapter analyzes the result of using a classical n-gram model or neural language model, such as part-of-speech tags and class-based models, within an APE system.
An ablation study was performed to understand the contribution of each model alone and in combination.
The chapter concludes with answers to questions about which of the two technologies is superior, and what combination of models is most effective for improving the final translation quality.

Chapter 6 investigates recent technologies based on deep neural networks for APE.
The chapter describes our contribution to an existing neural MT architecture and its training procedure to address task-specific challenges.
The challenges include \textit{i)} learning from limited, sparse in-domain data; \textit{ii)} modelling multiple inputs to leverage both source and MT; and \textit{iii)} avoiding over-correction to preserve the quality of MT output.
This enhanced architecture became the core component of our submission to the APE shared task at WMT 2017, where it outperformed all other submissions.
The overall system architecture is described in detail.
We describe the ensemble mechanism to leverage several neural models and an additional post-processing step to re-rank n-best APE hypotheses.

Chapter 7 extends our end-to-end deep-learning-based APE solution to address the problem of constraint decoding, in which prior knowledge is used to guide the decoding process. Specifically, we leveraged quality judgments of MT text as prior knowledge to mitigate the problem of over-correction. We then performed a first systematic analysis of different strategies to integrate quality estimation and APE to improve the quality of translation. The joint contribution of the two technologies was analyzed in various settings. A light integration was pursued when QE was used either to trigger the automatic correction of MT text or to validate an automatic correction by comparing it with the original MT output. A tighter integration was pursued when QE was used to inform the automatic correction process by identifying problematic passages in the MT text. Depending on the applied strategy, QE predictions were produced at the word or sentence level.

Chapter 8 focuses on an online learning framework rather than the batch setup that was discussed in previous chapters.
Usually, APE evaluation in a batch setup is performed in a controlled setting, where the representativeness of the training set for the test data is crucial for good performance.
Real-life scenarios, however, do not guarantee favorable learning conditions. 
Ideally, to be integrated in a real professional translation workflow, APE tools should be flexible enough to cope with continuous streams of diverse data from different domains or genres.
To manage these challenges, an online APE framework is proposed.
The framework is robust to data diversity - that is, able to learn and apply correction rules in the right contexts.
It also evolves over time by continuously extending and refining its knowledge.

Chapter 9 concludes the thesis by highlighting the lessons learned during the study period.
The chapter summarizes our overall contributions to the field of APE.
In addition, we discus unresolved problems and their possible solutions for future investigation.

\section{Contribution}
A part of the work discussed in this thesis has been presented and published at several conferences, workshops, and shared tasks during the PhD period.
Here follows a list of selected papers along with their short summary and highlight of my individual contribution:

\begin{itemize}
    \item \textbf{Rajen Chatterjee}, Marion Weller, Matteo Negri, Marco Turchi. Exploring the Planet of the APEs: a Comparative Study of State-of-the-art Methods for MT Automatic Post-Editing. In \textit{Proceedings of the $53^{rd}$ Annual Meeting of the Association for Computational Linguistics and the $7^{th}$ International Joint Conference on Natural Language Processing}, 2015, vol. 2, pp 156-161.\cite{P15-2026}\\
    \textbf{Summary:} This paper provides a first systematic study of the two variants of the classical APE approaches over multiple language pairs to understand their potential and to leverage their complementarity.\\
    \textbf{Contribution:} I created the constrained APE dataset across six language pairs from a publicly available corpora, where all the dataset share the same source sentences thus making several analysis comparable across languages.
    I ran all the monolingual and context-aware experiments, evaluated them, and performed further analysis.
    I also proposed an approach to combine both the classical APE variants to leverage their complementarity, which resulted in a much better performance over each individual variants.
    \item \textbf{Rajen Chatterjee}, Marco Turchi and Matteo Negri. The FBK Participation in the WMT15 Automatic Post-editing Shared Task. In \textit{Proceedings of the Tenth Workshop on Statistical Machine Translation}, 2015, pp 210-215.\cite{chatterjee-turchi-negri:2015:WMT}\\
    \textbf{Summary:} This paper addresses the problem of learning reliable post-editing rules from a more challenging dataset belonging to a broader domain and containing post-edits that are obtained via crowdsourcing, which often leads to a noisy data set.\\
    \textbf{Contribution:} I created the joint representation based dataset to train a context-aware APE system in addition to a monolingual APE system.
    I conducted all the experiments for language model selection as well as to evaluate different pruning strategies to filter out noisy rules.
    The dense features for capturing similarity, reliability, and usefulness aspects were designed by the co-authors and I performed the experiments to analyze the best combination of these features.
    \item \textbf{Rajen Chatterjee}, Gebremedhen Gebremelak, Matteo Negri, Marco Turchi. Online Automatic Post-Editing across Domains. In \textit{Proceedings of Third Italian Conference on Computational Linguistics {\&} Fifth Evaluation Campaign of Natural Language Processing and Speech Tools for Italian}, 2016, vol. 1749, paper 16.\cite{DBLP:conf/clic-it/ChatterjeeGNT16}\\
    \textbf{Summary:} This paper provides a first study on online learning for the APE task by proposing a novel framework that is robust enough to cope with the continuous stream of diverse data coming from different domains/genres.\\
    \textbf{Contribution:} I developed the online learning framework with the core contribution of instance selection mechanism. 
    I created the mixed domain dataset and conducted all the experiments which include running our system as well as other existing online learning systems on our dataset.
    I evaluated all the systems and performed further analysis.
    \item \textbf{Rajen Chatterjee},  Mihael Arcan, Matteo Negri and Marco Turchi. Instance Selection for Online Automatic Post-Editing in a multi-domain scenario. In \textit{Proceedings of the Twelfth Conference of the Association for Machine Translation in the Americas}, 2016, vol 1, pp 1-15.\cite{AMTA-2016-1-15}\\
    \textbf{Summary:} This paper extends the previous online learning framework \cite{DBLP:conf/clic-it/ChatterjeeGNT16} for evaluating its potential in a cross-domain translation scenario, where the system is trained on one domain and tested on another domain.\\
    \textbf{Contribution:} I created the dataset and conducted all the experiments to evaluate the potential of our online learning framework on in-domain as well as cross domain data.
    I performed all the evaluation and analysis of batch, cdec, Thot and our framework.
    \item \textbf{Rajen Chatterjee}, Jose G. C. de Souza, Matteo Negri, Marco Turchi. The FBK Participation in the WMT 2016 Automatic Post-editing Shared Task. In \textit{Proceedings of the First Conference on Machine Translation}, 2016, pp 745-750.\cite{chatterjee-EtAl:2016:WMT}\\
    \textbf{Summary:} This paper presents a mechanism to combine the two technologies classical and deep learning, where the former can learn more efficiently from a smaller dataset and the latter can generalize the learned skills better.\\
    \textbf{Contribution:} I developed the core idea of integrating the two technologies (phrase-based and neural) in a factored machine translation framework.
    I trained all the models and conducted experiments to study the effect of several language models on the final system's performance.
    I also introduced the data augmentation technique to mitigate the problem of over-correction, which showed significant improvement.
    The quality estimation scores were provided by the co-authors and I used them to obtain the best threshold that showed maximum improvement.
    \item \textbf{Rajen Chatterjee}, Gebremedhen Gebremelak, Matteo Negri and Marco Turchi. Online Automatic Post-editing for MT in a Multi-Domain Translation Environment. In \textit{Proceedings of the $15^{th}$ Conference of the European Chapter of the Association for Computational Linguistics}, 2017, vol. 1, 2017, pp 525-535.\cite{E17-1050}\\
    \textbf{Summary:} This paper enhances the previous online learning framework \cite{DBLP:conf/clic-it/ChatterjeeGNT16} with a dynamic knowledge base to compute model's parameters on-the-fly along with a negative feedback mechanism to penalize the incorrect post-editing rules.\\
    \textbf{Contribution:} I developed the core framework in which the dynamic knowledge base was added by the co-authors. 
    I contributed partially in implementing  the negative feedback features, running experiments, and analyzing time and performance of different approaches.
    \item \textbf{Rajen Chatterjee}, Matteo Negri, Marco Turchi, Marcello Federico, Lucia Specia and Frédéric Blain. Guiding Neural Machine Translation Decoding with External Knowledge. In \textit{Proceedings of the Second Conference on Machine Translation}, 2017, pp 157-168.\cite{chatterjee-EtAl:2017:WMT1}\\
    \textbf{Summary:} This paper addresses the problem of constraint decoding in neural MT. Constraints are provided in the form of; \textit{i)} terminology translation: for translating in-domain terms in the MT task, and \textit{ii)} quality judgements: to preserve the correct machine-translated words in the APE task.\\
    \textbf{Contribution:} I developed the novel idea of integrating external knowledge in neural MT decoding using a look-ahead mechanism. 
    I implemented the entire decoding algorithm in deep learning, conducted all the experiments for both MT and APE task, and performed error analysis.
    \item \textbf{Rajen Chatterjee}, M. Amin Farajian, Matteo Negri, Marco Turchi, Ankit Srivastava and Santanu Pal. Multi-source Neural Automatic Post-Editing: FBK’s participation in the WMT 2017 APE shared task. In \textit{Proceedings of the Second Conference on Machine Translation}, 2017, vol. 2, pp 630-638.\cite{chatterjee-EtAl:2017:WMT2}\\
    \textbf{Summary:} This paper presents a deep learning approach for joint modelling both source and MT output. It also addresses the problem of over-correction by introducing a task-specific loss function.\\
    \textbf{Contribution:} I developed the core idea of creating a multi-source neural APE system for jointly learning from multiple inputs.
    I also proposed a task-specific loss function to tackle the problem of over-correction which showed positive impact in terms of precision.
    I extended an existing neural MT framework to support multi-source processing and also added a new loss function.
    I trained all single-source and multi-source models which were combined by a weighted ensemble technique.
    I also proposed a novel task-specific re-ranker using shallow features that further improved system's performance.
    \item \textbf{Rajen Chatterjee}, Matteo Negri, Marco Turchi, Frederic Blain, Lucia Specia. Combining Quality Estimation and Automatic Post-editing to Enhance Machine Translation Output. In \textit{Proceedings of the Thirteenth Conference of the Association for Machine Translation in the Americas}, 2018, vol 1, pp 26-38.\cite{AMTA-2018}\\
    \textbf{Summary:} This paper presents a first systematic study of combining quality estimation and automatic post-editing together for improving the quality of machine-translated text.\\
    \textbf{Contribution:} I designed the core idea of integrating QE knowledge in APE and developed the decoder for this purpose.
    I annotated the datasets to obtain oracle QE tags and also trained APE models.
    I performed all the experiments and contributed in further analysis.
\end{itemize}

\chapter{MT Background}
\label{chap:MT}
This chapter introduces the topic of machine translation (MT).
MT, with post-editing by humans, has become the \textit{de-facto} standard in the translation industry.
We show its application in a computer-assisted translation (CAT) framework adopted by language service providers (LSPs). 
This application can increase the productivity of translators by generating automatic recommendations.
Furthermore, we discuss the popular technologies running behind MT engines, such as phrase-based and neural-based, which also form the backbone of the automatic PE engines described in the following chapters.
An MT evaluation framework is then described, with a focus on automatic metrics and human evaluation setups to measure the performance of translation systems, which will be used in our automatic post-editing task.
Finally, we discuss various application areas of MT, with PE being the main focus in the remaining chapters.

\section{Translation Market}
Advancements in science and technology have stimulated the need for translation services, and have provided new solutions to fulfill this need.
The invention of telephone now calls for real-time speech translation to facilitate multilingual communication.
Similarly, the existence of the internet has led to the need to use different languages when browsing websites or viewing social networking sites.
Across the spectrum, from code breaking during the Second World War\footnote{Recall the use of machines to crack the German Enigma code way back in World War II} to business globalization, translation has been a key factor in communication challenges.
Considering that human society across the globe comprises an estimated 7 billion people, across the 193 countries, with 150 official languages, there is a vast demand for translation services.
A report by TAUS\footnote{\url{https://www.taus.net/think-tank/reports}} predicted that the market revenue for MT will reach about \$638 million by 2020, up from \$250 million in 2014, with an average growth rate of 16.9\% per year.
To cope with this increasing demand, LSPs have shifted the human translation process from being completely manual to instead being semi-automated with the help of MT engines.
Translators work with CAT tools such as MateCat\footnote{\url{https://www.matecat.com/}} or SDL Trados\footnote{\url{https://www.sdl.com/software-and-services/translation-software/sdl-trados-studio/}} to automatize and speed up the translation process.
Automation is achieved by showing suggestions to translators while they are typing (interactive MT) or by translating the entire content at once and then displaying it for PE.
The TAUS report also mentioned that the translation industry will move towards MT as PE in the future. We believe, our work on automating the PE process is well placed to support this foreseen real-world CAT scenario.

In general, MT is useful for several applications and the demand for it is set to increase as the technology improves.
Broadly, MT can be used for understanding content in a foreign language, publishing research work across the globe, or in day-to-day activities such as chat and emails.
Achieving human-quality translation for various languages remains an elusive goal.
However, it is possible to achieve high-quality translation for simpler domains, such as weather forecasts or summaries of sport events, and even for interactive applications - such as rail or flight information - where the vocabulary is constrained to few words.
For more complex domains like manuals, patents and news translations, MT engines can provide a raw translation that can be refined by humans.

\section{Machine Translation}
MT is a sub-field of computational linguistics that focuses on automatic text translation from one natural language to another.
The research began more than six decades ago, when early MT systems were based on the rule-based paradigm.
Many approaches were explored, including the following:
\textit{i)}  \textit{direct translation}, which uses a bilingual dictionary to replace the source word or phrase with a corresponding translation; 
\textit{ii)}  \textit{transfer}, which performs syntactic analysis on the source text and then transfers this analysis into the target language to synthesize the target text; and 
\textit{iii)}  \textit{interlingua}, which converts the source text to an abstract meaning representation and then generates the translation.
These approaches are language-dependent and thus hardly scalable across languages.
For every new language, a new set of rules must be formalized, which is expensive in terms of time, labor and cost.

The next most popular MT paradigm is based on statistical methods.
The goal of statistical MT (SMT) is to automatically learn translation rules from the data without any need for linguistic knowledge.
Therefore, this approach is scalable across languages.
The building blocks of SMT were formulated
by proposing a word-based SMT technique based on the noisy channel model framework \cite{brown1993mathematics}.
The goal is to select the translation that maximizes the posterior probability
\begin{equation}
    e_{best} = argmax_{e}p(e|f)
\end{equation}
where $e$ and $f$ are, respectively, the target (translation) and the source sentences.
In this approach, words form the atomic unit of translation, which may result in poor translation if one word must be translated to several target words or vice versa.
Also, the meaning of a word depends on the context, and this approach might not generate the correct sense for highly polysemous words.
The immediate solution to this problem is to move the atomic translation unit from the word-level to the phrase-level to capture a larger context during translation, \cite{koehn2004statistical}.
In the next section, we discuss the phrase-based MT approach. 
This was the state-of-the-art approach for a long time until the emergence of the neural MT paradigm, which now dominates the field.

\subsection{Phrase-based Statistical MT (PBSMT)}
\label{sec:PBSMT}
As the name suggests, in PBSMT, phrases form the atomic unit of translation.
These phrases are basically contiguous sequences of words, not necessarily linguistically motivated.
The translation occurs by first splitting an input sentence into a sequence of phrases, which are then mapped one-to-one to target phrases.
The phrase translations may be reordered to obey the target language syntax.
The input sentence can be split in several ways, and each split can generate multiple translation hypothesis.
Hence, the whole translation process eventually results in a vast search space for finding the best hypothesis.
To speed-up the decoding process, various heuristics are used, such as reducing the number of translation options for each input phrase or selecting a fixed number of hypotheses at each decoding step.

The PBSMT models are usually trained with a sentence-aligned parallel corpus.
Initially, a lexical translation model is estimated to capture a word-to-word translation probability distribution, which inherently also learns the word alignment for the parallel corpus.
Then the phrase pairs are extracted using word-alignment information, and a phrase translation model is estimated based on how often a specific phrase pair co-occurs in the corpus.
These models, along with additional components - such as a language model and reordering model - are combined in a log-linear framework, where each model is considered as a feature.
This framework makes it possible to integrate additional arbitrary features, such as the number of words or phrase pairs used to generate the translation.
All features are weighted in an optimization step, which usually follows a discriminative training procedure. 
The system can then be used for decoding test data.
Below, we discuss the training procedure for the core models, the tuning mechanism of  feature weights, and the decoding process of the phrase-based technology. \\
\textbf{Lexical Translation Model:}
A general approach to estimate the lexical translation model is to scan the parallel corpus, check the translations of each source word, collect counts and estimate a probability distribution using maximum likelihood estimation.
The problem is that most of the existing parallel corpora are sentence-aligned and lack word-to-word alignment.
Without alignment information we cannot compute the counts and ultimately would not be able to estimate the model.

This problem is addressed by the expectation maximization (EM) algorithm.
The EM algorithm is an iterative learning procedure that executes the following steps until the model converges: \textit{i)} it applies the model to the data (expectation step); and \textit{ii)} it learns the model from the data (maximization step).
The model is initialized with a uniform probability distribution, when no prior knowledge is available.
In the expectation step, the model is applied to the data to obtain word alignments.
Initially, all alignments are equally probable or have equal weights; later, as the algorithm proceeds, the alignment converges to the most likely translation.
In the maximization step, the model is learned from the word-aligned data using maximum likelihood estimation, which considers the counts obtained from alignments that were weighted by the model in the expectation step.
We show the mathematical formulation of the EM algorithm for IBM model 1, as presented in literature \cite{Koehn:2010:SMT:1734086}, which is the foundation for building an SMT system.
This model was originally proposed along with more advanced models, from IBM model 2 to 5 \cite{Brown:1993:MSM:972470.972474}, which further considered absolute alignment position, the fertility factor, relative alignment distance, and other deficiencies of model 1.

In the expectation step, the model is used to compute the probabilities of different word alignments given a sentence pair in the data.
More formally, we compute $p(a|\textbf{e,f})$ that is the probability of an alignment ($a$) given the source ($\textbf{f}$) and target ($\textbf{e}$) sentence.
By applying the chain rule, we get:
\begin{equation}
\label{eq:align-generic}
p(a|\textbf{e,f}) = \frac{p(\textbf{e},a|\textbf{f})}{p(\textbf{e}|\textbf{f})}
\end{equation}
where $p(\textbf{e}|\textbf{f})$ is the probability of translating the source sentence to target, which is computed by:
\begin{equation}
p(\textbf{e}|\textbf{f}) = \sum_{a}p(\textbf{e},a|\textbf{f})
\end{equation}
that is marginalizing the translation over all possible alignments, which on further expansion gives:
\begin{equation}
p(\textbf{e}|\textbf{f}) = \sum_{a(1)=0}^{l_{f}}\cdots\sum_{a(l_{\textbf{e}})=0}^{l_{\textbf{f}}}p(\textbf{e},a|\textbf{f})
\end{equation}
where $l_{f}$ and $l_{e}$ are the length of the source and target sentence respectively, and $a(x)$ is the alignment function that returns the source index to which the target index $x$ is aligned to.
\begin{equation}
p(\textbf{e}|\textbf{f}) = \frac{\epsilon}{(l_{f}+1)^{l_{e}}}\sum_{a(1)=0}^{l_{f}}\cdots\sum_{a(l_{e})=0}^{l_{f}}\prod_{j=1}^{l_{e}}t(e_{j}|f_{a(j)})
\label{eq:sop}
\end{equation}
where $f_{x}$ and $e_{y}$ are the $x^{th}$ source and $y^{th}$ target word, $t(e_{j}|f_{a(j)})$ is the translation probability, and $\epsilon$ is a normalization constant, so that $p(e,a|f)$ is a probability distribution.
This equation first computes translation probability for each alignment and then sums the probabilities over all the possible alignments.
It has an exponential number of operations which makes it infeasible to compute in practice.
Therefore, another way to re-write the equation is to first compute the translation probability of each target word given all the source words and then apply chain rule to obtain the final translation probability of the entire sentence.
\begin{equation}
\label{eq:prob_e_to_f}
p(\textbf{e}|\textbf{f}) = \frac{\epsilon}{(l_{f}+1)^{l_{e}}}\prod_{j=1}^{l_{e}}\sum_{i=0}^{l_{f}}t(e_{j}|f_{i})
\end{equation}
Now we use equation $\S$\ref{eq:prob_e_to_f} in equation $\S$\ref{eq:align-generic}
\begin{equation}
p(a|\textbf{e,f}) = \frac{\frac{\epsilon}{(l_{f}+1)^{l_{e}}}\prod_{j=1}^{l_{e}}t(e_{j}|f_{i})}{\frac{\epsilon}{(l_{f}+1)^{l_{e}}}\prod_{j=1}^{l_{e}}\sum_{i=0}^{l_{f}}t(e_{j}|f_{i})}
\end{equation}
\begin{equation}
\label{eq:e-step}
p(a|\textbf{e,f}) = \prod_{j=1}^{l_{e}}\frac{t(e_{j}|f_{a(j)})}{\sum_{i=0}^{l_{f}}t(e_{j}|f_{i})}
\end{equation}
The equation \ref{eq:e-step} is the E-step in the EM algorithm, and it shows how the model is applied on the data.

Once all possible alignments are computed, the M-step will learn the lexical translation model from this data.
For this purpose, the count of each source word ($f$) co-occurring with a target word ($e$) is estimated from a given sentence pair (\textbf{$\textbf{e, f}$}) as follows:
\begin{equation}
c(e|f;\textbf{e, f}) = \sum_{a}p(a|\textbf{e, f})\sum_{j=1}^{l_{e}}\delta(e, e_{j})\delta(f, f_{a(j)})
\end{equation}
where $\delta(x,y)$ is the Kronecker delta function which is 1 if $x=y$ and 0 otherwise.
Using the formula in equation $\S$\ref{eq:e-step} we get
\begin{equation}
c(e|f;\textbf{e, f}) = \frac{t(e|f)}{\sum_{i=0}^{l_{f}}t(e|f_{i})}\sum_{j=1}^{l_{e}}\delta(e,e_{j})\sum_{i=0}^{l_{f}}\delta(f,f_{i})
\end{equation}
Using this count formula, we can estimate the new lexical translation probability distribution by
\begin{equation}
\label{eq:m-step}
t(e|f;\textbf{e,f}) = \frac{\sum_{(\textbf{e,f})}c(e|f;\textbf{e,f})}{\sum_{e}\sum_{(\textbf{e,f})}c(e|f;\textbf{e,f})}
\end{equation}
We iterate through these two steps until the model converges.\\
\textbf{Phrase Translation Model:}
Two natural languages do not always have one-to-one word correspondence.
Often, one word should be translated to several words or vice versa, especially in the case of multi-words and idioms.
The lexical model that translates word-by-word is not the best option to address this problem.
A better option is to move the atomic unit of translation higher, from word-level to phrase-level representation.
Phrase-based translation can resolve translation ambiguity more successfully through the availability of a larger context.
It also assists by learning longer and longer phrases, maybe even memorizing an entire sentence to reproduce the exact translation.

\begin{figure*}[th]
    \centering
    \includegraphics[scale=0.5]{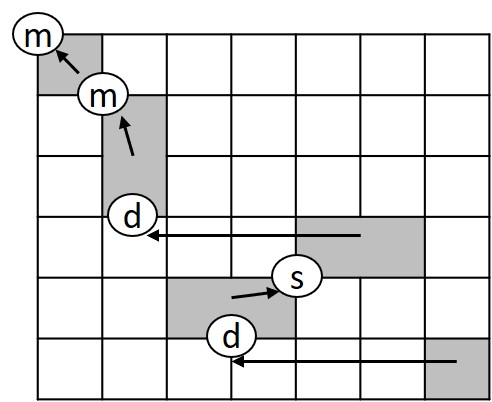}
    \caption{Different orientation of phrases in a lexicalized reordering model.}
    \label{fig:reordering}
\end{figure*}

To build the phrase translation model, first the valid phrase pairs were extracted from parallel sentences with the word alignment information.
Then the phrase translation probability distribution was computed using the maximum likelihood estimation technique.
\begin{equation}
p(\Bar{f}|\Bar{e}) = \frac{count(\Bar{e},\Bar{f})}{\sum_{\Bar{f_{i}}}count(\Bar{e},\Bar{f_{i}})}
\end{equation}
This gave the probability of translating the target phrase $\Bar{e}$ to the source phrase $\Bar{f}$.
Similarly, the reverse probability was estimated by the following equation:
\begin{equation}
p(\Bar{e}|\Bar{f}) = \frac{count(\Bar{f},\Bar{e})}{\sum_{\Bar{e_{i}}}count(\Bar{f},\Bar{e_{i}})}
\end{equation}
These bi-directional phrase translation models were used together as a feature function in the log-linear framework.\\
\textbf{Reordering Model:}
Translation generated by the above models does not follow the target language syntax.
The reordering model carefully reorders the translated phrases in appropriate position to improve the fluency of the output.
This is crucial for syntactically diverse language pairs, such as pairs involving like Japanese/German or English, where the verb at the end of a source sentence must be reordered during translation into English.
Here, we discuss the lexicalized reordering model, which is most widely used in PBSMT.
This model conditions the reordering of an actual phrase pair, and considers three orientations: \textit{i)} monotone order ($m$); \textit{ii)} swap with previous phrase ($s$); and \textit{iii)} discontinuous ($d$).
Basically, the model predicts the orientation type given the phrase pairs $p_{o}(orientation|\Bar{f},\Bar{e})$, as estimated by equation
\begin{equation}
p_{o}(orientation|\Bar{f},\Bar{e}) = \frac{\sigma * p(orientation) + count(orientation,\Bar{e},\Bar{f})}{\sigma + \sum_{o}count(o,\Bar{e},\Bar{f})}
\end{equation}
where $\sigma$ is a smoothing factor and $p(orientation)$ is given by
\begin{equation}
    p(orientation) = \frac{\sum_{\Bar{f}}\sum_{\Bar{e}}count(orientation,\Bar{e},\Bar{f})}{\sum_{o}\sum_{\Bar{f}}\sum_{\Bar{e}}count(o,\Bar{e},\Bar{f})}
\end{equation}
The orientation type is detected from the word alignment information shown in Figure \ref{fig:reordering}, where rows and columns indicate the target and source words, respectively.
An orientation is monotone (m) if a word alignment points to the top left; swap (s) if the alignment points to the top right; and discontinuous (d) if it is neither of those.
Further variations are discussed in the literature \cite{Koehn:2010:SMT:1734086}.\\
\textbf{Language Model:}
The language model (LM) measures the fluency of a sentence.
It is among the essential components of a PBSMT system.
This model helps to generate syntactically correct word order and, to select the correct translation for a specific context.
The most common LM mechanism in PBSMT is based on n-gram which uses statistics regarding the odds of one word following another.
Formally, the probability of a word $w_{n}$ given previous $m$ words is computed as follows:
\begin{equation}
    p(w_{n}|w_{n-m},...w_{n-2},w_{n-1}) = \frac{count(w_{n-m},...w_{n-2},w_{n-1},w_{n})}{\sum_{w}count(w_{n-m},...w_{n-2},w_{n-1},w)}
\end{equation}
Because we train the model with finite training data, it is possible to observe n-grams in the test set that are not observed during training; thus, the above model would assign them 0 probability.
A simple approach to overcome this issue is to use add-1 smoothing \cite{lidstone1920note,johnson1932probability}.
This step adds a fixed number (e.g. 1) to every n-gram occurrence and then estimates the probability.
Currently, the most widely used smoothing technique is Kneser-Ney \cite{kneser1995improved}.
It considers the diversity of word histories.
A word with a diverse history - that is, having many unique phrases that precede it - will obtain a higher probability.
The probability is computed as follows:
\begin{equation}
    p_{kn}(w) = \frac{N_{1+}(w)}{\sum_{w_{i}}N_{1+}(w_{i},w)}
\end{equation}
where $N_{1+} = |{w_{i}:c(w_{i},w)>0}|$\\
In this thesis, we use the IRSTLM \cite{federico2008irstlm} and KenLM \cite{heafield2011kenlm} toolkits for training n-gram LM with Kneser-Ney smoothing.\\
\textbf{PBSMT Decoding:}
The decoding process in PBSMT starts with by segmenting a source sentence into phrases.
For each source phrase, a phrase translation table that contains the source and target phrase pairs is queried, to retrieve translation options that are scored by different models.
Then, the phrase translations are stitched together to form the final translation.
Because a source sentence is segmented in different ways to generate sequences of different phrases, where phrases often have several translation options, this eventually leads to a vast search space.
Heuristics are used to reduce the search space and to generate an n-best list.
The highest scoring hypothesis in this n-best list is selected as the final translation.

The decoding in PBSMT is based on the log-linear model, a popular framework in ML, which takes the following form:
\begin{equation}
    p(x) = exp\sum_{i=1}^{n}\lambda_{i}h_{i}(x)
\end{equation}
where $n$ is the total number of features, and
$h_{i}$ and $\lambda_{i}$ represent a feature and its corresponding weight, respectively.
The goal is to select the translation $e$ that has the highest score.
\begin{equation}
    e_{best} = argmax_{e}p(e,a|f)
\end{equation}
\begin{equation}
\begin{aligned}
    p(e,a|f) = exp[\lambda_{PT}\sum_{i=1}^{I}log\ \textbf{PT}(\Bar{f_{i}}|\Bar{e_{i}}) 
    + \lambda_{LT}\sum_{i=1}^{I}log\ \textbf{LT}(\Bar{f_{i}}|\Bar{e_{i})}
    \\+ \lambda_{RO}\sum_{i=1}^{I}log\ \textbf{RO}(\Bar{f_{i}}|\Bar{e_{i}})
    + \lambda_{LM}\sum_{i=1}^{|e|}log\ \textbf{LM}(\Bar{e_{i}}|e_{1}...e_{i-1})
\end{aligned}
\end{equation}
where $PT$, $LT$, $RO$, and $LM$ refer to phrase translation, lexical translation, reordering, and language models, respectively.
In addition to these main models, other features can be added in this framework. 
Examples of additional features are word penalty, which controls the length of the translation, and phrase penalty, which helps to decide whether to translate into longer or shorter input phrases.
All phrase-based systems reported in this thesis were built with the Moses toolkit \cite{koehn2007moses}, which facilitates training, tuning, and decoding.\\
\textbf{Feature Weight Optimization:}
All the models discussed above contribute to the final decoding stage.
However, their importance varies depending on the text genre, domain, and languages.
For a close language pair like French and English, the phrase-translation feature might be more important than the reordering feature.
By contrast, for a structurally diverse language pair like English and German, the reordering feature might have higher importance.
Therefore, instead of each model being weighted equally, the model weights are usually tuned over a small held-out development set before the test data are decoded.
This is achieved by a discriminative training procedure like minimum error rate training (MERT) \cite{och2003minimum} or margin infused relaxed algorithm (MIRA) \cite{cherry2012batch,crammer2003ultraconservative,hasler2011margin}.
These two procedure are the most common tuning algorithms used in the PBSMT framework.
MERT is effective for tuning a few feature weights, typically ranging from 10 to 20, whereas MIRA is more scalable to millions of parameters.
The goal of these optimizers is to minimize translation error, measured with automatic MT evaluation metrics like BLEU and TER, which are discussed in detail in Section \ref{sec:MTeval}.

\subsection{Neural Machine Translation (NMT)}
\label{sec:NMT}
\begin{figure*}[th]
    \centering
    \includegraphics[scale=0.35]{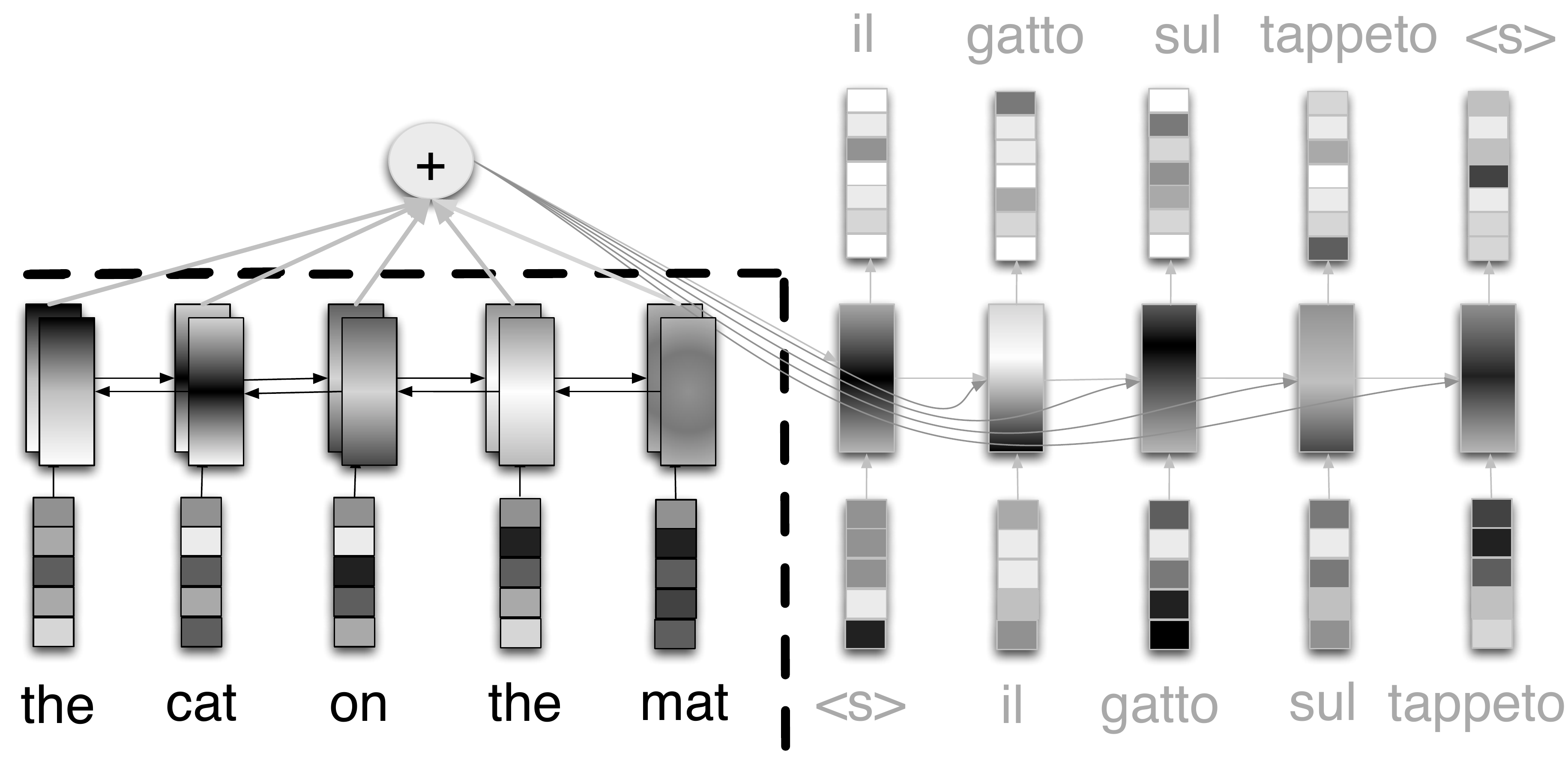}
    \caption{Time unfolded representation of the neural MT encoder-decoder architecture with bi-directional hidden states and attention model.}
    \label{fig:NMT}
\end{figure*}

Neural networks, especially deep neural networks (DNN), have emerged as a promising technology for solving various ML tasks.
Early neural networks include convolution \cite{lecun1998gradient} for image processing, recurrent \cite{hopfield1982neural} and recursive \cite{goller1996learning} for sequence modelling.
These networks have been enhanced over the years with advanced technologies, such as gated recurrent unit \cite{cho2014learning}, residual network \cite{he2016deep}, and capsule network \cite{sabour2017dynamic}, among others.
In the last few years, DNN has gained popularity in MT by achieving state-of-the-art results \cite{bojar-EtAl:2016:WMT1,bojar-EtAl:2017:WMT1,bojar-EtAl:2018:WMT1,wu2016google,gehring2017convolutional, vaswani2017attention}.
This cutting-edge technology has better generalization power than the phrase-based MT technique, as it operates in a continuous vector space representation instead of the discrete raw text of phrase-based MT.
The neural MT technology used in this work was based on the state-of-the-art sequence-to-sequence encoder-decoder architecture, with an attention mechanism as proposed in \cite{bahdanau2014neural} and extended by other researchers \cite{Luong2015StanfordNM,sennrich-haddow-birch:2016:WMT}. 
The core architecture is summarized in a temporal representation, shown in Figure~\ref{fig:NMT}.
The bottom left panel depicts the encoder network, which reads one input word at a time, encodes it into an embedding vector, and further encodes it together with its left (right) context into forward (backward) hidden state. 
Once the full input has been read, the decoder network (right panel) starts to generate the translation. 
The first word is generated by initializing the output sequence with the start symbol {\tt <s>}, which is mapped to an embedding and  further encoded into a (recursive) hidden state of the decoder, which also depends on the source context.
The source context is obtained by summing the hidden states of the encoder, where each state is first weighted by an attention model \cite{bahdanau2014neural} before the actual sum is performed.

The hidden state of the decoder is mapped into a probability distribution over the output vocabulary, using a scoring matrix and softmax operator. 
The most probable word ({\tt il}) is selected as the first output, and is used as input for the next inference step. 
The process continues 
until the sentence boundary symbol ({\tt <s>}) is emitted. 
In reality, the process is more complicated than the one shown in Figure \ref{fig:NMT} because a beam of $k$ most likely words is picked, and is used to infer the top $k$ words of the next output word.  

Now, we provide a mathematical definition of the computations performed by the network.  
Neural MT is aimed optimizing the parameters of a model to maximize the likelihood of the observed data.
The ultimate goal is to estimate a conditional probability model $p_{\Theta}(y|x)$, where  $\Theta$ is the parameter set of the model (weights and biases of the network), $y$ is a target sentence, and $x$ is a source sentence. 
Thus, the objective function is:
\begin{equation}
\underset{\Theta}{argmax}\frac{1}{N}\sum_{n=1}^{N}log(p_{\Theta}(y_{n}|x_{n}));
\end{equation}
where $N$ is the total number of sentence pairs in the training corpus.
The conditional probability is computed as follows:
\begin{equation}
p_{\Theta}(y|x) = \prod_{t=1}^{T_{y}}p_{\Theta}(y_{t}|y_{<t},x)
\end{equation}
where $T_{y}$ is the number of words in the target sentence. 
Given all the previous target words $y_{<t}$ and the source $x$, the probability of target word $y_{t}$ is modelled by the decoder network as follows:
\begin{equation}
p_{\Theta}(y_{t}|y_{<t},x) = g(\dot{y}_{t-1},s_{t},c_{t})
\end{equation}
where $\dot{y}_{t-1}$ is the word embedding of the previous target word, $s_{t}$ is the hidden state of the decoder, and $c_{t}$ refers to the source context vector (encoding of the source sentence $x$) at time $t$.
The decoder state $s_{t}$ is computed by a conditional gated recurrent unit (GRU) \cite{cho2014learning} in two steps.
First, \textit{i)} the previous hidden state and the previous target word embedding are  used to compute an intermediate hidden state by a GRU unit, as follows:
\begin{equation}
s'_{t} = f'(s_{t-1}, \dot{y}_{t-1})
\label{eq:ist}
\end{equation}
Second, \textit{ii)} the intermediate hidden state and the source context vector are passed to another GRU to compute the final hidden state of the decoder. 
The calculation can be summarized as follows:
\begin{equation}
s_{t} = f(s'_{t-1}, c_{t})
\label{eq:st}
\end{equation}
The source context vector is a weighted sum of all the hidden states of a bi-directional encoder \cite{bahdanau2014neural}.\footnote{In rest of the thesis, by encoder we mean bi-directional encoder}
\begin{equation}
c_{t} = \sum_{j=1}^{T_{x}}a_{tj}h_{j}
\end{equation}
where $a_{tj}$ is the attention weight given to the $j$-th encoder hidden state at decoding time $t$, and $T_{x}$ is the number of words in the source sentence.
The attention weight represents the importance of the $j$-th hidden state of the encoder in generating the target word at time $t$.
The weight is drawn from a probability distribution over all hidden states of the encoder, computed by applying a softmax operator over all the scores of the hidden units of encoder.
\begin{equation}
a_{tj} = \frac{exp(e_{tj})}{\sum_{k=1}^{T_{x}}exp(e_{tk})}
\end{equation}
where $e_{tj}$ and $e_{tk}$ are the scores of the $j$-th and $k$-th hidden units of the encoder at time-step $t$.
These scores are a function of the intermediate hidden state of the decoder (mentioned in Equation \ref{eq:ist}) and the hidden state of the encoder, as shown below:
\begin{equation}
e_{tj} = a(s'_{t-1}, h_{j})
\end{equation}
The hidden state $h_{j}$ of the $j$-th source word is a concatenation of the hidden states of the forward and backward encoders:
\begin{equation}
h_{j} = [\overrightarrow{h_{j}}; \overleftarrow{h_{j}}]
\end{equation}
where $\overrightarrow{h_{j}}$ and $\overleftarrow{h_{j}}$ are the hidden state of the forward and backward encoders, respectively.
These hidden states are computed by the GRU unit, which uses the previous or next hidden state and the word-embedding of the $j$-th 
source word ($\dot{x}_{j}$):
\begin{equation}
\overrightarrow{h}_{j} = f(\dot{x}_{j}, \overrightarrow{h}_{j-1})
\end{equation}
\begin{equation}
\overleftarrow{h}_{j} = f(\dot{x}_{j}, \overleftarrow{h}_{j+1})
\end{equation}
All neural systems reported in this thesis were built with the Nematus toolkit \cite{sennrich-haddow-birch:2016:WMT}.
Our contributions to the neural system were also built using this toolkit.

\section{MT Evaluation}
\label{sec:MTeval}
Human evaluation of MT considers various aspects of translation quality, such as adequacy and fluency \cite{hovy1999toward,white1994arpa}.
Thus, the evaluation is quite expensive and can be time-consuming.
This causes a bottleneck for MT developers to evaluate their ideas and for MT users to select the best engines for specific language pairs.
To overcome this issue, in this section we discuss the two most widely used automatic MT evaluation metrics, namely BLEU and TER. 
They are inexpensive, language independent, and strongly correlate with human judgment.
We also discuss a simpler version of human evaluation that uses the direct assessment technique, which is fast and reliable.

\subsection{Automatic Evaluation}
\label{sec:metric_auto}
What constitutes a good automatic MT evaluation metric is still an open question. 
Different metrics have been proposed over the years to address specific issues in measuring translation quality \cite{bojar-EtAl:2015:WMT,bojar-graham-kamran:2017:WMT,bojar-EtAl:2016:WMT2,specia-EtAl:2018:WMT}.
A special shared task to examine automatic metrics is organized every year at WMT \cite{callisonburch-EtAl:2012:WMT,bojar-EtAl:2013:WMT,bojar-EtAl:2014:W14-33,bojar-EtAl:2015:WMT,bojar-EtAl:2016:WMT2,bojar-graham-kamran:2017:WMT}.
For this task, participants are provided with MT output and corresponding human reference translations.
Participants are asked to submit the system-level or sentence-level automatic metric score for the MT output.
The submissions are ranked based on a correlation score, computed with reference to human judgments.
Out of several reference-based evaluation metrics, such as BLEU \cite{papineni2002bleu}, TER \cite{snover2006study}, CharacTER \cite{wang-EtAl:2016:WMT}, METEOR \cite{banerjee2005meteor}, and BEER \cite{stanojevic-simaan:2014:W14-33}, we use the first two in this thesis.
They are discussed in detail below:

\begin{itemize}
\item \textbf{BLEU \cite{papineni2002bleu}:}
The Bilingual Evaluation Understudy (BLEU) is based on modified n-gram precision. 
It identifies how many n-grams in the candidate translation occur in the reference translation, for an entire test set. 
It is thus considered a corpus-level evaluation metric.
The modified n-gram precision ($p_{n}$) is computed as follows:

{\centering
{\fontsize{0.4cm}{1em}
\begin{math}
p_{n} = \frac{\sum\limits_{C\in\{Candidate\}}\sum\limits_{\text{n-gram} \in\{C\}}Count_{clip}(\text{n-gram})}{\sum\limits_{C'\in\{Candidate\}}\sum\limits_{\text{n-gram'} \in\{C'\}}Count(\text{n-gram'})}
\end{math}
}

}
where $Count_{clip}$ indicates how many candidate n-grams were found in the reference.
To assign importance to each n-gram, the exponentially weighted (w) average of the logarithm of modified precision is considered.
To penalize shorter translations of the reference, a brevity penalty ($BP$) is added to the metric.
The BLEU score is then computed as follows:

{\centering
{\fontsize{0.4cm}{1em}
\begin{math}
BLEU = BP * exp^{\sum_{n=1}^{N}w_{n}log(p_{n})}
\end{math}
\begin{math}
where\ BP = \begin{cases} 
1, & if c>r \\ 
e^{(1-r/c)} & if c\leq r
\end{cases}
\end{math}
}

}
Where $c$ and $r$ refer to the length of the candidate translation hypothesis and the reference translations, respectively.
Being a measure of the overlap between a translation hypothesis and a reference translation, higher BLEU scores indicate better performance.
In this thesis, the BLEU score was computed using the \textit{multi-bleu.perl} package\footnote{\url{https://github.com/moses-smt/mosesdecoder/blob/master/scripts/generic/multi-bleu.perl}}, and statistical significance was computed using  paired bootstrap resampling  \cite{koehn2004statistical}.

\item \textbf{TER \cite{snover2006study}:}
Translation Error Rate (TER) measures the number of edits required to change a translation hypothesis to one of the references.
It is defined as follows:

{\centering
{\fontsize{0.4cm}{1em}
\begin{math}
TER = \frac{\#\ of\ edits}{average\ \#\ of\ reference\ words}
\end{math}
}

}

The number of edits consists of insertions, deletions, and substitution of single words as well as, shifts in word sequences.
Being an error metric, lower TER score indicate better performance.
In this thesis, the TER score was computed using TERcom\footnote{\url{http://www.cs.umd.edu/~snover/tercom/}} software, and statistical significance was computed using stratified approximate randomization \cite{clark2011better}.

The TER tool basically computes a score for each sentence and then calculates an average for the whole test set.
The sentence-level scores are helpful for comparing the translation quality of two systems for each source sentence.
This comparison is particularly useful if a translation system re-translates the output of a given MT system with the aim of further improving the translation quality.
In such cases, it is interesting to know how many of the modified translations led to an improvement in the re-translation process, which is basically the precision of the second translation system.
More formally, precision is computed as the ratio of the number of translations that were improved, over all the translations that were modified, as shown below.\footnote{Two translation that receive different TER score is considered as modified.}

{\centering
{\fontsize{0.5cm}{1em}
\begin{math}
Precision = \dfrac{Number\ of\ Sentences\ Improved}{Number\ of\ Sentences\ Modified}
\end{math}
}

}
A precision score of 1 indicates that all translations modified during re-translation resulted in improvements.
We use this measure quite often to compare the performance of an automatic PE system, which can be viewed as a second translation system, with that of an MT system.
\end{itemize}
BLEU is the most widely used automatic metric but it has several shortcomings: lack of synonym matching, inability to detect proper word order, and lack of reliability for sentence-level evaluation.
Some of these limitations are overcome by the TER metric.
However, unlike BLEU, TER does not exploit all the
references but scores each hypothesis against its reference in isolation.

\subsection{Human Evaluation}
Although automatic metrics are faster and cheaper, human evaluation provides stronger evidence about the quality of a translation.
Therefore, it is considered the main evaluation metric for ranking systems that are submitted for the shared translation task at the machine translation conference (WMT).
However, manual evaluation is not only expensive but requires domain knowledge.
For example, ranking-based evaluation requires comparing the output of different systems and ranking them according to the adequacy and fluency of the translations.
Evaluation through a PE approach requires the editor to edit a translation to correct errors in the system's output. 
Systems that require the least PE effort are considered the best.
Although this method of evaluation is  time-consuming, it provides insights into the types of errors made by systems.
Here we discuss a simpler version of human evaluation, carried out using the direct assessment (DA) technique \cite{graham-EtAl:2013:LAW7-ID,NLE:9961497}.
DA is used to assess the quality of the output of MT systems and produces a ranking based on human  judgment. It also analyzes how humans perceive TER/BLEU performance differences between the systems.
DA elicits human assessments of translation adequacy on an analogue rating scale (0--100), where human assessors are asked to rate how adequately the MT system output expresses the meaning of the reference translation.
Individual assessments collected through an analogue scale contain some degree of random error and cannot alone provide a reliable estimate of translation quality. 
However, when enough DA judgments are combined into mean scores, positive and negative random error (that is truly random) cancels itself out in score distributions, yielding estimates of translation
quality that are highly reliable \cite{grahametal:15}.

\section{Summary}
In this chapter, we provided background about MT, which along with human PE has become the \textit{de-facto} standard in the translation industry.
When LSPs and translators use MT engines in a CAT framework, they can speed-up their work and this increases the productivity of the translation industry.
Although human-quality translation remains an elusive goal, it is achievable for simple domains having constrained language with a limited vocabulary.
The translation quality achieved by current MT technology is, however, quite useful for ``gisting'', where the goal is to give an idea of the context.
We discussed the two most popular MT paradigms, namely \textit{phrase-based} and \textit{neural}, which form the basis of our work on automatic post-editing.
The former has been the most widely used technology for more than a decade, whereas the latter has emerged as a strong alternative that provides state-of-the-art performance.
At the end of the chapter, we discussed automatic and human evaluation metrics to measure the quality of MT output.
Automatic metrics provide a faster and cheaper means of evaluating a system's performance. 
However, human evaluation is more effective and is used as the main metric to evaluate systems submitted for the translation task in the MT conference (WMT).

\chapter{Automatic Post-Editing}
In this chapter, we introduce the APE task and motivate its applicability within a CAT framework for assisting professional translators.
We highlight its benefits and show how it complements an MT engine to improve translation quality.
We review prior work that addresses several challenges in APE, and cover different paradigms, learning modes, domains of data, among other aspects.
Finally, we discuss the evolution of APE through shared tasks at the conference on machine translation (WMT), which witnessed a paradigm shift from phrase-based to neural-based technology, leading to state-of-the-art results.

\section{Introduction}
In the previous chapter, we discussed the current trend among LSPs to semi-automate the translation process with the help of CAT tools.
Instead of translating from scratch, MT engines are first used to obtain a ``raw'' translation, which is then sent to the translators for further verification and correction to yield a publishable text.
This two-stage approach is effective when the MT quality is high, thereby reducing the human PE effort.
Given the advancement in MT technology in recent years, with significant gains in translation performance, it is predicted that the translation market will shift towards MT post-editing in the near future.\footnote{\url{https://www.taus.net/think-tank/reports}}
The semi-automated translation process with PE generates vast parallel data, comprising MT output and human post-edits, as a by-product of the translation workflow.
These data can be leveraged to learn error correction rules and apply them on new MT data, ranging from fixing typos to adopting terminology to a specific domain, or even modeling the personal style of an individual translator.
This way of automatically post-editing MT output can reduce overheads incurred by repeated manual corrections and may eventually improve translation productivity.

\begin{figure}
\center
\includegraphics[width=9cm, height=8cm]{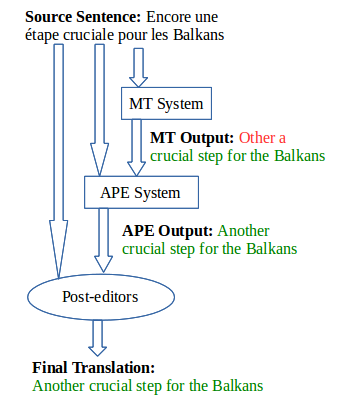}
\caption{\label{fig:CAT}APE component in CAT framework(fr-en)}
\end{figure}

APE can be seen as a ``monolingual'' translation task \cite{simard2007statistical}, and the same MT technology can be used to develop APE systems.
Unlike MT, where the system learns bilingual translations from source and target pairs (\textit{src}, \textit{trg}), APE learns to correct errors from MT text and human PEs (\textit{mt}, \textit{pe}) or from triplets (\textit{src}, \textit{mt}, \textit{pe}) to leverage the source context too.
Most of the current state-of-the-art systems are built using triplets \cite{chatterjee-EtAl:2017:WMT2,junczysdowmunt-grundkiewicz:2018:WMT,tebbifakhr-EtAl:2018:WMT}.
Once developed, an APE engine can be deployed as an interface between the MT system and the post-editor, such that it receives the source and corresponding MT text (\textit{src}, \textit{mt}) and in return provides the APE version for human correction, as shown in Figure \ref{fig:CAT}.
In the figure, the source text ``\textit{Encore une étape cruciale pour les Balkans}'' is first translated by the MT system as ``\textit{Other a crucial step for the Balkans}''.
The error in the MT output ``\textit{Other a}'' is corrected by the APE system, and the final translation ``\textit{Another crucial step for the Balkans}'' is given to the post-editor.
Because the APE system has fixed all the MT errors in this given example, the human post-editor reviews it without performing further modifications.

\section{Motivation}
As mentioned in Chapter \ref{chap:intro}, MT systems are imperfect and tend to make systematic errors that human post-editors must correct before publication.
Moreover, these systems are usually modelled as a function of source to target, where the same input causes the system to generate the same output.
Therefore, an erroneous mapping between source and target causes the system to re-generate the same error over and over again for the same input.
Repeated MT errors make the human PE task cumbersome.
Here, an APE tool is useful to correct frequent MT errors and thus reduce the workload of humans.
This tool becomes the only viable option to improve translation quality when the underlying MT engine is a ``black-box''.
This situation is observed when LSPs use third-party translation engines, which cannot be adapted to customers' data.
From an application point of view, the APE task is motivated by the following possible uses ~\cite{P15-2026}:
\begin{itemize}
\item Improve MT output by exploiting information  unavailable  to  the  decoder,  or  by  performing deeper text analysis that is too expensive at the decoding stage;
\item Cope with systematic errors of an MT system whose decoding process is not accessible;
 \item Provide professional translators with improved MT output quality to reduce human PE effort;
\item Adapt the output of a general-purpose MT system to the lexicon or style requested in a specific application domain.
\end{itemize}

\section{Literature Review}
This section categorizes prior work in APE, based on various factors involved in system design.

\subsection{APE Paradigm: Rule-based, Phrase-based, or Neural}
\label{subsec:paradigm}
Translation industries within  multilingual environments need their systems to be portable across domains.
Rule-based APE systems \cite{knight1994automated,elming2006transformation}, generally have precise PE rules as they are manually written.
However, these hand-crafted rules are insufficient to capture all possible scenarios.
Moreover, for each new language, a new set of rules must be written, which increases the complexity in terms of cost and time.
By contrast, data-driven approaches - such as phrase-based and neural learn error-correction rules automatically from the data.
This helps to speed-up the system development cycle and, given data availability, makes the system portable across languages. However, the cost is less reliable rules.
Phrase-based technology has been used in most work since the beginning of APE \cite{simard2007statistical, dugast2007statistical,simard2007rule,dugast2009statistical,lagarda2009statistical,pilevar2011using,terumasa2007rule,bechara2011statistical,bechara2012evaluation,chatterjee-turchi-negri:2015:WMT,chatterjee-EtAl:2016:WMT}.
It provides faster training and decoding capabilities and can learn efficiently from small datasets containing a few ten of thousands of training samples (especially for narrow domains).
Neural technology has emerged as a stronger alternative to phrase-based methods, achieving state-of-the-art performance in APE \cite{junczysdowmunt-grundkiewicz:2016:WMT,chatterjee-EtAl:2017:WMT2,pal-EtAl:2016:P16-2,pal-EtAl:2017:EACLshort,junczysdowmunt-grundkiewicz:2017:WMT}.
However systems based on deep neural nets are expensive to train in terms of computation power, as more graphics processing units (GPUs) and time (from several days to weeks) are required, compared with phrase-based systems.
They also need large training datasets to improve their generalization power.
Small datasets often lead to the problem of overfitting, where the system performs well on the training data but poor on the unseen test set.

The first rule-based approach was proposed by \cite{knight1994automated}, who had addressed the problem of article selection in English noun phrases when translating from Japanese to English.
They extracted noun phrases from the Wall Street Journal and built decision trees using various linguistic features, achieving 78\% accuracy for financial data.
Later, \cite{allen2000toward} proposed to use human PE data to learn PE rules, an approach which was later implemented in \cite{elming2006transformation}.
They used a transformation-based learning (TBL) technique to learn context-dependent substitution rules based on the word forms and part of speech tags, and obtained an improvement of 4.6\% in the BLEU score above a baseline system that kept MT output untouched.

After that, phrase-based statistical APE (SAPE) dominated the field for a while.
Many studies \cite{simard2007statistical, dugast2007statistical,simard2007rule,dugast2009statistical,lagarda2009statistical,pilevar2011using,terumasa2007rule} showed that SAPE can significantly improve the translation quality of a rule-based MT (RBMT) system, although some studies \cite{bechara2011statistical,bechara2012evaluation} showed that SAPE of SMT offered limited improvement in the translation quality.\footnote{Although both phrase-based and neural technologies are based on the statistics of data, when we refer to statistical systems we mean systems based on phrase-based technology, neural-based systems are explicitly mentioned.}
Our study \cite{P15-2026} showed that for domain-specific data, SAPE significantly improved the translation quality of SMT output.
This motivated us to initially investigate the statistical phrase-based paradigm. It has the potential to improve both the rule-based and the statistical MT output and can easily be scaled in a multilingual translation environment.

Recent research on deep learning techniques for the APE task has shown neural technology to hold more promise than phrase-based for improving the translation quality of PBSMT systems \cite{berard-besacier-pietquin:2017:WMT,chatterjee-EtAl:2017:WMT2,hokamp:2017:WMT,junczysdowmunt-grundkiewicz:2016:WMT,junczysdowmunt-grundkiewicz:2017:WMT,pal-EtAl:2017:EACLshort,pal-EtAl:2016:P16-2,tan-EtAl:2017:WMT2,varis-bojar:2017:WMT}.
All submissions in the 2017 and 2018 APE shared tasks were based on neural technology \cite{bojar-EtAl:2017:WMT1,bojar-EtAl:2018:WMT1}.
Most of them were built around the standard sequence-to-sequence encoder-decoder architecture (see Section \ref{sec:NMT}), and the state-of-the-art system added a multi-source plugin for the encoder.
To mitigate the problem of limited training data, most of the submissions used synthetic data to build generic models, which were then fine-tuned using task-specific data.

\subsection{MT Engine Accessibility: Glass-box vs. Black-box}
The metadata and internal decoding trace of an MT engine provides crucial information for designing the APE architecture.
When this information is accessible - that is, a glass-box scenario, the APE system has the privilege to encode various MT system level information. Such information includes word/phrase alignment between input and output text; top-k candidate translations; and scores for various components, such as the translation model, language model and other features.
These additional signals, along with human feedback make the glass-box a favorable condition for developing good APE systems.
However, a \textit{tight coupling} with the MT engine might restrict APE tools from being easily plugged-in with other MT back-ends.
In this scenario, it is possible to directly update the MT system itself.
However, re-training huge MT models that are trained on billions of tokens might not be  feasible for customizing translation for a particular customer.
In the case of a black-box scenario, where no information about the MT engine is known a priori, the APE system must depend on triplets (\textit{src}, \textit{mt}, \textit{pe}) to learn error-correction rules.
Being loosely coupled with the MT engine, the APE tool might be compatible with other MT back-ends, especially when different back-ends share the same technology.
This scenario is more challenging and is often faced by LSPs who rely on third-party translation engines, which do not allow for customization or fine-tuning to a customer's requirements.
In this thesis, we consider the black-box scenario to evaluate the potential of APE for improving translation quality, and to enhance existing technology address several challenges. These issues are discussed in the next chapters.

Most of the SAPE and NAPE work discussed in Section \ref{subsec:paradigm} successfully improved the translation quality of black-box MT systems, without considering any system-level information.
The glass-box scenario has not been explored much, mainly because APE solutions depend on a specific MT architecture and are not easy to generalize.
Mundt et al. \cite{mundt2012learning} proposed an APE solution to detect and correct named entities and content words in the source that are dropped (not translated) or mistranslated.
Alignment information from the MT decoding trace is used to detect the errors, and the correction is performed by retrieving valid options from a lookup table and inserting them in the most probable position given by a classifier.
Yang et al. \cite{yang2014novel} proposed a rule refinement method in which they extracted error-correction rules between the MT output and the reference using the TER metric.
These rules, when combined with the original SMT rules, showed a positive impact on the final translation quality.

\subsection{Type of Post-Editing Data: Simulated vs. Real}
The type of PE data has a vast influence on a system's performance.
The simulated PE data are easily available in the form of reference translations of the source sentences inside a parallel corpus.
Because target references are created independently of any MT system, they do not explicitly capture MT PE rules.
This type of data is  available in large quantities but it is not highly useful for building a good APE system because the translations do not reflect the correction of MT errors \cite{potet2012towards}.
In contrast, \textit{real} PE data are generated during the translation process. An MT engine is first used to obtain a raw translation, which is later verified and, if needed, corrected by post-editors.
This type of data captures the error-correction patterns; hence, it is highly useful for building APE systems.
The origin of such data affects the final translation quality.
Post-edited data of a professional quality will maintain consistency in applying corrections throughout the text. Crowd-sourced PEs would be of lower quality, with inconsistent and possibly incorrect corrections \cite{bojar-EtAl:2016:WMT1}.

Experiments with real PE data were first performed \cite{simard2007statistical} to develop a SAPE system to improve the translation of an RBMT system.
According to the findings, the SAPE system improved the output of the RBMT system, and the combination of RBMT with SAPE resulted in a better quality of translation than a standalone PBSMT system.
The impact of real and simulated PE data for improving the output of a PBSMT system using a SAPE tool was first examined by \cite{potet2012towards}.
They discovered that SAPE did not improve the translation of a general domain, irrespective of the type of data (simulated or real) used during the training phase.
However, the simulated data caused greater damage than the real data.
The positive gains of using real PE data were also observed in the outcomes of two rounds of the APE shared task \cite{bojar-EtAl:2016:WMT1,bojar-EtAl:2017:WMT1}, in which participants improved the output of an English-German (IT domain) PBSMT system by between 5 and 10 BLEU points.

\subsection{Domain of the Data: Generic vs. Specific}
The domain of the data has several attributes that can affect the learning ability of the APE system.
Domain can be broadly categorized into generic and specific to aid in clustering the different attributes, and to understand their relationship with a system's performance.
Here, we consider a generic domain to be a mix of different domains or topics, such as ``news'' or ``parliamentary proceedings''.
Large amounts of generic data with simulated PEs are freely available on the internet and can be used to learn a basic translation mapping between two languages. However, this is not useful to fix MT errors.
Because generic data covers diverse topics, they are usually sparse with a low repetition rate (RR) \cite{Cettolo:2014:amta} making it difficult to collect enough statistical data to estimate reliable scores for the PE rules.\footnote{RR measures the repetitiveness inside a text by looking at the rate of non-singleton n-gram types (n=1. . .4) and combining them using the geometric mean. Larger value means more repetitions in the text.}
In contrast, domain-specific data are more constrained and coherent in terms of vocabulary size, phrase structure, and text style or format. In addition, they usually have a higher RR, which facilitates learning reliable PE rules.

The contrasting natures of the two domains and their effects on system performance are evident in the results of the APE shared tasks \cite{bojar-EtAl:2016:WMT1}.
The pilot round \cite{bojar-EtAl:2015:WMT} released generic domain news data with crowdsourced PE having a low RR of 3.0.
The dataset was so challenging that none of the submitted systems (at that time using the phrase-based approach) could improve the translation quality.
In contrast, the outcome of the next APE round \cite{bojar-EtAl:2016:WMT2} was impressive; the evaluation was performed on domain-specific data (information technology) with professional PEs that had a higher RR.
Some of the submitted runs and the official APE baseline (phrase-based technology) were able to improve the quality of the MT text significantly.

Most researchers have concluded that a generic SAPE system performs well on the output of the RBMT system \cite{simard2007statistical,simard2007rule,terumasa2007rule,dugast2007statistical}.
This conclusion is not confirmed when the MT back-end uses phrase-based technology.
Post-processing PBSMT output with the SAPE technique for generic data yields limited improvement and sometimes damages the overall MT quality \cite{bechara2011statistical,bechara2012evaluation,potet2012towards,bojar-EtAl:2015:WMT}.
Because one of the main applications of MT is localization and other areas focus on domain-specific translation, most APE studies involve domain-specific data.
In \cite{de2008statistical}, the authors used the \textit{Labour Agreement} corpus to train a domain specific SAPE, which showed significant improvements over both RBMT and SMT. 
They also compared a general SAPE system with a domain-specific SAPE and found the latter more effective than the former.
A similar conclusion was drawn by \cite{potet2012towards}. 
They showed that a domain-specific SAPE built on a ``water science'' corpus, significantly-reduced the error rate by 20\% over the baseline PBSMT system. 
By contrast, the generic SAPE system showed no positive effect.

\subsection{Learning Mode: Offline vs. Online}
The potential of APE is generally investigated in controlled environments, in which static APE systems are trained on a batch of training samples and evaluated on a held-out test set.
In this offline (batch) learning mode, the representativeness of the training data with respect to the test data is crucial in achieving good performance.
The more the test data diverge from the training set, the more the system's performance deteriorates.
Ideally, to be integrated in a real professional translation  workflow,  APE  tools  should  be  flexible  enough  to handle continuous streams of diverse data coming from different domains or genres.
The real-time changes in the nature of data are modeled in an online learning framework, by incorporating feedback from each test sample back into the system before the next sample is processed.
Online learning mechanisms can easily be integrated with CAT tools. This makes the PE task easier for translators, through learning from past sample and improving the translation quality of the next sample.

Most of the works discussed in the above sections were based on batch learning.
Because APE is gaining attention from industries \cite{DBLP:journals/corr/CregoKKRYSABCDE16,grangier-EtAl:2017:arxiv,mathur-ueffing-leusch:2017:INLG2017}, researchers are targeting more realistic online learning scenarios.
Hardt and Elming \cite{hardt2010incremental} stated that online learning has inherent benefits.
First, PE rules learned from a sentence can significantly improve the translation quality of neighboring sentences, as they share a similar context within the document.
Second, the more recent a PE rule is, the more valuable its contribution.
Simard and Foster \cite{simard2013pepr} later proposed an online learning SAPE framework called ``PEPr (post-edit propagation)''.
It operates at the document level.
For each input document, a new set of PE rules is extracted when the document is processed.
This strategy showed significant improvement in translating technical documents, probably due to the high RR of phrases in the same document.
Recently, \cite{lagarda2015translating} performed large scale SAPE experiments in an online learning mode.
They assessed their technique on different corpora, namely EMEA, Xerox and i3media, which covered diverse domains (health, technical, and news). Many languages were used, as were various MT paradigms (rule-based, web-based or statistical).
Their study showed that SAPE systems for health and technical domains significantly outperformed others because of the high RR of these documents.

\section{Findings from APE shared task at WMT}
\label{sec:apefindings}
To monitor the state-of-the-art performance of APE systems, we host an APE shared task every year since 2015 at WMT - where other tasks are also host such as news translation, quality estimation, and evaluation metrics \cite{bojar-EtAl:2015:WMT,bojar-EtAl:2016:WMT1,bojar-EtAl:2017:WMT1}.
The systems submitted to the APE shared task are evaluated over benchmark datasets using a fair evaluation framework consisting of automatic and manual evaluation. 
Past experiences and lessons learned from the outcome of each round are discussed here. 
This section briefly summarizes the shared-task findings for 2015 to 2017, with a focus on \textit{i)} the impact of various data attributes on system performance, and \textit{ii)} the evolution of technology over the years.

The APE shared task focuses on the challenges posed by the ``black-box'' scenario, in which the MT system is unknown and cannot be modified. 
In this scenario, APE methods must operate at the downstream level - that is, \textit{after} MT decoding. 
The methods can apply either rule-based techniques, statistical, or neural approaches to exploit knowledge acquired from human PEs that are provided as training material.
The training and development data consist of (\textit{source, target, human PE}) triplets, and the participants are asked to return automatic PEs for a test set of unseen (\textit{source, target}) pairs.
Overall, the potential of APE to improve MT is evaluated with both generic and domain-specific data. The submissions are evaluated using automatic metrics, such as TER and BLEU, with TER being used as a primary metric to rank all submissions.
To further produce rankings based on human judgements, and to analyze how humans perceive TER and BLEU performance differences between the submitted systems, a human evaluation is also performed.
The official baseline results are the TER and BLEU scores  calculated by comparing the raw MT output with the human PEs. In practice, the baseline APE system is a ``\textit{do-nothing''} system that leaves all the test targets unmodified.
In addition, participating systems are evaluated against a system that is a re-implementation of an earlier approach ~\cite{simard2007statistical}. This can be seen as a monolingual translation task that aims to re-translate the MT output for further improvement.
This APE system was built with the Moses toolkit~\cite{koehn2007moses};  translation and reordering models are estimated following the Moses protocol, with a default setup using MGIZA++~\cite{gao2008parallel} for word alignment.
Language models were built with the KenLM toolkit ~\cite{heafield2011kenlm} for standard $n$-gram modeling with an $n$-gram length of 5. Finally, the  system was tuned on the development set, optimizing TER/BLEU with Minimum-Error-Rate Training~\cite{och2003minimum}.

\subsection{The Pilot Round (2015)}
The pilot round of the shared task \cite{bojar-EtAl:2015:WMT} released a generic domain news dataset for  English to Spanish translation, containing $\sim$12,000, $\sim$1,000 and $\sim$2,000 samples respectively in the training, development, and test sets. The post-edits were collected through crowdsourcing.
Four teams participated and submitted a total of seven runs.
All the submissions were based on the same phrase-based technology, with some variations.
The official results achieved by the participating systems are reported in Table~\ref{tab:APEresultsCS}.
\begin{table*}[t]{
\centering

\begin{tabular}{l|c}
ID                  & \begin{tabular}[c]{@{}l@{}}Avg. TER \end{tabular} \\ \hline
\rowcolor[HTML]{EFEFEF} Baseline     & 22.913                                           \\ 
FBK     Primary               & 23.228                                                  \\
LIMSI   Primary               & 23.331                                                  \\
USAAR-SAPE                    & 23.426                                                  \\
LIMSI   Contrastive           & 23.573                                                  \\
Abu-MaTran Primary            & 23.639                                                  \\ 
FBK     Contrastive           & 23.649                                                  \\ 
\rowcolor[HTML]{EFEFEF} {\color[HTML]{656565}(Simard et. al)~\cite{simard2007statistical}} & {\color[HTML]{656565} 23.839}      \\ 
Abu-MaTran Contrastive         & 24.715                                                 \\ \hline
\end{tabular}
\caption{\label{tab:APEresultsCS} Official results for the WMT15 Automatic Post-editing  task.}
}
\end{table*}
The rankings revealed an unexpected outcome: none of the submitted runs could beat the baseline. 
In addition, all differences from the baseline were statistically significant. 
In practice, this means that what the systems learned from the available data was not reliable enough to yield valid corrections of the test instances. 
As shown in Table~\ref{tab:APEresultsCS}, the performance achieved by a baseline implementation of~\cite{simard2007statistical} was  $23.8$ TER. 
Compared to  this score, most of the submitted runs performed better, with statistically significant results, indicating that the novelties introduced by each system were useful.

To better understand why APE did not yield the expected gain, the possible relation between participants' results and the nature of the data was investigated. The quantity, sparsity, domain and origin of the data were considered.
For this purpose the Autodesk Post-Editing Data corpus was used which possesses orthogonal features compared to the task data.\footnote{The corpus, (\url{https://autodesk.app.box.com/Autodesk-Post Editing})}
Specially, Autodesk data mainly cover the domain of \texttt{software user manuals} (that is, a restricted domain rather than a general one like \texttt{news}). In addition, PEs are performed by professional translators and are thus expected to be better than those of a crowdsourced workforce.
Different monolingual indicators were compared for both the datasets, namely token/type ratio, and RR.
The lower token/type ratio and RR of the task data (9.5 and 3 respectively), compared to Autodesk data (21 and 8.4 respectively), indicated
that information was repeated less often.
This monolingual analysis, however, did not indicate the repetitiveness of (\textit{error,correction}) pairs.
To investigate this aspect, a re-implemented approach was trained on both data sets ~\cite{simard2007statistical}. 
The distribution of the frequency of the  translation options in the phrase table is considered a good indicator of the level of repetitiveness of (\textit{error,correction}) pairs.
Many translation options that appear only once or a few times indicate a higher level of sparseness. 
As expected, due to the small training set, most of the translation options in both phrase tables were singletons (95.2\% and 84.6\% respectively for the APE task and Autodesk data).
Nevertheless, the Autodesk phrase table was more compact (731k versus 1,066k) and contained 10\% fewer singletons than the APE table. This confirms that the APE task data were sparser. Hence, it might be easier to learn applicable correction patterns from the Autodesk domain-specific data. Indeed, when the Autodesk  system was evaluated on its own test set, an error reduction of $3.55$ TER points was observed compared with the ``do-nothing'' baseline.
This result confirms the intuitions about the usefulness of repetitive data. It also showed that, in restricted-domain scenarios at least, APE can be  successfully used as an aid to improve the output of an MT system. 

The impact of PE origin (\textit{i.e} crowdsourced vs professional) was also investigated.
Unlike the Autodesk data, for which PEs were created by professional translators, the APE task data contained crowdsourced MT corrections by unknown translators, who likely were non-expert. One risk, given the high variability of valid MT corrections, is that the crowdsourced workforce upholds PE attitudes and criteria that  differ from those of professional translators. Professionals tend to maximize their productivity by making only necessary and sufficient corrections to  improve the translation quality. In addition, they follow consistent translation criteria, especially  for domain terminology. Such practices result in coherent and minimally edited  data, from which learning and drawing statistics are easier. The same is not assured when crowdsourced workers are involved, who do not have specific time or consistency constraints. 
The possible consequence is that non-professional PEs and the correction patterns learned from them will display relatively less coherence, higher noise, and higher sparsity. 
To assess the potential impact of these issues on data representativeness and, in turn, on task difficulty, a subset of the APE test instances (221 triplets randomly sampled) was analyzed. 
The target sentences of the subset were post-edited by professional translators. 
The analysis focused on TER scores computed between \textit{i)} the original target sentences and their crowdsourced PEs (avg. TER = $26.02$); \textit{ii)} the target sentences and their  professional PEs (avg. TER = $23.85$); \textit{iii)} the crowdsourced PEs and the professional ones, using the latter as references (avg. TER = $29.18$). The measured values indicate an attitude of non-professionals to correct \textit{more} and \textit{differently} compared to professional translators. Interestingly, the distance between crowdsourced versus professional PEs was larger than the distance between original target sentences and experts' corrections ($29.18$ vs. $23.85$). If we consider the output of professional translators as a gold standard for coherent  and  minimally post-edited data, these scores suggest greater difficulty in handling crowdsourced corrections. 

\subsection{The Second Round (2016)}
The second round of the shared task released a domain-specific (IT) dataset for English-German language pairs. They contained $\sim$12,000, $\sim$1,000, and $\sim$2,000 samples respectively in training, development, and test sets, where PEs were collected from professional translators \cite{bojar-EtAl:2016:WMT1}.
Six teams participated and submitted a total of 11 runs.
Most submissions had the same backbone as the previous round (phrase-based technology), whereas one or two explored the potential of deep neural networks.
The official results achieved by the participating systems are reported in Table \ref{tab:APEresultsCS2016}.
\begin{table}[h!]
\centering
\begin{tabular}{l|c|c}
ID                  & \begin{tabular}[c]{@{}l@{}}Avg. TER ($\downarrow$) \end{tabular} & \begin{tabular}[c]{@{}l@{}}BLEU ($\uparrow$) \end{tabular} \\ \hline
AMU Primary & 21.52 & 67.65 \\
AMU Contrastive & 23.06 & 66.09  \\
FBK Contrastive & 23.92 & 64.75  \\
FBK Primary &  23.94 & 64.75 \\
USAAR Primary & 24.14 & 64.10 \\
USAAR Constrastive & 24.14 & 64.00 \\
CUNI Primary &  24.31 & 63.32 \\
\rowcolor[HTML]{EFEFEF}{\color[HTML]{656565}(Simard et al.)\cite{simard2007statistical}} & {\color[HTML]{656565} 24.64 }     &  {\color[HTML]{656565} 63.47 } \\ 
\rowcolor[HTML]{EFEFEF} Baseline     & 24.76 & 62.11                                           \\ 

DCU Contrastive & 26.79 & 58.60  \\
JUSAAR Primary & 26.92 & 59.44  \\
JUSAAR Contrastive & 26.97 & 59.18  \\
DCU Primary & 28.97 & 55.19  \\ \hline
\end{tabular}
\caption{\label{tab:APEresultsCS2016} Official results for the WMT16 Automatic Post-editing  task.}
\end{table}
In the pilot round~\cite{bojar-EtAl:2015:WMT}, none of the runs had been able to beat the baselines. In contrast, in this round, half the participants achieved this goal by producing APE sentences that resulted in lower TER (with a maximum of -3.24 points) and higher BLEU scores (up to +5.54 points). All differences relative to such baselines were statistically significant. These results suggests that the correction patterns learned from the data were reliable enough to allow most systems to effectively correct the original MT output. 

The next question was whether the improvements observed in this round were due to the new dataset - that is, domain-specific texts and professional PEs. They could also have resulted from a real technology jump, that is, the use of neural end-to-end APE systems, factored, or operational sequential models. A partial answer was given by the performance of the phrase-based approach \cite{simard2007statistical}, which we ran on the data of both rounds of the APE task with the same implementation. Although the results for the two test sets were difficult to compare, partly due to the different language settings, the overall TER scores and relative distances (with respect to other submitted runs) provided some indications.
For the pilot  test set, the basic statistical APE method damaged the original MT output quality, with a TER increment of about 1 point. 
On 2016 data, it achieved a small improvement. 
This result suggests that - as hypothesized before - the repetitiveness featured by the selected data can facilitate the work of APE systems. 
In the new scenario, repetition rates for SRC, TGT, and PE were more than double the values measured  in the previous round.
Thus, these methods can learn a larger number of reliable and re-applicable correction patterns from the training data.
However,  the large improvements obtained by the top runs were only reached by moving from the basic statistical MT backbone to a new and more reliable APE solution based on deep neural nets. 
The winning system used vast artificial data created by a round-trip translation process, in which the source text was first translated to target, and then the target was translated back to source using phrase-based MT systems.
With this latest approach, the distance between the baseline and the top-ranked systems increased from 0.6 to 3.12 TER points. On the one hand, the new data made the task easier; on the other hand the deployed solutions and the increased distance in results (compared with the basic statistical APE approach) indicate a substantial step forward.

\subsection{The Third Round (2017)}
\label{sec:aperound2017}
The third round of the shared task
extended the previous round by including German-English as a new language direction, in addition to English-German \cite{bojar-EtAl:2017:WMT1}.
For both directions, participants operated with domain-specific data, namely information technology for EN-DE and medical for DE-EN.
The PEs were collected from professional translators.
For English-German, the training and test sets contained 11,000 and 2,000 triplets, respectively. The data released in the previous round of the task (15,000 instances) and the  artificially generated PE triplets  (4 million instances) \cite{junczysdowmunt-grundkiewicz:2016:WMT} were also provided as additional training material.  
For German-English, training and development sets contained 25,000 and 1,000 triplets respectively, and the test set consisted of 2,000 instances.
Seven teams participated in the English-German task and submitted a total of 15 runs. Two of those teams also participated in the German-English task with five submitted runs.
The official results are shown in Table \ref{tab:APEresultsENDE2017} and Table \ref{tab:APEresultsDEEN} respectively for EN-DE and DE-EN language pairs.
\paragraph{English-German}
Compared to previous rounds of the APE task, the most noticeable aspect is that in this round, for the first time,  all participants beat the MT baseline with their primary submission. This steady improvement was mainly driven by the widespread migration to the neural approach. In this round, the gains regarding English-German data were even larger, with the winning system scoring -4.88 TER and +7.58 BLEU points better than the MT baseline. The technology advancement  is evident if we look at our second term of comparison: the re-implementation of the phrase-based approach \cite{simard2007statistical}. In the previous round, English-German, the results of this method were better than the baseline and were placed in the middle of the official participants' rankings. By contrast, in this round the results were almost identical to the baseline. However, they were worse than the baseline in terms of TER (+0.21) and were slightly better than baseline in terms of BLEU (+0.48). They were competitive regarding the contrastive submission of one participant only. We considered the  distance between the same phrase-based approach and the baseline as an indicator of task difficulty across different rounds of the task. 
We hypothesized that the good results achieved by this round's participants were mainly due to improved techniques rather than ``easier'' test data.
Indeed, for English-German where a comparison with the previous round was possible, the close repetition rate and BLEU scores revealed a similar level of difficulty for the test data.
\begin{table}[t!]
\centering
\begin{tabular}{l|c|c}
ID                  & \begin{tabular}[c]{@{}l@{}}Avg. TER ($\downarrow$) \end{tabular} & \begin{tabular}[c]{@{}l@{}}BLEU ($\uparrow$)  \end{tabular} \\ \hline

FBK Primary                 &   19.6                &   70.07   \\
AMU Primary                 &   19.77               &   69.5   \\
AMU Contrastive             &   19.83               &   69.38   \\
DCU Primary                 &   20.11               &   69.19   \\
DCU Contrastive             &   20.25               &   69.33   \\
FBK Contrastive             &   20.3                &   69.11   \\
FBK\_USAAR Contr.     &   21.55               &   67.28   \\
USAAR Primary               &   23.05               &   65.01   \\
LIG Primary                 &   23.22               &   65.12   \\
JXNU Primary                &   23.31               &   65.66   \\
LIG Contrastive-Forced             &   23.51               &   64.52   \\
LIG Contrastive-Chained             &   23.66               &   64.46   \\
CUNI Primary                &   24.03               &   64.28   \\
USAAR Contrastive           &   24.17               &   63.55   \\
\rowcolor[HTML]{EFEFEF} Baseline                    &   24.48               &   62.49   \\
\rowcolor[HTML]{EFEFEF}{\color[HTML]{656565}(Simard et al.)\cite{simard2007statistical}}    &   {\color[HTML]{656565} 24.69}       &   {\color[HTML]{656565} 62.97}   \\
CUNI Contrastive            &   25.94               &   61.65   \\ 
\end{tabular}
\caption{\label{tab:APEresultsENDE2017} 
Results for the WMT17 APE \textbf{EN-DE}  task.
}
\end{table}
\paragraph{German-English}
For German-English, the improvement of the top submission relative to the baseline was smaller (-0.26 TER, +0.28 BLEU) but still statistically significant. 
Smaller gains obtained by systems that used the same approaches as those for the English-German task confirmed that the level of difficulty differed for the two language directions. 
The interaction between low repetition rate (6.2 RR) and high translation quality for the initial MT output (79.5 BLEU) certainly played a role in reducing the gap between the primary submissions and the ``\textit{do-nothing}'' MT baseline. This would be an interesting aspect for more thorough exploration. 
However, in this case, the lowest results achieved by the phrase-based APE baseline confirm that switching to neural methods was a technological advancement in the right direction.

Overall, lessons learned from early works on statistical APE methods include the following points: \textit{i)} learning from human PE is more effective than learning from (independent) reference translations; \textit{ii)} learning from, and applying APE to, domain-specific data is more promising than working on general-domain data; and \textit{iii)} correcting the output of rule-based MT systems is easier than improving translations from statistical MT.
A missing aspect, however, is the  assessment of different methods under comparable conditions, and on different language pairs. We address these topics in the next chapter.
\begin{table}[t!]
\centering
\begin{tabular}{l|c|c}
ID                  & \begin{tabular}[c]{@{}l@{}}Avg. TER ($\downarrow$) \end{tabular} & \begin{tabular}[c]{@{}l@{}}BLEU ($\uparrow$)  \end{tabular} \\ \hline
FBK Primary                 &   15.29                &   79.82   \\
FBK Contrastive             &   15.31                &   79.64   \\
LIG Primary                 &   15.53               &   79.49   \\
\rowcolor[HTML]{EFEFEF} Baseline                    &   15.55               &   79.54   \\
LIG Contrastive-Forced             &   15.62               &   79.48   \\
LIG Contrastive-Chained             &   15.68               &   79.35   \\
\rowcolor[HTML]{EFEFEF}{\color[HTML]{656565}(Simard et al.)\cite{simard2007statistical}}    &   {\color[HTML]{656565} 15.74}       &   {\color[HTML]{656565} 79.28}   \\
\end{tabular}
\caption{\label{tab:APEresultsDEEN} 
Results for the WMT17 APE \textbf{DE-EN}  task.}
\end{table}
\section{Summary}
In this chapter, we introduced the APE task, which aims to simulate the human PE process of transforming raw MT output to a publishable text by automatically fixing systematic MT errors.
We discussed the benefits of combining such post-processing systems to MT engines for leveraging human feedback, as well as the possibility of applying external tools that are either expensive or cannot be to integrated with MT engines.
Several APE challenges related to data, technology, and application were briefly discussed, and are investigated in more detail in the next chapters.
We then reviewed prior work to capture different aspects of APE.
Different paradigms such as rule-based, phrase-based, and neural-based solutions were discussed.
Initial studies on APE were based on the rule-based approach, which is not scalable for multiple languages; hence, most work has focused on data-driven phrase-based methods.
Phrase-based technology gave state-of-the-art results for more than a decade until the growth of deep neural nets.
Other aspects explored in this chapter include the accessibility of MT engines (glass-box vs black-box), PE data origin (simulated vs real), domain (generic vs specific), and learning mode (offline vs online).
Finally, we summarized the APE task findings from the conference on machine translation. 
Our discussion focused on the challenges posed by the nature of data (quality, quantity, repetitiveness, domain, and origin) and the shift in paradigms from phrase-based to end-to-end deep learning based solutions.

\chapter{Phrase-based APE}
\label{chap:phrase-basedAPE}
In this chapter, we report our initial investigation of the effectiveness of phrase-based APE as a downstream task to improve the translation quality of a PBSMT output.
Prior studies reported contrasting views about the effectiveness of applying APE to the output of a phrase-based MT system.
No solid evidence is available about if and when APE might be useful. 
We thus performed the first systematic study to fill this gap, by evaluating two variants of phrase-based APE - namely \textit{monolingual} and \textit{context-aware}) - across many language pairs.
Furthermore, we propose a mechanism to combine these variants to mutually benefit from each other.
This technology tends to learn noisy and unreliable PE rules when operating on sparse datasets.
This issue is addressed in the second half of the chapter through our novel task-specific dense features.
The resulting approach was the core of our submission to the WMT 2015 APE shared task.

\section{Introduction}
As mentioned in the previous chapter, APE can be seen as a translation task for transforming raw MT output into better translations.
Therefore, most APE systems are based on MT technology. 
In this chapter, we explore the phrase-based MT approach to the APE task that has been widely used for more than a decade.
Earlier work on APE \cite{simard2007rule, dugast2007statistical,bechara2012evaluation,potet2012towards} reported contrasting results when using this technology to improve the output of a phrase-based MT system, as opposed to confirmed positive results when the underlying MT system was based on a rule-based paradigm.
To understand the potential of phrase-based APE of PBSMT output, we examined two approaches. 
One was the ``monolingual'' translation method \cite{simard2007statistical} that was widely used in most of the early works.
The other variant was the ``context-aware'' solution \cite{bechara2011statistical}, where the source words are annotated by the MT words to yield a new source corpus, which no longer represents a monolingual text but a mix of two languages.
To our knowledge, this represents the most significant variant of \cite{simard2007statistical} in the phrase-based technology.
We discuss these approaches in detail in the next sections.
The main contribution of this chapter is a systematic analysis of different APE approaches, which were tested in controlled conditions over several language pairs.
These approaches tend to learn noisy and unreliable PE rules when trained on sparse data, especially for generic domains.
For example, a source and target phrase pair occurring only once in the training data will have a probability of 1, but this might not be a reliable pair due to poor statistics (count of 1).
The phrase pair might also result from faulty word alignment and thus, should be avoided during decoding, but the high probability can force its use.
To address this issue, we propose novel task-specific features that measure similarity, reliability, and usefulness of the rule to help the system achieve better translation options.

\section{Monolingual APE} In this variant, the  underlying idea is that APE components can be trained in the same way as statistical MT systems, that is, starting from ``parallel data''. Since the goal is to transform raw MT output into its correct version, the parallel data  consist of MT output as source texts ($\textbf{f'}$) and correct (human quality) sentences as target ($\textbf{f}$). 
In \cite{simard2007statistical}, the authors used these pairs ($\textbf{f',f}$) to train a phrase-based APE system, which is then applied to correct the output of a commercial rule-based MT system. 
Positive evaluation results were reported for English-French language pairs and even better results for French-English pairs. In both cases, statistical APE yields significant BLEU and TER improvements over the original MT output. However, because the training and test data for the two language directions differ in content and size, the measured performance variations cannot be directly ascribed to the effectiveness of the method in the two settings.

\section{Context-aware APE} 
\label{subsec:m2}
A limitation of the above ``monolingual translation'' approach is that the basic statistical APE pipeline is trained only on data in the target language, disregarding information about the source language.
Correction rules learned from ($\textbf{f',f}$)  pairs lose the connection  between translated words or phrases  and the corresponding source terms ($e$). This implies that information lost or distorted in the translation process is beyond reach of the APE component, and the resulting errors are impossible to recover.

To overcome this issue, \cite{bechara2011statistical} proposed a ``context-aware'' variant to represent the data. For each word  $f'$, the corresponding source word or phrase $e$ is identified through word alignment, and is used to obtain a joint representation of $f'$ given $e$ ($f'\#e$). The result is an intermediate language $F'\#E$ that represents the new source side of the parallel data used to train the statistical APE component. 
An example to illustrate this joint representation is shown in Table \ref{tab:joint-rep}.
The source \textit{``See Paint on 3D models .''} and the MT output \textit{``Siehe Bemalen von 3D-Modellen .''} are combined based on word alignment information to form \textit{``Siehe\#See Bemalen\#Paint\_on von\#on 3D-Modellen\#3D\_models .''}.
Although in principle this method is more precise, it can be affected by two problems. First, preserving the source context comes at the cost of a larger vocabulary size and thus higher data sparseness. While the basic statistical APE pipeline combines and exploits the counts of all co-occurrences of $f'$ and $f$ in the parallel data, its context-aware variant considers each $f'\#e{_i}$ as a separate term, thus breaking down the co-occurrence counts of $f'$ and $f$ into smaller numbers. Second, all these counts can be influenced by word alignment errors. To cope with data sparseness and unreliable word alignment, \cite{bechara2011statistical} experimented with different thresholds set on word alignment strengths to filter context information. 
Specifically, they discarded the ($f'\#e, f$) pairs in which the $f'\#e$ alignment score was smaller than the threshold.
\begin{table}[h]
\begin{center}
\small
\begin{tabular}{|c|m{10cm}|}\hline
\bf Source & See Paint on 3D models .
\\\hline
\bf MT output & Siehe Bemalen von 3D-Modellen .
\\\hline
\bf Joint Representation & Siehe\#See Bemalen\#Paint\_on von\#on 3D-Modellen\#3D\_models .
\\\hline
\end{tabular}
\end{center}
\caption{\label{tab:joint-rep} An example of joint representation used in \textit{context-aware} translation.}
\end{table}

When the approach was applied to correct the output of a statistical phrase-based MT system, the evaluation results were ambiguous. 
For French-English,  significant improvements (up to 2.0 BLEU points) were observed, relative to both the baseline (original MT output)  and the basic ``monolingual'' method  \cite{simard2007statistical}. 
For English-French, however, the performance dropped slightly. 
Moreover, follow-up experiments  using the same method  \cite{BecharaThesis} did not confirm the findings.
In light of these ambiguous results and  the lack of a systematic comparison between the two APE methods, our objective was to replicate the studies.\footnote{This is done based on the description provided by the published works. Discrepancies with the actual methods are possible, due to our misinterpretation or to wrong guesses about details that are missing in the papers.}
We conducted a fair comparison, in a controlled evaluation setting, involving different language pairs.

\section{Effectiveness of APE for Domain-specific Data}
In this section, we use the above two APE variants to investigate the effectiveness of APE for domain-specific data.
Our work draws on previous findings; that is, we learned from in-domain PE data and applied APE to phrase-based MT.
Yet it fills a gap in prior research by providing a fair comparative study between different methods in controlled conditions. 
The key enabling factor was the availability - for the first time - of in-domain data that consisted of the \textit{same source sentences}, \textit{machine-translated in several languages}, and \textit{post-edited by professional translators}.

A missing aspect in previous evaluations  was the  assessment of different methods, \textit{i)} under comparable conditions, and  \textit{ii)} using different language pairs featuring varying levels of MT quality. Focusing on statistical APE 
methods, we proposed the first systematic analysis of two approaches.  To understand their potential, we compared them in the same conditions over six language pairs having English as the source.
Our results provided evidence of the following: \textit{i)} consistent improvements for all language pairs; \textit{ii)} a relationship between the extent of the gain and MT output quality; \textit{iii)} slight but statistically significant performance differences between the two methods; and \textit{iv)} possible complementarity between the methods.
To ensure a sound analysis, our experimental setup consisted of a dataset of the same English source sentences, with automatic translations into six languages and respective manual PEs by professional translators. Overall, this represented the ideal condition to complement earlier research and provide answers to previousy missing questions, such as:\\
\textbf{Q1:} \textit{Does APE yield consistent MT quality improvements across different language pairs?}\\
\textbf{Q2:} \textit{What is the relation between the original MT output quality and the APE results?}\\
\textbf{Q3:} \textit{Which of the two analyzed APE methods shows the best potential?}

\subsection{Reimplementing the two APE variants}
\label{subsec:reimplementing}
To obtain the  statistical APE  pipeline that represents the backbone of both methods, we used the Moses toolkit to develop the systems.  Our training data consisted of (\textit{source}, \textit{MT output}, \textit{post-edits}) triplets for six language pairs, all having English as the source. 
Although the  monolingual translation approach \textbf{(APE-1)} used only the last two elements of the triplet, all of them played a role in the context-aware variant \textbf{(APE-2)}. Apart from the differing data representation, the training process was identical.
Translation and reordering models were estimated following the Moses protocol with default setup using MGIZA++ \cite{gao2008parallel} for word 
alignment.\footnote{In APE-1, MGIZA++ is used to align $f'$ and $f$. In APE-2 it is used  to align  $f'$ and $e$, and then $f'\#e$ and $f$.}
For language modeling we used the KenLM toolkit \cite{heafield2011kenlm} for standard $n$-gram modeling with an $n$-gram length of 5. The APE system for each target language  was tuned on  comparable development sets, optimizing TER with minimum error rate training (MERT) \cite{och2003minimum} using the PE sentences as references.

\subsection{Experimental Setting}
\paragraph*{Data.} We experimented with the Autodesk Post-Editing Data corpus,\footnote{\url{https://autodesk.app.box.com/Autodesk-PostEditing}} which  mainly covers the domain of software user manuals. English sentences were translated into several languages yielding 30,000 to 410,000 translations per language.
The translations were performed by  Autodesk's in-house MT system \cite{zhechev2012machine} and were post-edited by professional translators.

\begin{table}[t]
\begin{center}
\begin{tabular}{|l|c|c|c|}
\hline \bf Lang. & \bf No. & \bf Vocab. & \bf No. \\ 
 & \bf  tokens & \bf  Size & \bf Lemmas \\\hline
En & 210,491 & 10,727 & 8,260\\\hline \hline
Cs & 202,475 & 16,716 & 10,137 \\\hline
De &211,149 & 17,563 & 14,368 \\ \hline
Es & 252,020 & 11,075 & 6,683\\\hline
Fr & 263,690 & 10,928 & 7,213 \\\hline
It & 239,912 & 10,703 & 6,549 \\\hline
Pl & 206,016 & 17,027 & 10,430 \\
\hline
\end{tabular}
\end{center}
\caption{\label{tab:data} Data statistics for each language.}
\end{table}

Our experiments were run on six language pairs having English as the source and  Czech, German, Spanish, French, Italian or Polish as the target.  To set up a controlled environment, we extracted all the (\textit{source}, \textit{MT output}, \textit{post-edits}) triplets sharing the same source (En) sentences, across all language pairs. Table \ref{tab:data} provides statistics about the resulting \textit{tri-parallel} corpora. After randomly shuffling the triplets, we created training data (12.2K triplets), development data (2K), and test data (2K) that shared exactly the same source sentences across languages. Training and evaluation of our APE systems were performed on true-case data.

To guarantee similar experimental conditions in the six language settings, we also trained comparable target language models from external data.
Indeed, the 12.2K PEs would not be enough to train reliable LMs.
We built our LMs from approximately 2.5M 
translations of the same English sentences collected from Europarl \cite{koehn2005europarl}, DGT-Translation Memory \cite{STEINBERGER12.814}, JRC Acquis \cite{steinberger06}, OPUS IT \cite{tiedemann2012}, and other Autodesk data common to all languages.

\begin{table*}[t]
\begin{center}
\begin{tabular}{|l|c||c|c|c||c|c|c||c|}
\hline \bf  & \bf MT Baseline & \multicolumn{3}{ |c|| }{\bf APE-1}  & \multicolumn{3}{ |c|| }{\bf APE-2}  & \bf Oracle \\\hline
&  \textit{TER}  &   \textit{TER} & \textit{$\Delta$} & \textit{\% $\Delta$}  &   \textit{TER}  & \textit{$\Delta$} & \textit{\% $\Delta$} &  \textit{TER} \\\hline
\bf EN-DE & 46.46 & 43.07 & -3.39 & 7.3  & \textbf{42.79}$^{\ast}$  & -3.67 & 7.9 & 40.17\\\hline
\bf EN-CS & 44.06 & 39.38 & -4.68 & 10.62 & \textbf{39.10}$^{\ast}$  & -4.96 & 11.25 & 36.32\\\hline
  \bf EN-PL & 43.02 & 38.24 & -4.78 & 11.11 & \textbf{37.75}$^{\ast}$  & -5.27 & 12.25 & 35.05\\\hline
\bf EN-IT & 34.44 & 30.43 & -4.01 & 11.64 & \textbf{30.13}$^{\ast}$  & -4.31 & 12.55 & 28.33\\\hline
\bf EN-FR & 32.76 & 29.70 & -3.06 & 9.34 & \textbf{29.51}  & -3.25 & 9.92 & 27.12\\\hline
\bf EN-ES & 30.90 & 26.69 & -4.21 & 13.62 & \textbf{26.35}$^{\ast}$  & -4.55 & 14.72 & 24.34\\\hline
\end{tabular}
\end{center}
\caption{\label{tab:autodesk_results}  Performance of the MT baseline and the APE methods for each language pair. Results for APE-2 marked with the ``$^{\ast}$'' symbol are statistically significant compared to APE-1.}
\end{table*}

\paragraph*{Baseline.} 
Similar to all previous works on APE, our baseline was the MT output \textit{as-is}.  Hence, baseline scores for each language pair corresponded to the TER computed between  the original MT output, produced by the ``black-box'' Autodesk in-house system, and human PEs.

\subsection{Results}
Table~\ref{tab:autodesk_results} presents our results, with  language pairs  ordered according to the respective baseline 
TER. The positive answer to \textbf{Q1} (\textit{``Does  APE  yield  consistent improvements to MT output?''}) is evident: both APE methods consistently improved MT quality for all  language pairs. 
TER reductions  ranged from 3.06 to 5.27 points. 
Quality improvements were statistically significant ($p<0.05$) as  measured by a bootstrap resampling test \cite{clark2011better}.

In answer to \textbf{Q2} (\textit{``What is the relation between original MT quality and APE results?''}), our controlled experiments indicated that the higher  the MT quality, the higher the improvement was.
This interesting result may seem counter-intuitive because more room for improvement is expected for sentences of poor quality.
However, it revealed that that learning from - and correcting - noisy data affected by many errors was difficult for statistical APE methods. This finding was absent for EN-FR, for which a reasonably good MT quality did not induce a  gain in performance comparable to  language pairs featuring similar MT 
 TER (EN-IT and EN-ES). 
 Further analysis of the data showed that all the target languages except French maintained coherent behavior with regard to domain-specific English terms, which were always either preserved (Italian) or translated (other languages). 
 French showed an alternation between the two patterns.
 An example is  the English ``\textit{workflow}'', which appeared in the French PEs either untranslated (21 sentences) or translated as ``\textit{flux de travail}'' (34 sentences). In contrast, in the other language directions, all occurrences of  ``\textit{workflow}''  were either translated consistently or were consistently kept in English.
These frequent ambiguities were difficult to manage, especially if the two forms occur roughly equally in the training data.
This might account for the smaller gains in quality observed for EN-FR than for other language pairs.

In answer to \textbf{Q3} (\textit{``Which method has the highest potential?''}), we observed slight 
TER reductions when moving from ``monolingual'' to ``context-aware'' variants.\footnote{Filtering the context information with thresholds between 0.6 and 0.8 leads to the best results for all languages.} Although small (between 0.19 and 0.49
TER points), such gains were statistically significant ($p<0.05$), except for EN-FR ($p<0.07$). This suggests that linking the MT words to the source terms can help to recover adequacy errors that are beyond the capabilities of APE-1. 

To better understand to what extent the two methods behaved differently, we calculated the results of an \textit{Oracle} system, similar to one proposed in literature \cite{rubinoEAMT2012}.
For each test sentence, we selected the best PE (lower TER) produced by the two approaches.
As shown in the last column of Table~\ref{tab:results}, this oracle achieved a significant TER reduction (from 1.8 to 2.78 points) for all the language pairs. We interpreted such gains as signs of possible complementarity between the two methods, which would be worth investigating. 

As mentioned in Section~\ref{subsec:m2}, an advantage of the monolingual approach is its robust estimation of translation parameters.  In contrast, by exploiting contextual information from the source, the context-aware method is more precise but can be affected by data sparsity issues due to its large vocabulary. In an attempt to use a less sparse model at the level of word alignment, we trained an APE system based on the context-aware representation, but with word alignment computed as in the case of the monolingual setup. 
This method of combining the two variants was tested on three language pairs (EN-DE, EN-FR and EN-CS).
The TER was reduced by 0.75 for EN-DE, by 0.60 for EN-CS, and by 0.53 for EN-FR, compared with the context-aware variant.
This result seems to validate our intuition about the possible complementarity of the two variants.

\section{Effectiveness of APE for Generic Data}
The previous section discussed the positive effect of APE on the output of a PBSMT system for domain-specific data.
For generic data, the challenge increases.
The system has to deal with increased sparsity, higher vocabulary size, and lower repetition rates.
This section evaluates whether the APE technology can obtain a positive outcome in this challenging scenario too.
For this study, we used the generic domain dataset released in the pilot round of the APE shared task \cite{bojar-EtAl:2015:WMT}, which featured characteristics suitable to make a fair evaluation.
Given that the domain was generic, the data were already sparse, and because the PEs were collected through crowdsourcing, they were further vulnerable to noise.
We devised mechanisms through novel task-specific dense features to learn a reliable model from this sparse and noisy dataset.
Our contributions became part of the system submission at the pilot round of the APE shared task at WMT 2015 \cite{bojar-EtAl:2015:WMT}.

\subsection{Experimental Settings}
\textbf{Data:} 
The training, development, and the test data consisted of $\sim$11K, 500, and 500 triplets respectively.
Our evaluation was based on the performance achieved on this test set.
Training and evaluation of our APE systems were performed on the true-case data.
Our APE pipeline consisted of various stages like language model selection, phrase table pruning, and feature designing, as discussed in the following sections.
We followed an incremental strategy to develop the APE systems.
At each stage of the APE pipeline, we found the best configuration of a component on a development set and proceeded to explore the next component.
\\
\textbf{Baseline:}
Our baseline was the MT output \textit{as-is}.
We used the corresponding human PE corpus for evaluating systems' performance, which yielded a \textbf{23.10} TER score.

\subsection{APE Pipeline}
\label{sec:pipeline}
In this section we describe various components that we explore at each stage of the pipeline.

\paragraph{Language Model Selection (APE-LM)}
We used various datasets to train multiple language models to see which had the highest impact on translation quality.
All the LMs were trained using the IRSTLM toolkit \cite{federico2008irstlm} having an order of 5-gram with Kneser-Ney smoothing.
The dataset varied in quality and quantity as described below.
\begin{itemize}
\item \textbf{LM1}
was built using only the training data ($\sim$11K). 
Although limited in amount, the data were quite reliable since they were sampled from the same distribution of the test set.
\item \textbf{LM2}
was built from the News Commentary corpus, which containis $\sim$200K sentences, downloaded from the WMT 2013 translation task.\footnote{\label{wmt2013}http://www.statmt.org/wmt13/translation-task.html} This corpus belongs to the same domain of the APE data but it is created under different conditions, involving professional translators and translating the source sentence from scratch.
This made it significantly different from the data used to build LM1.  
\item \textbf{LM3} (Big data)
was built from the News Crawl data from 2007-2012, contributing to $\sim$13M sentences, downloaded from the WMT 2013 translation task.
This dataset had vast amounts of news crawled from the Web and covering several topics.

\item \textbf{LM1+LM2+LM3:}
In this setting, all the previous language models were simultaneously used by the APE systems.
However, they were weighted through an optimization step over a held-out development set. 
\end{itemize}

\begin{table}[h]
\begin{center}
\begin{tabular}{|l|c|c|}
\hline \bf  & \bf APE-1 & \bf APE-2 \\\hline
\bf LM1 & \bf 23.95  & \bf 24.59 \\\hline
LM2 & 23.96 & 24.62 \\\hline
LM3 & 24.06 & 24.66 \\\hline
LM1+LM2+LM3 & 24.05 & 24.69 \\\hline
\end{tabular}
\end{center}
\caption{\label{tab:LM} Performance (TER score) of the APE systems using various LMs on dev set}
\end{table}

The impact of using different LMs for both variants of the phrase-based APE is shown in Table \ref{tab:LM}; again, APE-1 and APE-2 correspond to ``monolingual'' and ``context-aware'' variants respectively. 
We noticed that the performance of APE systems did not show much variation for different LMs.
This could be due to the fact that the \textit{news commentary} and \textit{news crawl} data might not closely resemble the shared task data.
For this reason, the in-domain LM1 was selected and was used to evaluate the next component. 
\\
\paragraph{Pruning Strategy (APE-LM1-Prun)}
\label{sec:prun}
To remove unreliable translation rules generated from the data obtained through crowd-sourcing, pruning strategies were investigated.
First, we tested the classic pruning technique \cite{johnson2007improving} which is based on the significance
testing of phrase pair co-occurrence in the
parallel corpus.
According to our experiments, this technique was too aggressive when applied on limited amounts of sparse data. 
Nearly 5\% of the phrase table was retained after pruning, mainly with self-rules - that is, translation options  containing the same source and target phrases.

We thus developed a novel feature for pruning, which measured the usefulness of a translation option present in the phrase table.
For each translation option in the phrase table, all the parallel sentences were retrieved from the training data such that the source phrase of the translation option was present in the source sentence of the parallel corpus. 
We then substituted the target phrase of the translation option in the source sentence of the parallel corpus and then computed the TER score with respect to the corresponding target sentence.
If TER increased, we incremented the \textit{neg-count} by 1; if TER decreased we incremented the \textit{pos-count} by 1.
Finally, we computed the \textit{neg-impact} and the \textit{pos-impact} as follows:
\\
\\
\textit{neg-impact} $=$ $\dfrac{\textit{neg-count}}{\textit{Number of Retrieved Sentences}}$
\\
\\
\textit{pos-impact} $=$ $\dfrac{\textit{pos-count}}{\textit{Number of Retrieved Sentences}}$
\\
\\
Once these ratios were computed for all the translation options, we filtered the phrase table by thresholding on the \textit{neg-impact} to remove rules which were not useful; the higher the \textit{neg-impact}, the less useful they were. 
All translation options with a score greater than or equal to the threshold value were filtered out. 
We applied this pruning strategy for both the APE methods over various threshold values.

\begin{table}[h]
\begin{center}
\begin{tabular}{|c|c|c|c|}
\hline \bf Threshold & \bf TER & \bf Number of sentences modified & \bf Precision \\\hline
0.8 & 23.90 & 88 & 0.12 \\\hline
0.6 & 23.91 & 90 & 0.13 \\\hline
\bf 0.4 & 23.98 & 94 & \bf 0.15 \\\hline
0.2 & 23.77 & 70 & 0.12 \\\hline
\end{tabular}
\end{center}
\caption{\label{tab:Filter1} Performance (TER score) of the APE-1-LM1 after pruning at various threshold values on dev set}
\end{table}

\begin{table}[h]
\begin{center}
\begin{tabular}{|c|c|c|c|}
\hline \bf Threshold & \bf TER & \bf \# sentences modified & \bf Precision \\\hline
0.8 & 24.29 & 130 & 0.20 \\\hline
0.6 & 23.99 & 103 & 0.18 \\\hline
0.4 & 23.66 & 70 & 0.18 \\\hline
\bf 0.2 & 23.46 & 50 & \bf 0.22 \\\hline
\end{tabular}
\end{center}
\caption{\label{tab:Filter2} Performance (TER score) of the APE-2-LM1 after pruning at various threshold values on dev set}
\end{table}

Table \ref{tab:Filter1} and Table \ref{tab:Filter2} show the performance after pruning the APE-1-LM1 and APE-2-LM1 systems respectively.
In Table \ref{tab:Filter1}, the TER scores for various threshold values remain close to each other, so to select the best threshold value, we based our decision on precision (Section \ref{sec:metric_auto}).
For APE-1, we selected the threshold value of 0.4, which showed the highest precision, namely \textbf{APE-1-LM1-Prun0.4}.
For APE-2, it is evident from the result in Table \ref{tab:Filter2} that the threshold value of 0.2 was the best in terms of both the TER score (reduction by 1.13 points) and precision (\textbf{APE-2-LM1-Prun0.2}).
These results suggest that our pruning technique had a larger impact on the APE-2 method than APE-1. This is motivated by the fact that the context-aware approach is affected by the data sparseness problem, resulting in many unreliable translation options that can be removed from the phrase table.

\paragraph{New Dense Features Design}
The final stage of our APE pipeline was the feature design.
When a translation system is trained using Moses, it generates a translation model consisting of default dense features, like phrase translation probability (direct and indirect) and lexical translation probability (direct and indirect). 
In the task of APE, where the source and target phrases are available in the same language, we can leverage this information to provide the decoder with useful insights.
In this direction, we designed four task-specific dense features to raise the ``awareness'' of the decoder.  
\begin{itemize}
\item \textbf{Similarity ($f1$):}\\
The $f1$ feature is quite similar to one proposed in the literature \cite{grundkiewicz2014amu}, which measured the similarity between the source and target phrases of the translation options.
The score for $f1$ is computed as follows:
\\
\centerline{$f1_{score}$ = $e^{1-ter(s,t)}$}
\\
where $ter$ measures the number of edit operations required to translate the source phrase $s$ into the target phrase $t$, and is computed using TER\cite{snover2006study}.

\item \textbf{Reliability ($f2.1$ and $f2.2$) :}\\
We allowed the model to learn the reliability of the translation option by providing it with the statistics of quality, in terms of HTER, for the parallel sentences used to learn that specific translation option.
The higher the quality, the higher the likelihood for learning reliable rules.

To achieve this, we extracted all parallel sentences and their HTER scores, which contained the source 
and the target phrases respectively (in the source and the target sentence).
The scores were then used to compute the following two features:\\
\textbf{Median ($f2.1$):}
The median of the HTER values of all the retrieved pairs.
\\
\textbf{Standard Deviation ($f2.2$):}
The standard deviation of the HTER values of all the retrieved pairs.

\item \textbf{Usefulness ($f3$):}
As discussed in Section \ref{sec:prun}, we used \textit{pos-impact} as a feature to measure the positive impact of a translation option over the training set.
The higher the positive impact, the higher its usefulness.
\end{itemize}
We studied the impact of individual features when applied one at a time and when used all together.

\begin{table}[h]
\begin{center}
\begin{tabular}{|c|c|c|c|}
\hline \bf Features & \bf TER & \bf \# of sentences modified & \bf Precision \\\hline
$f1$ & 23.87 & 81 & 0.16 \\\hline
\textbf{\textit{f2.1}}, \textbf{\textit{f2.2}} & 23.92 & 94 & \bf 0.19 \\\hline
$f3$ & 23.88 & 82 & 0.14 \\\hline
$f1$, $f2.1$, $f2.2$, $f3$ & 23.97 & 85 & 0.12 \\\hline
\end{tabular}
\end{center}
\caption{\label{tab:Feat1} Performance (TER score) of the APE-1-LM1-Prun0.4 for different features}
\end{table}

\begin{table}[h]
\begin{center}
\begin{tabular}{|c|c|c|c|}
\hline \bf Features & \bf TER & \bf Number of sentences modified & \bf Precision \\\hline
\textbf{\textit{f1}} & 23.50 & 52 & \bf 0.27 \\\hline
$f2.1$, $f2.2$ & 23.50 & 53 & 0.20 \\\hline
$f3.1$ & 23.52 & 59 & 0.22 \\\hline
$f1$, $f2.1$, $f2.2$, $f3.1$ & 23.52 & 54 & 0.19 \\\hline
\end{tabular}
\end{center}
\caption{\label{tab:Feat2} Performance (TER score) of the APE-2-LM1-Prun0.2 for different features on dev set}
\end{table}

Table \ref{tab:Feat1} and Table \ref{tab:Feat2} show the performance of various features for APE-1-LM1-prun0.4 and APE-2-LM1-Prun0.2 systems respectively.
For this dataset, we observed that the use of these features preserved the APE performance in terms of TER score.
A slight improvement was observed in terms of precision over both the APE systems, which indicated its contribution to improve the translation quality.

\subsection{Full-Fledged Systems}
\label{sec:finalSys}
From our previous experiments on the development set, we found the best configurations and used them as full-fledged systems for submission to the 2015 APE shared task.
We were allowed to submit two runs: primary and contrastive.
Our primary submission was the best system in Table \ref{tab:Feat2}, namely APE-2-LM1-Prun0.2-f1, and the contrastive system was the best system in Table \ref{tab:Feat1}, namely APE-1-LM1-Prun0.4-f2.1-f2.2.
According to the shared task evaluation report \cite{bojar-EtAl:2015:WMT}, the scores of our submitted systems were as shown in Table \ref{tab:eval}
\begin{table}[h]
\begin{center}
\begin{tabular}{|c|c|c|}
\hline \bf Systems & \bf Case Sensitive & \bf Case Insensitive \\\hline
Baseline (MT) & 22.91 & 22.22 \\\hline
APE Baseline \cite{simard2007statistical} & 23.83 & 23.13 \\\hline
Primary & 23.22 & 22.55 \\\hline
Contrastive & 23.64 & 22.94 \\\hline
\end{tabular}
\end{center}
\caption{\label{tab:eval} APE shared task evaluation score (TER)}
\end{table}

Although we could not beat the baseline (MT), we were able to obtain better results over the APE baseline \cite{simard2007statistical}.
This result indicates that our contributions led to an improvement in phrase-based APE technology.

\section{Summary}
In the first half of the chapter, we explored the potential of APE for domain-specific data in ideal conditions (quantity and quality of data) using the right equipment (state-of-the-art methods). The data comprised the same English sentences, machine-translated in six languages and post-edited by professional translators.
These data allowed us to  compare, for the first time different approaches in a fair setting.
The  two methods we analyzed helped us to measure  consistent improvements on all language pairs, as shown in TER reductions ranging from 7.3\% to 14.7\%.
This first study provided a starting point for future work. A promising direction to explore is  possible complementarity between the two methods, and the room for mutual improvement.  Now we just have a glimpse of the path (higher Oracle results, slight gains with a first  combination method), but positive preliminary results  confirm its existence.
This aspect of leveraging both methods is further explored in the next chapter, using a factored MT architecture

In the second half of the chapter, we explored the potential of APE for generic data.
Our contributions led to the submission at the pilot round of the APE shared task in 2015.
The shared task was challenging for many reasons, such as \textit{black-box MT, generic news domain data, and crowdsourced post-edits}.
Although we were unable to beat the MT baseline, we gained positive experiences through the shared task.
First, our primary APE system performed significantly better (0.61 TER reduction) than the official APE baseline \cite{simard2007statistical}.
Second, our novel dense feature (\textit{neg-impact}), which we used to prune the phrase table, showed significant improvement for the context-aware APE performance.
Third, other task-specific dense features, which measured the similarity and reliability of translation options, helped to improve the precision of our APE systems.

\chapter{Hybrid APE}
\label{chap:hybridAPE}
In this chapter, our focus moves away from pure phrase-based technology to a hybrid one that integrates a neural engine within the APE tool.
We show how these two paradigms can be merged to mutually benefit from each other.
Specifically, we used a factored machine translation framework to tie the phrase-based and neural technologies.
The former was used to develop a core translation model, while the latter contributed a language model.
Our study explored the same two APE variants discussed in the previous chapter, but with the novelty of introducing several target language models (POS-tag, class-based) built with different technologies to find the best configuration among them.
The resulting approach represents our submission to the WMT 2016 APE shared task.

\section{Introduction}
In the previous chapter, we discussed the pros and cons of the two APE variants we called ``monolingual'' and ``context-aware''.
Overall, the latter variant showed better performance; however, it faced two challenges.
The first was to preserve the source context results across many representations of the same \textit{mt} word, where each \textit{mt} word could be aligned to more than one \textit{source} word.
This causes an increase in data sparseness, which would eventually reduce the quality of word alignment between an (\textit{mt\#source}-\textit{pe}) pair.
The result would be a drop in the quality of PE rules.
Second, the joint representation (\textit{mt\#source}) itself may be affected by word alignment errors between (\textit{source}-\textit{mt}) pairs, which may mislead the learning of translation options.
Moreover, a technical problem with this representation occurs during tuning of the system. 
Because the input is a joint representation, it might be possible that the \textit{mt} in \textit{mt\#source} is a correct translation but the system might consider the whole word (\textit{mt\#source}) as OOV, thereby affecting the tuning process.
To address these issues and to better leverage the complementarity between the two alternative APE variants, we propose a more elegant approach, which combines the two variants in a factored translation model.

\section{Factored Phrase-based MT}
The factored MT model was proposed in \cite{koehn2007factored}. 
It enables the integration of additional annotations, called ``factors'', along with the source words.
These factors can be linguistic markup, automatically generated word classes, or any other information associated with each word. 
The factored translation framework supports two types of mapping: \textit{translation} and \textit{generation}.
Both mappings lead to the development of probabilistic models.
The former learns a bi-lingual model, such as the probability of a target word or stem given a source word or stem.
The latter learns a monolingual model of the target language, such as the probability of a target word given  a target stem and suffix.
This framework is especially useful to deal with data sparsity, where not all variations of a word are seen in the training data. 
In such situations, factored models can divide the main problem of translating words into sub-problems - such as first translating the stem and suffix of a source word and then generating the final target word from this partial translation information.
For the latter step, even vast target language monolingual data can be leveraged.
This mechanism of factored translation model made it suitable to explore in our APE task.

\section{Factored Phrase-based APE}
\label{sec:factoredAPE}
To build our factored APE systems, we pre-processed the training data to obtain the factored representation.
A fragment of our parallel corpus with factored representation is shown in Table \ref{tab:factor_eg}.
\begin{table*}[t]
\begin{center}
\small
\begin{tabular}{|>{\centering\arraybackslash}m{8cm}|>{\centering\arraybackslash}m{8cm}|}\hline
\multicolumn{2}{|c|}{\bf Parallel Corpus} \\\hline
\bf Source (\verb;mt_word|source_word;) & \bf Target (\verb;pe_word|pos-tag|class-id;) \\\hline
\begin{lstlisting}
Siehe|See Bemalen|Paint_on von|on 3D-Modellen|3D_models .|.
\end{lstlisting} & 
\begin{lstlisting}
Siehe|ADV|104 "|$(|373 Bemalen|NN|40 von|APPR|382 3D-Modellen|NN|137 .|$.|451 "|$(|373
\end{lstlisting}\\\hline
\begin{lstlisting}
Bildrate|Framerate des|of_the Videos|video MP4|MP4 .|.
\end{lstlisting} &
\begin{lstlisting}
Bildrate|NN|339 des|ART|407 MP4-Videos|NN|41 .|$.|451
\end{lstlisting}\\\hline 
\end{tabular}
\end{center}
\caption{\label{tab:factor_eg} Example of parallel corpus with factored representation.}
\end{table*}
The source side of the parallel corpus had 2 factors (\textit{mt\_word} and \textit{source\_word}, similar to the joint representation) and the target side contained 3 factors (\textit{pe\_word}, \textit{pos-tag}, and \textit{class-id}).\footnote{\textit{class-id} denotes a cluster of semantically related words.}
In this representation, we defined the following features:
\begin{itemize}

\item A word alignment mapping between \textit{mt\_word}{\textless}-{\textgreater}\textit{pe\_word}.
This helped to mitigate the problem of word alignment between (\textit{mt\#source}-\textit{pe}) in the \textit{context-aware} APE approach.

\item A translation mapping between \textit{mt\_word}{\textless}-{\textgreater}\textit{pe\_word} (\textit{monolingual} variant) and \textit{mt\_word{\textbar}source\_word}{\textless}-{\textgreater}\textit{pe\_word} (\textit{context-aware} variant). 
This allowed us to leverage both models during decoding.

\item A generation mapping between \textit{pe\_word}{\textless}-{\textgreater}\textit{pos-tag}, and \textit{pe\_word}{\textless}-{\textgreater}\textit{class-id}. 
This allowed us to improve the fluency of the translations by scoring them with part-of-speech tags as well as class-based language models.
\end{itemize}

\subsection{Factor Creation}
The factor on the source side of the parallel corpus was obtained after the approach to produce the joint representation for \textit{context-aware} APE. 
For the target side, we introduced two factors that measured fluency of translation at the syntactic level.
The first was POS-tag ($\sim$50 tags) obtained using the TreeTagger \cite{schmid1995improvements}.
The second was word-class ID ($\sim$500 classes), obtained using the \textit{mkcls}\footnote{\url{https://github.com/clab/mkcls}} tool, which clusters words based on bi-gram contextual similarity.
These factors were used to learn generation models ($P(pos\text{-}tag|pe)$ and $P(class\text{-}id|pe)$) to generate corresponding target factors for the test sentence, which were scored by their respective LMs during decoding.

\subsection{Neural Language Model}
As mentioned earlier, we were interested in studying the effect of different LMs on the performance of an APE system.
We built the n-gram LMs using the KenLM toolkit as described in Chapter 2 (Section \ref{sec:PBSMT}).
Here, we focus on developing LMs using the neural technology.
We selected the neural probabilistic language model (NPLM) toolkit for this task \cite{vaswani-EtAl:2013:EMNLP}.
NPLM is based on a feedforward neural network \cite{bengio2003neural} with a noise-contrastive estimation (NCE) technique \cite{gutmann2010noise}  that replaces the more common softmax layer with a binary classification task to speed up the training process.

\begin{figure}
\center
\includegraphics[width=10cm, height=8cm]{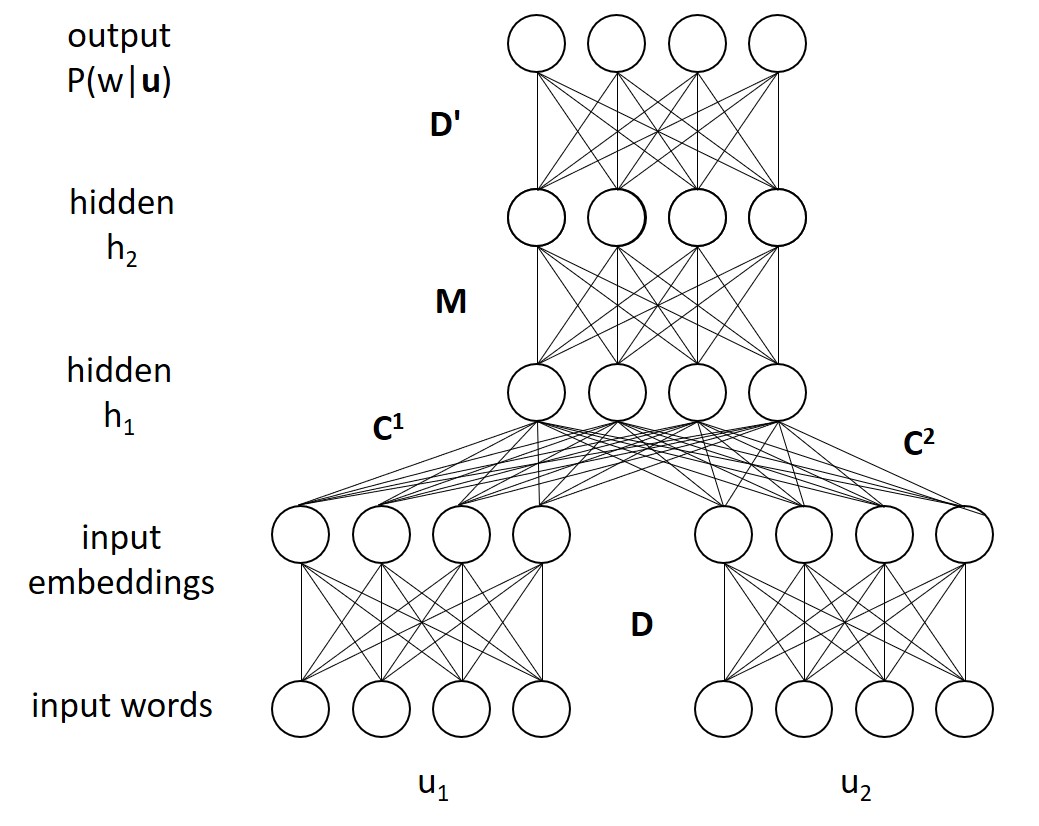}
\caption{\label{fig:nplm} Neural Probabilistic Language Model}
\end{figure}

Figure \ref{fig:nplm} depicts the feedforward neural LM. 
The input is a sequence of one-hot representations of words in context $\textbf{u}$, and the output is the probability of a target word $w$.
The probability is computed by the following softmax function which normalizes the score of a word $p(w|\textbf{u})$ by a constant $Z(\textbf{u})$.

{\centering
\begin{math}
P(w|\textbf{u}) = \frac{1}{Z(\textbf{u})}p(w|\textbf{u})
\end{math}

}
The normalization constant is the sum of the scores of all target words in the vocabulary.

{\centering
\begin{math}
Z(\textbf{u}) = \sum_{w^{'}}p(w^{'}|\textbf{u})
\end{math}

}
The score of a target word is computed by multiplying the vector of the second hidden layer $h_{2}$ with an output word embedding matrix $\textbf{D}'$ and then adding bias $\textbf{b}$ to it.

{\centering
\begin{math}
p(w|\textbf{u}) = exp(\textbf{D}'h_{2} + \textbf{b})
\end{math}

}
The vector of the second hidden layer is computed by multiplying the vector of the first hidden layer $h_{1}$ with a weight matrix $M$.

{\centering
\begin{math}
h_{2} = \o(\textbf{M}h_{1})
\end{math}

}
The vector of the first hidden layer is a function of the context matrix $\textbf{C}^{j}$ and the input-word embedding matrix $\textbf{D}$, to which a rectified linear unit is applied elementwise \cite{nair2010rectified}, as shown below. 

{\centering
\begin{math}
h_{1} = \phi \left ( \sum_{j=1}^{n-1}\textbf{C}^{j}\textbf{D}u_{j} \right )
\end{math}

}

To avoid the repeated summation involved in the normalization constant $Z(\textbf{u})$, noise-contrastive estimation was used, which considers $Z(\textbf{u})$ as another parameter of the network.
In NCE, the problem is shifted to a binary classification problem, where the network must predict whether a given sample belongs to the training data (Class=1) or noise (Class=0).
For this, $k$ noise samples $w_{i1},...,w_{ik}$ were added from a distribution $q(w)$ for each training sample $\textbf{u}_{i}w_{i}$ in the data.

{\centering
\begin{math}
P(C=1,w|\textbf{u}) = \frac{1}{1+k}.\frac{1}{Z(\textbf{u})}p(w|\textbf{u})
\end{math}

}

{\centering
\begin{math}
P(C=0,w|\textbf{u}) = \frac{k}{1+k}.q(w)
\end{math}

}
The objective was to maximize the conditional likelihood.
The parameters $\theta$ and $Z(\textbf{u})$ were updated through backpropagation with the following loss function.

{\centering
\begin{math}
L = \sum_{i=1}^{N}\left ( log P(C=1|\textbf{u}_{i}w_{i}) + \sum_{j=1}^{k}log P(C=0|\textbf{u}_{i}w_{ij}) \right )
\end{math}

}

\section{Experimental Setting}
\label{sec:data-n-eval}
In this section, we describe the experimental setting of our hybrid APE system submitted to the APE shared task at WMT 2016 \cite{bojar-EtAl:2016:WMT1}.
This system was built using the factored translation framework as described in Section \ref{sec:factoredAPE} to integrate classic phrase-based models along with a neural language model.\\\\
\textbf{Data:} The APE shared task training data \cite{bojar-EtAl:2016:WMT1} for English-German consisted of 12K triplets (\textit{src, mt, pe}).
We split the development data into 400 and 600 triplets, selected randomly, to tune and evaluate our APE systems.
We used \textit{pe} from the training data to build a 5-gram word-based language model using the KenLM toolkit, and 8-gram POS-tag and class-based language model using both KENLM and NPLM \cite{vaswani-EtAl:2013:EMNLP} toolkits.
To build the joint representation (\textit{mt\#src}) and to obtain source factors (\textit{mt{\textbar}src}), we used the word alignment model trained on the  (\textit{src}-\textit{mt}) pairs of the training data.
To develop the APE systems, we used the MOSES toolkit with the alignment heuristic set to \textit{grow-diag-final-and} and reordering to \textit{msd-bidirectional-fe}.
To build word alignment models, we used MGIZA++.
We used MERT to tune the feature weights with the aim of optimizing the TER score.
\\\\
\textbf{Baseline:}
We considered the MT system as a baseline, along with ``monolingual'' (APE-1) and ``context-aware'' (APE-2) variants.
The baseline results reported in Table \ref{tab:baseline-DA} (``baseline'' column) show that the naive \textit{monolingual} APE system already outperformed the MT baseline by 1.5 BLEU points.
However, the low precision of the APE systems indicates that they were prone to over-correction.
\begin{table}[h]
\begin{center}
\small

\begin{tabular}{|l|ccc|ccc|}\hline
 & \multicolumn{3}{c}{\bf Baseline} & \multicolumn{3}{|c|}{\bf Data Augmentation} \\
 & \bf TER & \bf BLEU & \bf Precision & \bf TER & \bf BLEU & \bf Precision \\\hline
MT system & 24.80 & 63.07 & - & - & - & -\\
APE-1 & 24.73  & 64.55 & 55.55 & 24.46  & 64.74 & 63.27 \\
APE-2 & 24.68 & 64.01 & 54.01 & \bf 24.08$\dagger$ & \bf 64.88$\dagger$ & \bf 70.41  \\\hline
\end{tabular}
\end{center}
\caption{\label{tab:baseline-DA} Performance of the APE systems on development set (``$\dagger$'' indicates statistically significant differences wrt. Baseline with p\textless 0.05).}
\end{table}
\\
\section{Results}
\textbf{Data Augmentation:}
To mitigate the problem of over-correction, the APE system should learn to preserve the correct parts of the MT segment.
To this aim, we augmented the parallel corpus with the PEs (12K) on both source and target side, so that the system could learn to identity patterns which helped to preserve correct MT words.
To summarize, our final training corpus consisted of 12K \textit{mt-pe} or \textit{mt\#src-pe} pairs, to learn PE rules, and an additional 12K \textit{pe-pe} or \textit{pe\#src-pe} pairs, to preserve correct MT words.
Replicating the baseline APE systems with the data augmentation technique yielded significant improvements in all the evaluation metrics, as reported in Table \ref{tab:baseline-DA} (``data augmentation'' column). 
Therefore, we used the augmented training data in all further experiments.
Among the two variants, we noticed that the APE-2 obtained maximum benefit with an absolute precision improvement of 16.4\%  (an increase from 54.01\% to 70.41\%).
\\\\
\textbf{Factored APE models:}
Both the APE variants had their own strengths and weaknesses, as discussed in the previous chapter. 
To leverage their complementarity, we used a factored translation approach (described in Section \ref{sec:factoredAPE}).
Before combining the two variants, we replicated the \textit{context-aware} variant - which was the best of the two - in the factored architecture, along with the integration of different target LMs.
We studied the effect on the performance of the APE system when using a 8-gram POS-tag and class-based LMs built with both statistical and neural technologies, while keeping a standard n-gram word-based model in all the setups.
The results are reported in Table \ref{tab:LM-performance}.
It is evident that the neural LM performed better than the n-gram statistical models.
The combination of both POS-tag and class-based neural LMs provided slightly better precision than individual neural LMs.

\begin{table*}[h]
\begin{center}
\small

\resizebox{\textwidth}{!}{%
\begin{tabular}{|c||c|c|c||c|c|c||c|c|c|}\hline
\bf & \multicolumn{3}{ |c|| }{\bf  POS-tag LM} & \multicolumn{3}{|c||}{\bf Class-based LM} & \multicolumn{3}{|c|}{\bf POS-tag \& Class-based LM} \\\hline\hline
\bf Approach & TER & BLEU & Precision & TER & BLEU & Precision & TER & BLEU & Precision \\\hline
N-gram & 24.20  & 64.29 & 63.88 & 24.28 & 65.08 & 67.27 & 24.22 & 65.12  & 70.25 \\\hline
Neural & 24.06 & 65.27 & 71.85 & 24.07 & 65.04 & 68.92 & \bf 24.07 & \bf 65.31 & \bf 72.72 \\\hline
\end{tabular}
}
\end{center}
\caption{\label{tab:LM-performance} Performance of the Factored APE-2 for various LMs (statistical word-based LM is present in all the experiments by default).}
\end{table*}

We therefore decided to use the neural POS-tag and the class-based LMs along with an n-gram word-based LM, for both variants (\textit{monolingual} and \textit{context-aware}) in the factored architecture.
The translation models of both variants were used together during decoding, with the help of the multiple decoding feature available in the MOSES toolkit.
The results of this factored APE system for various tuning strategies are shown in Table \ref{tab:factored-combine}.
The tuning strategies we tested were \textit{i)} MERT to optimize TER; \textit{ii)} MERT to optimize BLEU; and \textit{iii)} MIRA to optimize BLEU.
The TER score was almost the same for different tuning strategies, but a slight improvement in BLEU score was observed with MIRA, therefore, we select it for further exploration. 
\begin{table}[h]
\begin{center}
\small

\begin{tabular}{|l|c|c|c|}\hline
\bf Optimization & \bf TER & \bf BLEU & \bf Precision \\\hline \hline
MERT-TER & 24.03 & 65.03 & 69.71 \\ \hline
MERT-BLEU & 24.07 & 65.47 & 65.67 \\\hline
\bf \parbox[t]{3cm}{MIRA-BLEU} & \bf 24.04  & \bf 65.56 & \bf 67.47 \\\hline
\end{tabular}
\end{center}
\caption{\label{tab:factored-combine} Performance of the combined factored model for various tuning configurations.}
\end{table}
\\
\textbf{Factored APE model with quality estimation:}
\label{sec:qe}
The previous experiment (MIRA-BLEU, Table \ref{tab:factored-combine}) showed that most of the APE segments were better than MT (67.47\% precision).
However, a few MT segments were deteriorated by APE.
A mechanism to select the best translation among MT and APE could provide an optimal solution, yielding 100\% precision in the final output.
To this end, we built a sentence-level quality estimation (QE)  \cite{Mehdad2012a,Turchi2014,cdesouza-EtAl:2015:ACL-IJCNLP} that would assign a quality score to a translation.\footnote{QE is explained more in depth in Chapter \ref{chap:QEandAPE}.}
To train the QE model, we first extracted 79 system-independent features that comprised three aspects of the QE problem: \textit{i)} fluency (e.g. language model perplexity of the whole translated sentence); \textit{ii)} complexity (e.g. average token length of the source sentence); and \textit{iii)} adequacy (e.g. ratio between the number of nouns in the source and translation sentences). 
These features were obtained with the QuEst feature extractor implementation \cite{Specia2013}.
We used the information to train a regression model to predict the actual PE effort, as measured by the TER between the MT-generated translation or the factored APE output and  PE version. The regression model was trained using the extremely randomized trees \cite{Geurts2006} implementation of Scikit-learn library \cite{Pedregosa2011}. 
This method had yielded competitive results in sentence-level QE shared-tasks in previous years \cite{Souza2013,Souza2014a}. 
To select the final translation, we checked whether the predicted score of MT output was lower\footnote{Lower is better since we are predicting TER scores} than the predicted score of the APE output by at least \textit{k} TER points (threshold).
We experimented with different threshold values, as reported in Table \ref{tab:qe}.
Using QE with a threshold of 5 performed slightly better than other competitors in terms of both TER and BLEU metrics.
\begin{table}[h]
\begin{center}
\small

\begin{tabular}{|p{2.5cm}|c|c|c|}\hline
\bf Threshold & \bf TER & \bf BLEU & \bf Precision \\\hline
1 & 24.18 & 65.09 & 72.13 \\ \hline
2 & 24.15 & 65.34 & 70.88 \\\hline
3 & 24.09 & 65.51 & 68.15 \\\hline
4 & 24.02 & 65.59 & 68.94 \\\hline
\bf 5 (Primary) & \bf 23.99 & \bf 65.65 & \bf 67.83 \\\hline
6 & 24.01 & 65.64 & 67.98 \\\hline \hline
Contrastive (w/o QE) & 24.04  & 65.56 & 67.47 \\\hline 
Baseline (MT) & 24.80 & 63.07 & - \\\hline
\end{tabular}
\end{center}
\caption{\label{tab:qe} Performance of the APE system with quality estimation for various thresholds.}
\end{table}
\\\\
\textbf{Full-Fledged System}
\label{sec:final-sub}
The shared task evaluation used 2,000 unseen samples consisting of (\textit{source}-\textit{mt}) pairs from the same domain of the training data.
Our primary submission was a factored APE system that \textit{i)} was trained with a data augmentation technique; \textit{ii)}  leveraged the two statistical phrase-based variants (\textit{monolingual}, and \textit{context-aware}); \textit{iii)} used a neural POS-tag and class-based LM along with the statistical word-based LM; and \textit{iv)} used a quality estimation model.
Our contrastive submission was similar to the primary one but lacked the QE model.
According to the official results of the shared task (Table \ref{tab:task-results}) \cite{bojar-EtAl:2016:WMT1}, both of our submissions achieved similar performances, with a minimal difference in TER.
They provided significant improvements of 3.31\% for TER and 4.25\% for BLEU (relative) compared with the baseline MT system. 
We also observed that the use of QE in our primary submission did not yield the expected improvements.
Probably, the sentence-level QE performance was insufficient to accurately distinguish which of the two translations was better.
We explore this aspect of integrating QE and APE in-depth in Chapter \ref{chap:QEandAPE}.

\begin{table}[h]
\begin{center}
\small

\begin{tabular}{|l|c|c|}\hline
\bf & \bf TER & \bf BLEU \\\hline
Baseline (MT) & 24.76 & 62.11 \\ \hline
Baseline (APE) & 24.64 & 63.47 \\\hline
\bf Primary & \bf 23.94$\dagger$ & \bf 64.75$\dagger$ \\\hline
\bf Contrastive & \bf 23.92$\dagger$ & \bf 64.75$\dagger$ \\\hline
\end{tabular}
\end{center}
\caption{\label{tab:task-results} Results of our submission to the 2016 WMT APE shared task (``$\dagger$'' indicates statistically significant differences wrt. Baseline (MT) with p\textless 0.05). }
\end{table}

\section{Summary}
\label{sec:conclusion}
In this chapter, we presented a hybrid APE architecture that benefitted from both phrase-based and neural technologies.
The overall system was built within a factored MT framework that allowed us to combine \textit{i)} the two technologies; \textit{ii)} the two APE variants; and \textit{iii)} multiple LMs.
From our experiments on LMs, we learned that \textit{i)} using both POS-tag and class-based LM together was better than using either one alone; \textit{ii)} building the LMs with a neural-based approach was better than the traditional n-gram based solution; and \textit{iii)} the best LM combination achieved a 0.4 BLEU point improvement over the system that did not use these LMs.
The performance of our primary and contrastive submissions to the shared task were similar, with a significant improvement of 3.31\% TER and 4.25\% BLEU points (relative)  over the baseline MT system.
However, adding a layer of QE to our primary submission did not yield the expected improvement. We investigate this aspect of using QE with APE in-depth in Chapter \ref{chap:QEandAPE}, after discussing end-to-end neural APE solutions in the next chapter (Chapter \ref{chap:neuralAPE}).

\chapter{End-to-End Neural APE}
\label{chap:neuralAPE}
This chapter describes a major shift in the APE paradigm: moving from pure phrase-based and hybrid solutions, we present an end-to-end deep learning architecture for APE.
We show how to jointly model the two APE variants in this new framework.
We also discuss how to train an efficient model with limited amount of domain-specific data.
The resulting multi-source neural APE approach represents our submission to the WMT 2017 APE shared task, where it achieved the best performance in both human and automatic evaluation.

\section{Introduction}
In the previous chapters, we showed that the source context was crucial to correct errors in the MT segment.
Exploiting source information as an additional input can help the system to disambiguate error-correction rules. For example, the German phrase \textit{``mein Haus''} (EN: \textit{``my house''})  looks correct, but if the source phrase was \textit{``my home''} then the correct translation would be \textit{``mein Zuhaus''}. 
In this case, an APE system ignoring the source would have left the sub-optimal MT output untouched.
We have already discussed the effectiveness of jointly learning from source and MT output through the ``context-aware'' variant.
By integrating this notion in a neural approach to the problem, 
we present our multi-source neural automatic post-editing (NAPE) architecture.
Our architecture is an extension to an existing NMT implementation, discussed in Chapter \ref{chap:MT}, Section \ref{sec:NMT}.

\section{Multi-source Neural APE}
\label{sec:multisource}
To our knowledge, the first multi-source neural APE approach was proposed by \cite{libovicky-EtAl:2016:WMT}.
According to the authors, their neural network was unable to perform minimum edits required to correct the MT segment.
Rather, it paraphrased the input, which resulted in a high chance of performing unnecessary corrections that would be penalized by a reference-based evaluation metric against human PEs.
To overcome this problem, they represented the target as a minimum-length sequence of edit operation needed to turn the MT sentence into the reference PE.
Our multi-source APE implementation, which was built on top of the neural MT architecture ($\S$\ref{sec:NMT}), is similar to theirs \cite{libovicky-EtAl:2016:WMT}. However, our model extends theirs with a context dropout, and considers the target as a sequence of words rather than a minimum-length sequence.
We extended the existing architecture ($\S$\ref{sec:NMT}) to have two encoders, one for \textit{src} and the other for \textit{mt}.
Each encoder has its own attention layer that is used to compute the weighted context.
The \textit{src} and the \textit{mt} contexts ($c_{t}^{src}$ and $c_{t}^{mt}$ respectively) are then passed to a merging layer to obtain the final context ($c_{t-merge}$).
The merging layer basically concatenates both contexts and applies a linear transformation. Thus the final context possesses necessary information from both the inputs, as shown below:
\begin{equation}
c_{t-merge} = [c_{t}^{\textit{src}}; c_{t}^{\textit{mt}}] * W_{ct} + b_{ct}
\end{equation}
where, $W_{ct}, b_{ct}$ are respectively the weight and the bias of the merging layer.
The final context ($c_{t-merge}$) is used by the decoder to compute target word probabilities:
\begin{equation}
p_{\Theta}(y_{t}|y_{<t},x) = g(\dot{y}_{t-1},s_{t},c_{t-merge}) 
\end{equation}
\\
Context Dropout:
Dropout was proposed \cite{hinton2012improving} as a regularization technique for deep neural networks to avoid over-fitting.
The aim is to randomly drop some units - and their incoming and outgoing connections - from the neural network to prevent co-adaption on the training data.
This method has been shown to be highly effective for a wide range of supervised learning tasks in vision, speech recognition, document classification and computational biology \cite{srivastava2014dropout}.
When applying dropout with a recurrent neural network, using the same dropout mask at each timestep is better than ad hoc techniques where different dropouts are sampled at each time step \cite{gal2016theoretically}.
We added dropout for the source context at different layers of the network.
\begin{itemize}
    \item To compute the attention score, we applied a shared dropout to the hidden state of both encoders;
    \item  To compute the final hidden state of the decoder  we applied a dropout to the merged context of the encoders ($c_{t-merge}$).   
\end{itemize}
We observed that the use of source context dropout helped the model to avoid overfitting. It also allowed for more stable performance on the validation set when the model converged.

\subsection{Experiments}
\label{sec:exp}
In this section, we summarize how our  models were trained, tuned and combined to produce the multi-source APE submissions at the WMT 2017  APE  shared  task (for more details about the task and data please refer to Section \ref{sec:aperound2017}).

\paragraph{Data:}
For EN-DE, we used $\sim$4M artificial training data from literature \cite{junczysdowmunt-grundkiewicz:2016:WMT}. The data were created by a round-trip translation procedure to train generic models that were later fine-tuned with $\sim$500K artificial\footnote{\url{https://github.com/amunmt/amunmt/wiki/AmuNMT-for-Automatic-Post-Editing}} and 23K real PE training data (replicated 20 times in order to avoid bias towards artificial data) collected from the 2017 APE shared task and its previous rounds \cite{bojar-EtAl:2016:WMT1,bojar-EtAl:2017:WMT1}.\footnote{\url{http://www.statmt.org/wmt17/ape-task.html}} 
The development set released in the 2016 APE shared task was used to evaluate and compare different models' performance.
All the data were segmented using the byte pair encoding (BPE) technique to obtain sub-word units, to avoid the problem of out-of-vocabulary words\cite{sennrich-haddow-birch:2016:P16-12}.

For DE-EN, we created PE training data by a round-trip translation, using the sub-set of parallel data released in the medical task at WMT'14 \cite{bojar-EtAl:2014:W14-33}. 
The parallel data were used to build a phrase-based MT system (PBMT) for both EN-DE and DE-EN language directions. 
The monolingual English data (considered as \textit{pe}) were first translated into German (considered as \textit{source}) using the EN-DE PBMT system, and then back-translated into English (considered as \textit{mt}) using the DE-EN PBMT system. 
The parallel and monolingual data consisted of $\sim$2M segments.
To train the APE systems we concatenated the round-trip translated data, the parallel data (for which we considered the reference as the MT output), and the shared task training data (25K triplets). The data were replicated 160 times to provided 50\% real and 50\% artificial data to avoid any bias towards artificial data.
All the data were segmented in sub-word units, similar to the \textit{EN-DE} direction, and the systems were evaluated on the development set released for the 2017 APE shared task.

\paragraph{Hyper parameters:}
The hyper parameters of all the systems in both language directions were the same.
The vocabulary was created by selecting the 50K most frequent sub-words.
Word embedding and GRU hidden state size were set to 1024.
Network parameters were optimized with Adagrad \cite{adagrad}, with a learning rate of 0.01, following a study \cite{fbk-iwslt16} that empirically showed that Adagrad had a faster convergence rate and better performance than Adadelta \cite{zeiler2012adadelta}.
Source and target dropout was set to 10\%, 
whereas encoder and decoder hidden states, weighted source context, and embedding dropout was set to 20\% \cite{sennrich-haddow-birch:2016:WMT}.
After each epoch, the training data were shuffled and the batches were created after sorting 2000 samples to speed up the training.
The batch size was set to 100 samples, with a maximum sentence length of 50 sub-words.

\paragraph{Models:}
For both language directions, we trained four different networks to capture information that could be leveraged together through ensemble techniques.
The results of the single best model for EN-DE and DE-EN from each network type are reported in Tables \ref{tab:results-EN-DE} and \ref{tab:results-DE-EN}, respectively.
The performance trends among different networks were similar for both the language directions.
However, the variation was less visible for DE-EN, as the room for improvement was far less due to higher MT quality (15.58 TER and 79.46 BLEU scores).
We base our discussion for each model below on the results achieved with the development data for the EN-DE direction. The performance variations among different networks were more visible for this condition.
\begin{table}[h]
\begin{center}
\begin{tabular}{l|c|c|p{0.9cm}}\hline\hline
\bf Systems & \bf TER & \bf BLEU & \bf Prec. (\%) \\\hline
MT\_Baseline & 24.81 & 62.92 & - \\\hline
SRC\_PE & 26.66 & 61.91 & 49.07 \\
MT\_PE & 21.57 & 69.09 & 72.01 \\
MT+SRC\_PE & 19.77 & 70.72 & 78.22 \\
MT+SRC\_PE\_TSL & 20.07 & 70.52 & 78.77 \\\hline
Ens8 & 19.26 & 71.63 & 78.50 \\
Ens8+Re-rank-A & \bf 19.22$\dagger$ & \bf 71.89$\dagger$ & \bf 78.84 \\
Ens8+Re-rank-AB & 19.35 & 70.94 & 78.07 \\\hline\hline
\end{tabular}
\end{center}
\caption{\label{tab:results-EN-DE}Performance of the APE systems on dev. 2016 (EN-DE) (``$\dagger$'' indicates statistically significant differences wrt. MT\_Baseline with p\textless 0.05).}
\end{table}

\begin{table}[h]
\begin{center}
\begin{tabular}{l|c|c|p{1.1cm}}\hline\hline
\bf Systems & \bf TER & \bf BLEU & \bf Precision (\%) \\\hline
MT\_Baseline & 15.58 & 79.46 & - \\\hline
SRC\_PE & 28.50 & 58.17 & 20.22 \\
MT\_PE & 15.97 & 78.43 & 36.29 \\
MT+SRC\_PE & 15.61 & 78.59 & 44.67 \\
MT+SRC\_PE\_TSL & 15.89 & 78.48 & 42.58 \\\hline
Ens8 & 15.14 & 79.41 & 54.18 \\
Ens8+Re-rank-A & \bf 15.04$\dagger$ & \bf 80.00$\dagger$ & \bf 68.86 \\
\hline\hline
\end{tabular}
\end{center}
\caption{\label{tab:results-DE-EN}Performance of the APE systems on dev. 2017 (\textit{DE-EN}) (``$\dagger$'' indicates statistically significant differences wrt. MT\_Baseline with p\textless 0.05).}
\end{table}
\paragraph{SRC\_PE} 
This system was similar to a NMT system used for bilingual translation from a source language to a target language.
The parallel corpus consisted of source text and PEs of the MT segments.
The performance of this system was below the MT baseline, indicating that learning from the source text alone was insufficient to improve the translation quality.
This system probably generates alternative and potentially correct
translations, which diverge from the MT output and are thus penalized by automatic evaluation metrics that use human PEs as references.
This idea was confirmed because when we used the reference test set for evaluation,\footnote{\url{http://hdl.handle.net/11234/1-2334}} the APE system outperformed the MT baseline by +4.2 BLEU points (47.97 vs 43.79 for BLEU score).

\paragraph{MT\_PE}
This model represented the ``monolingual'' APE variant discussed in the previous chapters.
Source and target languages were the same, and the goal was to translate raw MT segments into their corrected version.
The results in Table \ref{tab:results-EN-DE} show that learning from MT was better than learning from the corresponding source sentences (-3.2 TER and +6.0 BLEU points higher than the MT baseline).
Although the performance gain was substantial, it did not indicate whether all the MT segments were improved.
To better understand this aspect,  we used the precision metric \ref{sec:metric_auto}.
A precision rate of 72\% for this system indicated that the most of the MT segments that were modified resulted in a better translation quality.

\paragraph{MT+SRC\_PE}
This model represented the ``context-aware'' APE variant discussed in the previous chapters.
However, we did not use the joint representation here (\textit{mt\#source}). Instead, the source and MT were encoded independently by their respective encoders to form a multi-source neural sequence-to-sequence model, 
as described in $\S$\ref{sec:multisource}.
Our multi-source neural architecture clearly outperformed the strong monolingual single-source model (-1.8 TER and +1.6 BLEU).
The improvement was also evident in precision (+8.2\%), which indicates that the source segment was useful for disambiguating whether the MT word should be corrected or left untouched, thus helping to mitigate the problem of over-correction.

\paragraph{MT+SRC\_PE\_TSL}
The low TER score of the MT baseline (24.8 and 15.5, respectively, for \textit{EN-DE} and \textit{DE-EN}) 
indicated that most of the MT words were correct.
To induce a conservative approach, or, in other words, to induce the APE system to preserve the correct MT words, we used a task-specific loss (TSL) function that considered the attention score of the MT words before computing the target word probabilities.
The attention scores can act as a reward to the target words that are present in the MT segment.
To this end, first we added the attention scores from the \textit{mt} encoder to the respective target words in the softmax layer.
Then, we applied softmax to obtain the target word probabilities.
This procedure is formally written as follows:
\begin{equation}
p_{\Theta}(y_{t}|y_{<t}, X^{src}, X^{mt}) = \frac{e_{dec}^{y_{t}}+ \sum e_{enc}^{y_{t}}}{\sum_{y'} (e_{dec}^{y'}+\sum e_{enc}^{y'})}
\end{equation}
where, $e_{dec}^{y}$ and $e_{enc}^{y}$ are, respectively,  the scores for the target word computed by the decoder layer and the attention layer of the \textit{mt} encoder ($e_{enc}^{y}$ = 0 if $y \notin MT$). 
Because a target word could occur many times in the MT segment, we summed the scores of all occurrences. If a target word was not present in the MT segment, the score was 0. 
The contribution of this model was visible in terms of precision for EN-DE experiments (Table \ref{tab:results-EN-DE}). 

\paragraph{Ensemble (Ens8)}
To leverage all the network architectures discussed above, we ensembled the two best models from each of them.
The networks were highly diverse in terms of information learned from the input representation, and we observed that weighting all the models equally did not yield an improvement over the single system.
Therefore, we generated 50-best hypotheses from the ensemble system and then tuned the model weights with Batch-MIRA \cite{cherry2012batch} on the development set to maximize the BLEU score.
We observed that, after three cycles of decoding and tuning, the performance converged.
The weighted ensemble of eight models further improved the translation quality (-0.8 TER and +1.1 BLEU) relative to the best single multi-source model (MT+SRC\_PE).

\paragraph{Re-ranking}
Following the improvements obtained by re-ranking n-best hypotheses \cite{pal-EtAl:2017:EACLshort}, we used a re-ranker in our final submissions.
To train the re-ranker, we used shallow features that can easily be extracted on-the-fly.
It captures different types of edit operations performed by an APE system on the MT output.
These features include: number of insertions, deletions, substitutions, and shifts, and the length ratio between the MT segment and each APE hypothesis (computed using TER).
In addition, to avoid over-correction by rewarding hypotheses that were closer to the MT segment, we computed the precision and recall of the APE hypotheses.
Precision is the percentage of words generated by the APE system that are present in the MT segment, and recall is the percentage of words in the MT segment that are generated by the APE system.
The feature weights were optimized with Batch-MIRA on the development set to maximize the BLEU score.
Re-ranking with these features yielded further improvements over the ensemble system.
To compare our models with the best neural-based system of the 2016 APE task, we evaluated our system on the 2016 APE test set.
The results of this evaluation are reported in Table \ref{tab:results-EN-DE-2016}.
We observed that this system achieved significant improvement over the MT baseline (-5.4 TER and 8.7 BLEU points) as well as the 2016 test set.
\begin{table}[h]
\begin{center}
\begin{tabular}{l|c|c}\hline\hline
\bf Systems & \bf TER & \bf BLEU \\\hline
MT\_Baseline & 24.76 & 62.11 \\
APE\_Baseline & 24.64 & 63.47 \\
Best APE (2016) & 21.52 & 67.65 \\ 
Ens8+Re-rank-A & \bf 19.32$\dagger$ & \bf 70.88$\dagger$ \\
\hline\hline
\end{tabular}
\end{center}
\caption{\label{tab:results-EN-DE-2016}Performance of the APE systems on the 2016 WMT test set (EN-DE) (``$\dagger$'' indicates statistically significant differences wrt. MT\_Baseline with p\textless 0.05).}
\end{table}

\subsection{Results on Test Data}
\label{sec:submission}
The shared task evaluation was performed on 2,000 unseen samples consisting of \textit{src} and \textit{mt} pairs obtained from the same domain of the training data.
Our primary submission was Ens8+Re-rank-A (see Tables \ref{tab:results-EN-DE} and \ref{tab:results-DE-EN}), which was a weighted ensemble of eight neural APE models (two best models from SRC\_PE, MT\_PE, MT+SRC\_PE, and MT+SRC\_PE\_TSL).
As a contrastive submission, we wanted to evaluate the performance of a simpler system with a higher throughput.
Therefore, we selected a single best multi-source model (MT+SRC\_PE) with a re-ranker that was based only on edit-distance features (labelled "Contrastive-A" in Table \ref{tab:officialresults}).
According to the shared task results, as reported in Table \ref{tab:officialresults}, our primary and contrastive submissions achieved significant improvement over the MT baseline for both language directions.
It is interesting to note that our contrastive-A submission, which was a simpler version than the full-fledged system, performed almost similarly to our primary submission for \textit{DE-EN} but slightly worse (+0.7 TER points) for \textit{EN-DE}.
\begin{table}[h]
\begin{center}
\begin{tabular}{l|c|c|c|c}\hline\hline
\multirow{2}{*}{\bf Systems} & \multicolumn{2}{c|}{EN-DE} & \multicolumn{2}{c}{DE-EN}\\\cline{2-5}
& \bf TER & \bf BLEU & \bf TER & \bf BLEU \\\hline
Baseline (MT) & 24.48 & 62.49 & 15.55 & 79.54 \\
Baseline (APE) & 24.69 & 62.97 & 15.74 & 79.28 \\\hline
Primary & \bf 19.6$\dagger$ & \bf 70.07$\dagger$ & \bf 15.29$\dagger$ & \bf 79.82$\dagger$ \\
Contrastive-A & 20.3$\dagger$ & 69.11$\dagger$ & 15.31 & 79.64 \\
\hline\hline
\end{tabular}
\end{center}
\caption{\label{tab:officialresults}Official results on 2017 test set (``$\dagger$'' indicates statistically significant differences wrt. Baseline (MT) with p\textless 0.05).}
\end{table}

\section{Summary}
In this chapter, we investigated the latest APE paradigm shift from phrase-based to neural nets.
We evaluated both technologies using the same dataset. The first was the official APE baseline implementation of a phrase-based approach; the second was our winning system based on deep neural nets, which gained more than 7 BLEU points over the baseline for the EN-DE language pair.
Our study confirmed that the state-of-the-art system is an end-to-end deep learning based solution to the task.
We showed how to jointly model both source and MT using a multi-source neural architecture, and how the shallow re-ranker features contributed to the final translation quality.
Although quite powerful, neural nets still suffer from the problem of over-correction.
This is more visible when the underlying MT quality is good, leaving little to no room for improvement, as confirmed by our experiments on the DE-EN language pair.
We extended our study to mitigate this problem by leveraging external resources, as discussed in the next chapter.

\chapter{Combining APE and QE for MT}
\label{chap:QEandAPE}
Previous chapters have discussed the effectiveness of APE systems implemented under different paradigms.
Moving from classical phrase-based methods to recent neural-based solutions provided a steady progress in state-of-the-art results.
However, a common problem faced by each of these technologies is over-correction.
We addressed this issue inherently in the system architecture, either by proposing task-specific dense features in the phrase-based paradigm (Chapter \ref{chap:phrase-basedAPE}) or by defining a new task-specific loss function in neural nets (Chapter \ref{chap:neuralAPE}).
In this chapter, we explicitly address the problem by leveraging external knowledge in the form of quality judgments of machine-translated text.
This is achieved by a novel ``guide'' mechanism that enhances the neural-based APE decoder with the ability to integrate external knowledge, without needing to re-train or fine-tune existing models. 
In the latter part of the chapter, we investigate different strategies for combining quality estimation and automatic post-editing to improve translation quality.  
The joint contribution of the two tasks was analyzed in various settings.
Here, QE served in one of the following roles:  \textit{i)} an \textit{activator} of APE corrections; \textit{ii)} a \textit{guidance} for APE corrections; or \textit{iii)} a \textit{selector} of the final output to be returned to the user. 

\section{Introduction}
\label{sec:intro}
In recent years, the steady progress of MT technology has motivated research on various ancillary tasks.
The wide adoption  of MT, especially  in the translation industry, has raised new challenges.
These are not only related to model training,  optimization, adaptation and evaluation, but also to aspects that are external to the core translation approach, such as quality estimation and automatic post-editing.

Both QE and APE have been successfully explored as standalone tasks in previous work, especially in the established WMT framework. In six editions of the WMT QE shared task (2012-2017), the MT quality prediction problem was formulated in different ways (\textit{e.g.} ranking, scoring) and  attacked at different levels of granularity (sentence, phrase, and word-levels). Constant state-of-the-art advancements are rendering QE a more appealing technology - for instance, to enhance the productivity of human translators who operate with computer-assisted translation tools \cite{Bentivogli:2016:IEEE}.
Three rounds of the  APE shared task at WMT (2015-2017) followed a similar trend, with improvements that reflected steady progress in the underlying technology developed by participants (Section \ref{sec:apefindings} covers more detail of each round).
Despite the growing interest in the two tasks and the fact that the proposed evaluation exercises shared the same training or test data, previous research on the two topics has mainly followed independent paths. 
The potential usefulness of leveraging both technologies together to achieve better MT has barely been explored.
No systematic analysis of possible combination strategies had been reported until this work. 
In this chapter, we investigate how QE and APE can be jointly deployed to boost the overall quality of the output produced by an MT system, without intervention on the actual MT technology.

\section{Quality Estimation}
\label{subsec:QE}
MT QE refers to the task of predicting the quality of MT text at run-time, without relying on hand-crafted reference translations \cite{Specia2010}. 
The possible uses of MT QE include: \textit{i)} deciding whether a given translation (or portions of it) is good enough for publishing \textit{as-is} or needs PE by professional translators (\textit{e.g.} in a computer-assisted translation environment); \textit{ii)} informing readers in the target language about the reliability of a translation; and \textit{iii)} selecting the best translation among options from many MT systems. 

QE is usually cast as a supervised learning task, in which systems trained on (\textit{input}, \textit{output}, \textit{quality\_label}) triplets must predict a quality label for unseen (\textit{input}, \textit{output}) test instances. In previous works, this problem was addressed from different perspectives - which in recent years have reflected the formulation of various sub-tasks proposed within the WMT shared evaluation framework.
The main differences concern the granularity and type of predictions as well as the underlying learning paradigm. The granularity of QE predictions  ranges from the document level to the sentence, phrase or word level. Relevant to this chapter are the sentence and word levels, the former being the most widely studied type of QE.

In sentence-level QE, the required predictions can be any of the following:
\textit{i)} PE effort estimates (\textit{e.g.} the expected number of editing operations required to correct the MT output);
\textit{ii)} PE time estimates (\textit{e.g.} the time required to make an automatic translation acceptable);
\textit{iii)} ranking of many translation options;
\textit{iv)} binary (``good''/``bad'') or Likert-scale scores (\textit{e.g.} scores ranging from 1 to 5, indicating the overall translation quality as perceived by a human). 
Depending on the type of prediction required, the proposed supervised learning approaches range from classification to regression and ranking. Together with the batch learning solutions that characterize most   of the proposed approaches, recent studies have explored the application of online and multitask learning methods 
\cite{Turchi2014,cdesouza-EtAl:2015:ACL-IJCNLP}.
The studied methods have targeted flexibility and robustness to domain differences between training and test data.

In word-level QE, the required predictions can either be binary ``good''/``bad'' labels for each MT output token, or  finer-grained multi-class labels that indicate the type of  error occurring in a specific word.  The problem, initially approached as a sequence labelling task \cite{luong-lecouteux-besacier:2013:WMT}, has  been successfully addressed with neural solutions, which now represent the state-of-the-art technology \cite{Souza2014a,kreutzer-schamoni-riezler:2015:WMT,martins-EtAl:2016:WMT,kim-lee-na:2017:WMT}.

\section{Integrating External Knowledge in NMT Decoding}
The need to enforce fixed translations of certain source words is a well-known problem in MT.
For instance, this is an issue in application scenarios in which the translation process must comply with specific terminology or style guides.
In addition to MT, the same need arises for APE, where one might want the APE engine to modify only the erroneous MT words, leaving the correct ones untouched.
Considering both tasks, we proposed a single solution that was compatible with both.
However, in this chapter we focus on the APE use-case.
Details about the MT task appear in our published paper \cite{chatterjee-EtAl:2017:WMT1}.
This ``constraint decoding'' problem has been addressed in the case of phrase-based technology, which explicitly manipulates symbolic representations of the  basic constituents (phrases) in the source and target languages.
Solutions to this problem are limited for neural technology.
We investigate  problems arising from the fact that NMT operates on implicit word and sentence representations in a continuous space, which makes influencing the process with external knowledge more complex.
In this section, we attempt to answer the following questions: \textit{i)} How to enforce the presence of a given translation recommendation  in the decoder's output;
\textit{ii)}  How to place these word(s) in the right position; and \textit{iii)}  How to guide the translation of out-of-vocabulary terms? 

\subsection{Related Work} 
\label{sec:relatedwork}
In phrase-based technology, the injection of external knowledge in the decoder is usually handled with the so-called XML markup.
This technique is used to guide the decoder by supplying the desired translation for certain source phrases.
The supplied translation choice can be injected in the output by using different  strategies. Examples include manipulating the phrase table, either by replacing entries that cover the specific source phrase or by adding alternative phrase translations, so that they compete. 

This problem has so far only been explored in NMT, and in most cases the proposed solutions integrate external knowledge at the training stage. 
Time-consuming training routines limit the suitability of this strategy for  applications that require real-time translations.
Monolingual data can be used to train a neural language model that is integrated in the NMT decoder by concatenating their hidden states. 
In one study \cite{arthur-neubig-nakamura:2016:EMNLP2016}, the probability of the next target word in the NMT decoder was biased by using lexicon probabilities computed from a bilingual lexicon. 
When the external knowledge takes the form of linguistic information such as POS tags or lemmas, separate embedding vectors can be computed for each piece of linguistic information and they can then be concatenated  without altering the decoder \cite{sennrich-haddow:2016:WMT}.
Other solutions exploit the strengths of PBSMT systems to improve NMT by pre-translating the source sentence. In \cite{NiehuesCHW16}, the NMT model was fed with a concatenation of the source and its PBSMT translation.
Some of these solutions led to improvements in performance, but they all require time-intensive training of the NMT models to use an enriched input representation or to optimize the parameters of the model. 
\cite{stahlberg-EtAl:2016:P16-2} proposed an approach that can be used at decoding time.
A hierarchical PBSMT system was used to generate the translation lattices, which are then re-scored by the NMT decoder. During decoding, the NMT posterior probabilities are adjusted using the posterior scores computed by the hierarchical model. 
However, by representing the additional information as a translation lattice, 
this approach does not allow the use of external knowledge in the form of quality judgements, as our model does.
A different technique is post-processing the translated sentences. \cite{jean-al-acl2015} and \cite{Luong2015StanfordNM} replaced the unknown words either with the most likely aligned source word or with a translation that was determined by another word-alignment model. 

The closest approach to ours is the one by \cite{hokamp-liu:2017:Long}.
They explored all the possible constraints (or translation options) at each time-step, making sure not to generate a constraint that had already been generated in a previous timestep.
Their approach generated all the constraints in the final output; thus, implicitly the model assumed that only one translation option would be provided as constraint for a specific source word or phrase. 
However, one real-life scenario is that the target language might be more inflected than the source language.
In this case, a source word can have several translation options and the decoder should decide on-the-fly which is best, depending on the source context. 
Our approach can handle both scenarios, which makes it highly suitable in practice.

\subsection{Guided NMT decoding}
\label{sec:guided-nmt}
The NMT workflow, as described in Chapter 2, has no easy way to integrate partial translations provided by an external resource. 
Unlike a PBSMT decoder, which recalls each source phrase translated at each step, the NMT decoder does not have this information.  The only indirect connection between the target word $y_t$ generated at time $t$ and the corresponding  source positions (words) is represented by the attention model weights $\alpha_{t,j}$.
These weights are used to create the context vector $c_t$ from the encoder hidden states. 
Again, unlike decoding in PBSMT, the NMT architecture
does not apply any coverage constraint on the source positions. 
Thus, there is no guarantee that the output generated by NMT covers (translates) each source word exactly once.

To overcome these problems, we present a technique called ``guided decoding'', which forces the decoder to generate specific translations that are provided as external knowledge. 
In the case of APE, where we address the problem of over-correction, the external resource comprises quality judgments of the MT text.
The quality labels indicate whether an MT word is correct, which is integrated in the input representation as shown below:
\\
\textit{\noindent <seg id="1"> <n translation=``I''> I </n> <n translation=``want''> want <n> <n translation=``to''> to </n> playing </seg>}
\\
An example might be where the MT output is ``I want to playing'' and the QE system labels the first three words as correct.
This information is integrated by annotating each correct word by itself within the XML tags. 
The annotations are placed in a ``\textit{n}'' tag that has the attribute ``\textit{translation}'' to hold the translation recommendation for the corresponding source word.
The decoder parses the XML input and creates two parallel input streams, one that contains MT words and another that contains the corresponding suggestions or an empty string.
Then, the overall decoding process is performed out as usual but 
with a different interaction between the beam search and the network. 
After a new beam of $K$-top target words is generated, the ``guide'' 
mechanism checks the $K$ hypotheses and their attention model weights.
If necessary, it influences the beam search with the external suggestions.
This is done by \textit{i)} prioritizing the hypotheses that can generate the suggestions provided; \textit{ii)} performing look-ahead steps with the beam search, to evaluate the current hypothesis and \textit{iii)} applying various strategies to manage out-of-vocabulary (OOV) terms.

\paragraph{Forcing the presence of a given term}
\label{sec:chal-1}
In PBSMT, XML markups can be easily handled.
When looking for translation options for each input phrase, the decoder checks both the external suggestions and the options in the phrase table. 
However, the NMT process is  too complex to follow a similar approach. 
When generating a target word,  NMT  assumes a continuous representation of the whole input sentence through a context vector. 
Specifically, all  input words can - in principle - contribute to generate a target word; in addition, different hypotheses may focus on different input words at the same decoding step. 
Thus, it is not guaranteed that the output at a given time-step depends solely on a specific input word.
It is also not clear how the external suggestions could be used. 
We address this issue by using the probability distribution of the input word positions, obtained from the attention model used to create the context vector. 
At each step of the beam search, for each $K$-generated target word, we looked for the most  probable input word position provided by the attention model. If the corresponding input word has a suggestion,  we replaced the target word by the given suggestion and updated the score of the hypotheses.
If not, we kept the original target word.

\paragraph{Placing the term in the right position}
\label{sec:chal-2}
The guiding mechanism allowed the decoder to generate a given translation by replacing options inside the beam. 
However, the method did not consider cases in which one input word position was involved in generating many  output words. 
This can happen when the decoder attends to a specific input word more than once - for example, an article and a noun in the output, which refer to the same noun in the input.
In these situations, it could happen that valid translation options are erroneously replaced and the external suggestion is reproduced many times in the output.

To address this problem and render our approach  robust to possible attention model nuances, we relaxed the hard replacement of a translation option if it differed from the provided suggestion. Specifically, if the 
conditions for a replacement occurred, we also checked whether the beam search would still generate the suggestion
from the current word, within a few steps. If so, we kept the current word in place because we knew that the actual suggestion would be generated soon. 
If the suggestion was not reachable,  we forced the replacement. 

\begin{algorithm}[!htp]
\begin{small}
\caption{Guided Beam Search Step}
\label{algo:guidedbeamsearch}
\begin{algorithmic}[1]

 \LineComment $K$: size of beam
\LineComment $L_{t}$: K lists of generated suggestions
\LineComment $N$: look-ahead step to check reachability 
\LineComment $S_{t}$ = $[y_{t}, s_{t}, q_{t}, b_{t}, \alpha_{t}]$: state information
\LineComment $y_{t}$: K target words
\LineComment $s_{t}$: K decoder layer hidden states
\LineComment $q_{t}$: K cumulative language model scores 
\LineComment $b_{t}$: K backtracking indexes
\LineComment $\alpha_{t}$: K highest-attention-indexes
\LineComment Global variable with suggestions:
\LineComment $\ \ \ \tilde{y}[j]$: target word for source position $j$ 
\Procedure{GuidedBeamSearch($K$,L$_{t-1}$,$N$,S$_{t-1}$)}{}
\LineComment{Perform a step of beam search}
\State $S_{t}$:= BeamSearchStep($S_{t-1}$)
\LineComment{Copy generated suggestions from parent}
\State $L_{t}$:=UpdateLists($b_{t}, L_{t-1}$)
\LineComment{for each entry of the beam}
\For {$k \in \{1,\ldots,K\}$}
\LineComment{Check suggestion for source word $\alpha_{t,k}$}
\If{$\tilde{y}[\alpha_{t,k}] \ne \emptyset \wedge \alpha_{t,k} \notin L_{t,k}$ }
\State $\tilde{y}:=\tilde{y}[\alpha_{t,k}]$
    \If{$y_{t,k} \ne \tilde{y}$}
    \LineComment{if $\tilde{y}$ is not generated by $N$ steps}
        \If{!Reachable($S_{t},\tilde{y},k,N$)}
            \LineComment{Force suggestion in beam}
            \State $y_{t,k}=\tilde{y}$;  
            \LineComment{Update suggestion list}                  
            \State Add($\alpha_{t,k},L_{t,k}$)
            \EndIf  
    \Else
        \LineComment{Suggestion is generated }
        \State Add($\alpha_{t,k}, L_{t,k}$)
    \EndIf
\EndIf
\EndFor
\State{return ($L_{t},S_{t}$)}
\EndProcedure

\end{algorithmic}
\end{small}

\end{algorithm}

Algorithm \ref{algo:guidedbeamsearch} illustrates the modified beam search process that generated the $K$  best hypotheses for the next target word. 
Starting from the beam at time ${t-1}$, a new state $S_{t}$ was computed and returned. 
The state contained the best $K$ target words ($y_t$), their corresponding decoder hidden states ($s_t$), cumulative language model scores ($q_t$), backtracking indexes to the parent entries in the previous state ($b_t$), and input indexes having the largest attention weight ($\alpha_t$). 
In addition, the modified beam search algorithm maintained, for each $K$ entry, the list of suggestions ($L_t$) that had been generated so far within that hypothesis. 
The algorithm accesses the global variable $\tilde{y}[j]$, which contains, for each source position $j$, either a provided target word suggestion or the empty word $\emptyset$. 
The algorithm proceeds by computing the normal beam search step (line 14), and initializes the lists of generated suggestions with the list of the corresponding parents (line 16) that are accessible through the backtracking indexes. 
The main loop (line 18) checks, for each beam  entry, the input position that received the highest weight by the corresponding attention model.  
If this input position ($\alpha_{t'k}$) corresponds to a non-empty suggestion and if the suggestion has not been generated by one of the predecessors of this entry, the algorithm decides whether this suggestion ($\tilde{y})$ must be forced in the beam. 
There are two cases for which action is taken. 
First, if the suggestion differs from the word in the beam (line 22) and the suggestion will not be generated by a next $N$ successor, the suggestion replaces the current word (line 26).  
The list of  suggestions generated by this hypothesis is updated accordingly. 
Second,  if the suggestion is equal to the word in the beam (line 30), then the suggestion was generated directly by the beam search  and the corresponding list is updated (line 32).  
The algorithm finally returns the updated lists of generated suggestions and the updated beam
search state. 

\begin{figure}[h]
\centering
\includegraphics[scale=0.375]{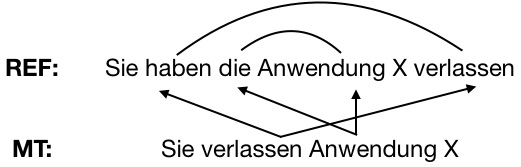}
\caption{An example showing continuous (``$die$ $Anwendung$'') and discontinuous (``$haben$...$verlassen$'') target phrases.}
\label{fig:dis-continous_phrase}
\end{figure}

This algorithm can generate both continuous and discontinuous target phrases, as follows.
\\\textbf{Continuous phrases} are those in which consecutive target words are pointed by the same input word. 
The phrase pair (``$Anwendung$'', ``$die$ $Anwendung$'') in Figure \ref{fig:dis-continous_phrase} indicates that the missing article ``$die$'' in the MT output should be generated along with ``$Anwendung$''.
With a look-ahead  window set to 1 in the algorithm, the decoder can generate the bi-gram phrase ``$die$ $Anwendung$''.
With larger look-ahead windows, longer phrases can be generated.
\\\textbf{Discontinuous phrases} are those in which  target words pointed by the same source word are intermingled with other words, for which the attention points elsewhere.
The phrase pair (``$verlassen$'', ``$haben$ ... $verlassen$'') in Figure \ref{fig:dis-continous_phrase} falls in this category, where the MT word ``$verlassen$'' should be corrected to a phrase that contains intermediate words.

\paragraph{Guiding the translation of OOV terms}
\label{sec:chal-4}
The last problem requires dealing with suggestions that are OOV words.
In NMT, it is common practice to replace OOV words by the unknown token (UNK) and use its corresponding embedding.
Two questions arise: \textit{i)} If an OOV suggestion is given by the  external resource, should the modified beam search force it into the beam?; \textit{ii)} Which target word embedding should be used in the next step?
To answer these questions, we implemented a lookup table to store all the OOV suggestions along with their unique \textit{id} before initializing the decoder.
These \textit{id}s are used for OOV suggestions by the beam search, rather than the \textit{id} associated by default with the UNK token.
In order to generate word embeddings for OOV terms, we tested several strategies:
\textit{i)} using the embedding of the unknown word, \textit{ii)} using the embedding of the best target word in the beam, \textit{iii)} using the embedding of the previous word ($y_{t-1}$), and \textit{iv)} using the average of the embeddings of all the previous words ($y_{1,..,t-1}$). 
The best results are obtained when using the embedding of the unknown word which, on  further investigation, resulted to be close to rare words in terms of cosine similarity.

\subsection{Experimental Settings}
\label{sec:T2APE}
As mentioned earlier, we applied guided decoding to an automatic PE task to mitigate the problem of over-correction.
The neural-based APE decoders would  benefit from external knowledge that indicated which words in the input were correct and should not be modified during decoding. 
We proposed using word-level binary QE labels \cite{Blatz:2004,wmt14-fbk} to annotate ``good'' words that should be preserved in the final output.
We evaluate our proposed decoding mechanism on domain-specific data using existing neural APE models as discussed below.

\paragraph{APE models}
We used an existing pre-trained model (EN-DE) that was a component of the best submission at the 2016 APE shared task at WMT \cite{junczysdowmunt-grundkiewicz:2016:WMT}.
This available model was trained with the Nematus toolkit over a dataset of $\sim$4M back-translated  pairs, and then adapted to the task-specific data segmented using  the BPE technique.

\paragraph{Test data}
In this experiment, we used the English-German IT domain data released at the WMT 2016 APE shared task \cite{bojar-EtAl:2016:WMT1}.
To annotate the test set, instead of relying on automatic  quality predictions from a QE system, we exploited oracle labels indicating ``good'' words (to be kept in the output) and ``bad'' words (to be replaced by the decoder).
To this end, we first aligned each MT output with the corresponding human PE using TER \cite{snover2006study}. 
Then, each MT word that was aligned with itself in the PE was annotated as ``good''. 
This resulted in many ``good'' labels (on average, 79.4\% of the sentence terms). 
It is worth noting that, by construction, the resulting quality labels were ``gold'' annotations that current QE systems can only approximate. These made them suitable for our testing purposes, as they allowed us to avoid the noise introduced by sub-optimal predictors. 
The BPE-level version of the test set was obtained by projecting the word-level QE tags into the sub-words; all sub-words of a word received the original word tag. If a sub-word was labelled  ``good'', we annotated it with itself to indicate that the decoder must generate the sub-word in the output.

\subsection{Results}
\label{subsec:T2disc}
Our  results   on  the APE task are reported in 
Table~\ref{tab:result-APE}.
Our first baseline (\texttt{Base-MT}) - which was the same as that used at WMT - corresponded to the original, untouched MT output.
Our second baseline (\texttt{Base-APE}) 
was a neural APE system  that was trained on (\textit{MT\_output}, \textit{MT\_post-edit}) pairs but ignored  the information from the QE annotations. \texttt{Base-APE} improved the \texttt{Base-MT} up to  3.14 BLEU points.

\begin{table}[h]
\begin{small}
\centering
\begin{tabular}{lll}
\hline \hline
   & BLEU ($\uparrow$) & TER ($\downarrow$)\\
  \hline
  \texttt{Base-MT} &62.11 &24.76 \\
  \texttt{Base-APE} & 65.25 &	23.67  \\
  \texttt{GDec\_base} & 62.68$\dagger$ & 23.97$\dagger$ \\
  \texttt{GDec\_base+OOV}  & 62.69$\dagger$ &	23.96$\dagger$ \\
  \texttt{GDec\_base+OOV+reach}  & \textbf{67.03$\dagger$} & \textbf{22.45$\dagger$}\\ \hline \hline
\end{tabular}
\caption{Performance of different decoders on the APE task measured in terms of TER ($\downarrow$) and BLEU score ($\uparrow$) (``$\dagger$'' indicates statistically significant differences wrt. \texttt{Base-APE} with p\textless 0.05).}
\label{tab:result-APE}
\end{small}
\end{table}

Our guided decoder was evaluated incrementally. 
\texttt{GDec\_base} forces the ``good'' words in the automatic translation to appear in the output according to the mechanism described in Section \ref{sec:chal-1}. This basic guidance mechanism yielded only marginal improvements over the \texttt{Base-MT} and was far behind the \texttt{Base-APE}. This can be explained by the many constraints (``good'' words to be kept), which strongly reduced the freedom of the decoder to generate surrounding words. This reasoning was confirmed by manual inspection;  many original MT segments were missing function words that depended on the ``good'' words present in the sentence. These 
insertions are easily performed by the unconstrained \texttt{Base-APE} decoder but are unreachable by \texttt{GDec\_base},  which is only able to keep the annotated words. 

\texttt{GDec\_base+OOV} integrates the mechanism to handle OOV annotations. 
Because the model was trained on the BPE segment corpus, the problem of OOV was already addressed by the model itself.
Thus, we did not observe a significant contribution by this mechanism, a finding that was in-line with our results for BPE in the MT task.

\texttt{GDec\_base+OOV+reach} is our full-fledged system, which manages repetitions and insertion positions. 
Its ability to better model the surroundings of the annotated words  enabled this technique to achieve statistically significant improvements (+1.78 BLEU, -1.22 TER) over the strong \texttt{Base-APE} decoder.

To better appreciate the ability of the APE decoder to leverage the QE labels and avoid over-correction, we compute the APE precision.
The \texttt{GDec\_base +OOV+reach} decoder gained 9 precision points over \texttt{Base-APE} (72\% vs 63\%).
This results confirmed that guided decoding, supported by QE labels, 
improved the APE output quality as well.

\begin{table*}[h]
\begin{tabular}{p{15.5cm}}
\hline  \hline
\textbf{Src:} Specifies the source for the glow .\\
\textbf{MT:} <n translation="gibt||Gibt">Gibt</n> die Quelle</n> für</n> das Glü@@ hen aus .\\
\textbf{Base:} Legt die Quelle für das Glühen fest .\\
\textbf{GDec:} Gibt die Quelle für das Glühen aus .\\
\textbf{Ref:} Gibt die Quelle für den Schein an .\\
\hline
\textbf{Src:} Map Japanese indirect fonts across platforms to ensure a similar appearance .\\
\textbf{MT:} ... zu einem ähnlichen Erscheinungsbild <n translation="gewährleisten"> gewährleisten </n> .\\
\textbf{Base:} ... " auf einem ähnlichen Erscheinungsbild an .\\
\textbf{GDec:} ... " auf , um ein ähnliches Erscheinungsbild zu gewährleisten .\\
\textbf{Ref:} ... zu , um ein ähnliches Erscheinungsbild zu gewährleisten .\\
\hline
\textbf{Src:} All values , even primitive values , are objects .\\
\textbf{MT:} alle Werte , auch Grund@@ werte , <n translation="handelt">handelt</n> <n translation="es">es</n> <n translation="sich">sich</n> <n translation="um">um</n> <n translation="Objekte">Objekte</n> .\\
\textbf{Base:} Alle Werte , auch Grundwerte , sind Objekte .\\
\textbf{GDec:} Alle Werte , auch Grundwerte , handelt es sich um Objekte .\\
\textbf{Ref:} Bei allen Werten , auch Grundwerten , handelt es sich um Objekte .\\
\hline \hline
\end{tabular}
\caption{Examples covering some cases where GDec improves over the baseline for APE task.}
\label{tab:ape-exp}
\end{table*}

\paragraph{Manual Analysis}
We manually analyzed the outputs generated by various APE systems.
Examples that  captured various aspects of the workings of GDec in this task are provided in Table \ref{tab:ape-exp}.
The labels Src, MT, Base, GDec, and Ref represent, respectively, the source sentence, MT output, baseline APE output, GDec full-fledged output, and the reference translation.
Example 1 shows the capacity of GDec to preserve the MT words in the final output that were correctly generated by the MT system.   
In this example, the word \textit{``Gibt''} (En: \textit{``Specifies''}) was preserved by GDec which would otherwise be translated to \textit{``Legt''} (En: \textit{``Sets''}) by the baseline system.
Example 2 shows that guiding the neural decoder by marking the MT word \textit{``gewährleisten''} (En: \textit{``ensure''}) as ``Good'' not only helped to preserve it in the final output, but also improved other parts of the translation.
For instance, \textit{``um ein ähnliches''} (En: \textit{``a similar''}) would otherwise be left untouched  by the baseline APE system.
Example 3 illustrates that GDec was useful for avoiding the problem of over-correction.
The MT segment in this example was almost a correct translation of the source sentence and should be left untouched, but the baseline APE system modified, thus damaging the overall translation quality.
However, when the MT word was annotated to itself by the XML tags, GDec preserved this word, thereby avoiding over-correction and retaining the translation quality.

\section{QE and APE Integration Strategies}
The interaction and possible joint contributions of QE and APE technology have barely been explored in the literature. 
This fact is surprising because both  tasks can rely on the same type of training data, consisting of (\textit{source}, \textit{MT}, \textit{post\_edited MT}) triplets.
In principle, this commonality allows for knowledge transfer and model sharing.
Building on this consideration,  \cite{Martins:TACL1113} exploited the synergies between the two related tasks by using the output of an APE system as an extra feature to boost the performance of a neural QE architecture. The intuition is that word-level quality labels can be automatically obtained through TER alignments between the translated and the PE sentence (used as a ``pseudo human PE''). The resulting APE-based QE system achieved state-of-the-art performance at the word and sentence-level QE tasks for the WMT 2015 and WMT 2016 datasets.
Another line of research that is closer to our work has focused on improving APE performance by leveraging word-level QE predictions. In \cite{hokamp:2017:WMT}, this was done by incorporating word-level features as factors in the input to a neural APE system.
Through an alternative approach, we explored our ``guided-decoding'' mechanism (discussed in the previous section) to guide a neural APE system with word-level binary QE judgments. 
Another solution for the integration of QE and APE was explored in our earlier work \cite{chatterjee-EtAl:2016:WMT}, discussed in Chapter \ref{chap:hybridAPE}, in which sentence-level quality predictions were used to select the best translation between the raw   output  versus  the correction produced by a factored phrase-based APE model.

To summarize, QE and APE can be combined in different ways to enhance MT quality. In the next sections, we identify and evaluate the following  three alternative strategies:

\begin{enumerate}
    \item QE  as  APE \textit{activator}: QE predictions are used to trigger  APE decoding when the estimated quality of the  MT segment is below a certain threshold.

    \item QE as \textit{guidance} for the APE decoder: QE labels are used to inform the APE decoding process by 
    discriminating which tokens in the MT output should be kept or changed.
    
    \item QE as \textit{selector}: QE predictions are  used to identify the best alternative between the  raw MT and its automatically corrected version. This decision can be performed either  at the level of entire sentences or for portions of the two alternative outputs. 
\end{enumerate}

\subsection{QE as APE activator}
\label{subsec:activator} 
In this first scenario, QE was used to enable a decision about whether to activate APE.
This was done by running a sentence-level QE on the MT segment to predict its TER score, then setting a threshold on this prediction. If the predicted TER was below the threshold, the translation 
was considered as good enough and the application of an APE step was unnecessary. 
By contrast, if the predicted TER was above the threshold, the APE decoder was run and its output was shown to the end user.

\subsection{QE as APE guidance}
\label{subsec:guidance}
In the previous ``QE as APE activator'' strategy, the APE decoder was not directly aided by QE to improve its performance because QE was applied before the correction process was run. This prevented QE from supporting APE to address specific problems such as over-correction.
To overcome this limitation, an alternative strategy is  a QE+APE pipeline, in which fine-grained word-level QE judgments for each  MT token were passed as additional information to the APE decoder.
The  aim was to  guide the decoder to change only those words marked as ``bad'' (that is, to retain correct MT tokens in the APE output).

We evaluated this combination strategy for both phrase-based and  neural methods.
In the case of phrase-based APE decoding, the XML markup technique\footnote{http://www.statmt.org/moses/?n=Advanced.Hybrid\#ntoc1} can easily be applied, as discussed in the above section. With this approach, word-level QE labels were directly passed to the APE decoder by specifying a fixed translation for a specific span of the source sentence. If the predicted label was ``good'', the suggested span contained the original MT words - that is, the decoder was forced to preserve them in the final output.
If the label was ``bad'', the corresponding MT word was not marked, thus giving the decoder the freedom to modify it. Among the various strategies to combine the suggested translation options and those proposed by the APE model, we experimented with the \textit{inclusive} setting (see Section \ref{sec:Results}).
Here, the proposed options competed against  other translation candidates in the phrase table.
When the APE system was based on a neural decoder, the XML markup strategy was implemented, following our novel guided decoding mechanism (discussed in the previous section). 
Similarly to the XML markup, the ``good'' labels were transformed in suggested spans containing the MT words, thus pushing the decoder towards using them.
For ``bad'' word-level predictions, in contrast, the decoder did not receive any constraint and was free to produce the most probable output.

\subsection{QE as MT/APE selector}
\label{subsec:selector}
A third possible strategy is to exploit QE predictions at a downstream level, after APE processing. After the APE decoder has generated its output, QE can be used to determine the best output option between the original MT and the PE segments. In our experiments, this was done in two ways.
The first method was to annotate only the MT segment with word-level QE labels,
and the second was to annotate both the MT and the APE outputs at the sentence or word-level. 

In the first scenario, the MT and APE segments are first word aligned. Then, using the MT output as a backbone, words are retained or modified based on their QE predicted labels. MT words labelled as ``good'' are retained, whereas, those marked as ``bad'' are replaced by the corresponding alignments from the APE output. 

In the second scenario, both the MT and APE sentences are labelled with sentence-level QE predictions (TER scores), and the one with the lower predicted TER score is selected as final output. To make the decision process more robust, a threshold  $\tau$ can be set on the difference in TER between the two segments. 
For instance, if the goal is to take a conservative approach favoring the MT system, a threshold can be set such that APE outputs are selected only if their predicted TER is much lower than that of the MT (TER$_{MT}$ - TER$_{APE}$ $>$  $\tau$).

When both the MT and APE sentences are annotated with  word-level binary labels, the tokens marked as ``good'' are selected from one of the two segments. In detail, the MT and APE segments are first word aligned and then the MT is taken as the backbone. For each MT word, the QE labels are compared. If the MT label is ``good'' and the APE is ``bad'', the MT word is used. 
If the MT label is ``bad'' and the APE is ``good'', the APE word is selected. In cases where the annotations are the same ( both ``bad'' or both ``good''), the MT word is chosen. In the case where either the MT or APE word is aligned to NULL with a ``good'' label, the word is added to the final output. Although this technique is rather simple, its performance is competitive, as shown in Section \ref{sec:Results}.

\subsection{Experimental Settings}
\label{sec:experiments}

\paragraph{Data} The experiments were performed using the EN-DE datasets released for the APE and QE shared tasks at WMT 2016.
The data is the same as used in Chapter \ref{chap:hybridAPE}.
The data belong to the Information Technology domain, and the PEs are created by professional translators. The training set contains 12K triplets, the development set 1K, and the test set 2K items. 
For the word-level QE task, the three sets had 21.4\%, 19.54\% and 19.31\%  ``bad'' labels, showing an unbalanced distribution towards the ``good'' quality tokens.

\paragraph{QE systems}
\label{subsec:QE_Exp}
To generate the sentence-level predictions, we used the best system submitted at the 2016 QE shared task \cite{kozlova-shmatova-frolov:2016:WMT}. It consists of a pipeline of two regressors.
The first one, given a set of features, predicts the BLEU score; the second one, given the predicted BLEU value, predicts the TER score. Several features are combined, including features extracted from the parse trees of the sentences, pseudo-references, back-translation, web-scale language model, and word alignments. 

At word-level, the best performing system at the 2016 QE shared was used \cite{martins-EtAl:2016:WMT}. It is a  stacked architecture that combines three neural models: one feed-forward and two recurrent models. The predictions of these three models are added as features in a feature-based linear sequential model. Syntactic dependency-based features are combined with the baseline features released by the task organizers.

In our experiments, we tested different QE-APE integration strategies using either the predicted QE annotations produced by the above systems or the gold (Oracle) labels released by the task organizers. The main reason for using oracle labels was to evaluate the improvements  in MT quality that would be possible given a perfect QE predictor.\footnote{It is important to note that there are several perfectly valid translations of the same input text, so the gold QE predictions that we use are a subset of possible oracle labels generated based on the available reference sentence.} 

\paragraph{APE systems}
\label{subsec:APE_Exp}
The outputs of two APE systems were used in the ``QE as APE activator'' and  ``QE as MT/APE selector'' experiments. 
The first system was our submission to the APE shared task at WMT 2016 \cite{chatterjee-EtAl:2016:WMT}, which was the best system among all the phrase-based solutions (see Chapter \ref{chap:hybridAPE}).
The second system was a neural-based APE solution to the same round of the shared task, which out-performed all other submissions. 
The system was based on an attentional encoder-decoder model trained with sub-word units. It was an ensemble of monolingual (\textit{MT} $\rightarrow$ \textit{pe}) and cross-lingual (\textit{source} $\rightarrow$ \textit{pe}) systems combined in a log-linear model. A task-specific feature based on string matching was added to the log-linear combination to control the faithfulness of the APE results with regard to the input. Unlike the PBSMT system, the neural model requires vast training data. 
These data are obtained by a ``round-trip translation'' process that generates source and MT segments starting from the reference sentences. In total, $\sim$4 million artificial triplets were used to train a generic neural APE system that was then fine-tuned on the task-specific (APE) data.  

In the ``QE as APE activator'' and ``QE as MT/APE selector'' strategies, the interaction between the QE and the APE systems was performed before and after decoding. 
For this reason, we directly took the submissions of the two systems to the APE shared task. 
In the tighter integration explored in the ``QE as APE guidance'' experiments, the test set must be post-edited when the QE information is available to the decoder. For this purpose, the PBSMT APE system - an instance of the Moses toolkit  with standard parameters - was trained  only on the task-specific data. 
This differed from the best PBSMT APE system at WMT 2016 because we used only an n-gram word-level language model. 
For the guided decoder, the implementation and settings described in the above section were used. 
This meant that the neural APE model was trained using the large ``round-trip translation'' dataset, and then adapted to task-specific data. The network parameters were as follows:
word embedding dimensionality of 600; hidden unit size of 1,024; maximum sentence length of 50; batch size of 80; and vocabulary size of 40K.  
The network parameters were optimized with Adadelta \cite{zeiler2012adadelta}. 
For the guided decoder, the best value for the look-ahead step was defined on the development set.
Each QE word-level annotation was projected to all the subword units. We used only a single \textit{MT} $\rightarrow$ \textit{pe} model instead of an ensemble of models.
It is important to note that both the APE systems are strong. In fact, they  showed significant improvements over the MT output, with +2.64 and +5.54 BLEU points, respectively, for the phrase-based and the neural-based systems.

\subsection{Results and Discussion}
\label{sec:Results}
In this section, the light and tight integration of QE and APE are evaluated to identify conditions where the translation quality can be enhanced. In all our experiments, the combined QE and APE systems were compared against \textit{i)} the original MT baseline and \textit{ii)} the APE system without QE (official WMT submissions when possible, our implementations in the results reported in Table \ref{tab:guidance}). Both PBSMT and neural APE were considered. 
In the APE shared task at WMT 2016, both systems outperformed the MT baseline significantly. The performance of all the systems was evaluated in terms of BLEU scores against the reference translation.

\paragraph{QE as APE activator.} 
In this set of experiments, we investigate whether the light integration of QE, as a way of triggering the automatic correction of MT texts, could  improve the translation quality. For this purpose, both sentence-level QE predictions and Oracle values were computed for each MT sentence. If the values were larger than a threshold, the APE decoder was run to improve the translation. Different TER thresholds on the QE scores were tested on the development set, ranging from 0 to 100 in step 5, and the best performing value was applied to the test set. The performances of the MT, APE (WMT systems), APE+QE predictions, and APE+QE$_{ORACLE}$ are reported in Table \ref{tab:activator}.

\begin{table}[ht]
\centering
\begin{tabular}{|l|ll|}\hline
      & \multicolumn{2}{|c|}{APE System}  \\
      & PBSMT         & Neural           \\\hline\hline
MT    & 62.11         & 62.11            \\\hline
APE (WMT systems) & \textbf{64.75}         & \textbf{67.65}                \\
APE + QE  & 64.47         & 67.19              \\\hline
APE + QE$_{ORACLE}$  &  64.58      & 67.56       \\
\hline
\end{tabular}
\caption{BLEU scores for QE as APE activator. These results are obtained using a threshold of 10 TER points.}
\label{tab:activator}
\end{table}

As reported in Table  \ref{tab:activator}, for the two APE+QE configurations, a marginal drop in performance indicated that using sentence-level QE did not aid this task.
The use of the  Oracle  labels (last row in table) was marginally better than the use of predicted QE scores, but worse than using the top performing APE systems without QE. This was true for both the PBSMT and neural APE systems. Our intuition was that  sentence-level QE scores provided information that was too coarse-grained and did not give any hint to the APE system about what was wrong in the MT output or how difficult it was to correct it. For instance, not running the APE decoder when the TER (either Oracle or predicted) was small did not prevent APE from correcting the few  errors present in the MT segment.

\paragraph{QE as APE guidance.} 
To better support APE, a tighter integration of the two technologies was obtained by injecting word-level QE annotations directly into the decoder. This was done using the XML markup in the case of the PBSMT model, and through  guided decoding for the neural model. 
The results are reported in Table \ref{tab:guidance}.\footnote{The  APE results are different compared to the ones reported in Table \ref{tab:activator} because our PBSMT APE only uses the word-level language model, and our neural APE is a single \textit{MT} $\rightarrow$ \textit{pe} system instead of an ensemble.}

\begin{table}[ht]
\centering
\begin{tabular}{|l|ll|}\hline
      & \multicolumn{2}{|c|}{APE System}  \\
      & PBSMT         & Neural           \\\hline\hline
MT    & 62.11         & 62.11            \\\hline
APE (our implementation) & 63.47         & 65.25                \\
APE + QE  & 63.57        & 65.50              \\\hline
APE + QE$_{ORACLE}$  &  \textbf{63.78}$\dagger$      & \textbf{67.03}$\dagger$       \\
\hline
\end{tabular}
\caption{BLEU scores for QE as APE guidance (``$\dagger$'' indicates statistically significant differences wrt. APE with p\textless 0.05). 
}
\label{tab:guidance}
\end{table}

Unlike the sentence-level QE predictions, the word-level  predictions were effective; their use resulted in a small but significant gain in performance over the APE system alone. This improvement also occured with the neural decoder, which was already a stronger APE module on its own. When using the Oracle labels,
further improvements were noted in the BLEU score. 
This trend was more evident for the neural system (+1.53 BLEU), bringing it closer to the ensemble APE system (67.75 BLEU) as shown in Table  \ref{tab:activator}. A possible explanation of this larger gain compared to the PBSMT APE is that the higher generalization capacities of the neural approach, which forces the APE system to perform many changes, can be controlled using information from QE. Moreover, the guided decoder proposes a tighter integration of the QE annotations than the XML markup, which is unable to decode phrases spanning across the suggested words and those for which a modification is required.  
It is worth noting that the observed relative improvements were obtained in addition to the  simpler implementations of PBSMT and neural APE systems. For this reason,  they are not directly comparable with the APE results shown in Tables \ref{tab:activator}, \ref{tab:selector_onlyMT}, \ref{tab:selectore_BothSent} and \ref{tab:selector_BothWord}, which were obtained from participants' official submissions at WMT 2016.

\paragraph{QE as MT/APE selector.} 
In the last round of experiments, we used
QE information after the APE decoding. 
For this configuration, two solutions were explored. 
In the first, shown in Table \ref{tab:selector_onlyMT}, only the MT segment was labelled with word-level QE annotations (predicted and Oracle). 
In the second solution, both  MT and APE sentences were annotated with either sentence-level or word-level QE information.
Table \ref{tab:selectore_BothSent}  shows the result for sentence-level annotation and Table \ref{tab:selector_BothWord} for word-level.  
The results in Table \ref{tab:selector_onlyMT} show that both predicted and Oracle word-level annotations of the MT output enhanced the quality obtained by the standard APE (WMT systems). Similar to the ``QE as APE guidance'' approach, the neural APE was more sensitive to  QE information, achieving significantly higher BLEU scores over the standard APE (+0.2 for predictions and +1.56 Oracle).
We hypothesized that these results stemmed from the tendency of neural APE systems to perform many modifications on the MT output, which are not always correct. 
QE information on the MT thus limits unnecessary changes made by the APE module.  

\begin{table}[ht]
\centering
\begin{tabular}{|l|ll|}\hline
      & \multicolumn{2}{|c|}{APE System}  \\
      & PBSMT         & Neural           \\\hline\hline
MT    & 62.11         & 62.11            \\\hline
APE (WMT systems)  & 64.75         & 67.65                \\
APE + QE  & 64.87$\dagger$         & 67.86$\dagger$              \\\hline
APE + QE$_{ORACLE}$  &  \textbf{65.13}$\dagger$     & \textbf{69.21}$\dagger$       \\
\hline
\end{tabular}
\caption{BLEU scores for QE as MT/APE selector. Word-level QE annotations are produced only for the MT segment (``$\dagger$'' indicates statistically significant differences wrt. APE with p\textless 0.05). 
}
\label{tab:selector_onlyMT}
\end{table}

When both the MT and APE segments were annotated with the sentence-level QE scores, a threshold was set to decide whether to show the MT or the APE translation to the end user. 
The best results were obtained with $\tau$ being equal to 5 for the predicted TER values and 1 for the Oralce TER values (Table \ref{tab:selectore_BothSent}).
These experiments confirmed that using the predicted sentence information was not useful;  both the PBSMT and neural APE+QE systems produced outputs that were  worse than the standard APE (WMT systems). 
When using the Oracle annotations, the BLEU scores were better than those obtained by the APE system alone.
They were also in line with the performance obtained by Oracle word-level QE information on the MT segments (Table \ref{tab:selector_onlyMT}, last row). These results indicate that better QE scores would be helpul in this setting.

\begin{table}[ht]
\centering
\begin{tabular}{|l|ll|}\hline
      & \multicolumn{2}{|c|}{APE System}  \\
      & PBSMT         & Neural           \\\hline\hline
MT    & 62.11         & 62.11            \\\hline
APE (WMT systems) &   64.75         & 67.65           \\
APE + QE  &  64.49        &   66.49           \\\hline
APE + QE$_{ORACLE}$  &   \textbf{65.26}$\dagger$     &    \textbf{69.50}$\dagger$    \\
\hline
\end{tabular}
\caption{BLEU scores for QE as MT/APE selector. Sentence-level QE annotations both on the MT and APE segments (``$\dagger$'' indicates statistically significant differences wrt. APE with p\textless 0.05).}
\label{tab:selectore_BothSent}
\end{table}

The final experiment entailed annotating both the MT and  APE segments with  word-level QE information,  and defining a simple strategy to merge the two outputs (see Section \ref{subsec:selector}). Table \ref{tab:selector_BothWord} shows that both PBSMT and neural APE systems took advantage of the QE labels, offering a slight improvement over the APE systems alone.
Unsurprisingly, larger gains with both techniques were obtained using the Oracle annotations, with a performance boost in BLEU scores of +0.68 for PBSMT and +3.34 for neural APE. 
Again,  neural APE achieved the best performance and largest potential improvement, confirming that the large variability of the applied changes was indeed an advantage and could be controlled using information from QE.

\begin{table}[ht]
\centering
\begin{tabular}{|l|ll|}\hline
      & \multicolumn{2}{|c|}{APE System}  \\
      & PBSMT         & Neural           \\\hline\hline
MT    & 62.11         & 62.11            \\\hline
APE (WMT systems) & 64.75         & 67.65                   \\
APE + QE  & 64.83         & 67.79              \\\hline
APE + QE$_{ORACLE}$  &  \textbf{65.51}$\dagger$      & \textbf{71.13}$\dagger$       \\
\hline
\end{tabular}
\caption{BLEU scores for QE as MT/APE selector. Word-level QE annotations both on the MT and APE segments (``$\dagger$'' indicates statistically significant differences wrt. APE with p\textless 0.05). 
}
\label{tab:selector_BothWord}
\end{table}

Overall, considering all the experiments reported in this second part of the chapter, our main findings can be summarized as follows:

\begin{itemize}
    \item Integration strategies that exploited word-level QE seemed more promising than those based on sentence-level QE. Our results showed that sentence-level QE information was too coarse to  support  APE decoding, whereas proving a QE annotation on each MT or APE token enhanced the overall translation quality. 
    \item At word-level, predicted QE labels yielded limited but constant gains of up to $\sim$0.2 BLEU points over standard APE systems. These values were small but supported the intuition that QE and APE integration offers positive potential. 
    \item Oracle results indicated that large scope exists for improvement, if better QE systems can be designed. Although it is not guaranteed that a QE model could achieve the Oracle performance, increasing the quality of QE annotations did result in significant improvement in the translation quality. 
\end{itemize}

\section{Summary}
In this chapter, we addressed the problem of over-correction by proposing a novel guided decoding mechanism to leverage external knowledge in the form of quality judgements.
We discussed the challenges faced in the neural paradigm, and showed that a naive integration of external knowledge based on attention information was insufficient.
A more sophisticated look-ahead mechanism was needed to guide the decoding process.
Specifically, we answered \textit{i)} how to enforce a given suggestion in the decoder's output; \textit{ii)} how to place these suggestion in the right position; and \textit{iii)} how to deal with out-of-vocabulary words.
Furthermore, we performed a systematic analysis of different techniques to combine QE and APE to achieve better MT quality. These strategies ranged from light integration, in which  QE was used either to trigger APE or to compare the APE with the original MT segment, to tighter integration, in which QE annotations were directly used to guide the inner workings of the APE decoder. Our experiments showed that QE can help APE to produce better MT outputs. Among the proposed strategies,  ``QE as guidance'' and ``QE as selector'' led to improvements in MT quality.
The use of word-level QE on both MT and neural APE resulted in the largest gains over the top WMT 2016 APE system (+0.2 BLEU score with the predicted QE annotations, and +3.34 BLEU score with the Oracle labels).

\chapter{Online Phrase-based APE}
\label{chap:onlineAPE}
In previous chapters, we studied the potential of APE systems in a controlled evaluation scenario.
In this setting, the representativeness of the training set with respect to the test data is a key factor to achieve good performance. Real-life scenarios, however, do not guarantee such favorable learning conditions. An example of APE tools being integrated in a real professional translation workflow might entail their playing role in a computer-assisted translation framework. For this, APE tools should be flexible enough to handle continuous streams of diverse data coming from different domains or genres. To address this problem, in this chapter we propose an online APE framework that is: \textit{i)} robust to data diversity (that is able to learn and apply correction rules in the right contexts) and \textit{ii)} able to evolve over time, by continuously extending and refining its knowledge. 

\section{Introduction}
The effectiveness of learning from relatively small amounts of PE data is evident from the impressive outcomes of the APE shared tasks at WMT \cite{bojar-EtAl:2017:WMT1,bojar-EtAl:2016:WMT1,bojar-EtAl:2015:WMT}. 
Different APE paradigms, like neural \cite{junczysdowmunt-grundkiewicz:2016:WMT}, hybrid \cite{chatterjee-EtAl:2016:WMT}, and phrase-based \cite{usaarAPE16}, all significantly improved the MT output quality in the IT domain. The gains ranged from   2.0 to 5.5 BLEU points.
Nevertheless, the positive outcomes of previous work on APE were built on a problem formulation that assumed operation would occur in a controlled laboratory environment. In such settings, systems are trained and evaluated across a coherent or homogeneous data set. 
Moving from this controlled scenario to real-world translation workflows, where training and test data can be produced by different MT systems, post-edited by various translators and belonging to several text genres, makes the task inherently more challenging because the APE systems must adapt to all these diversities in real-time. 
One problem is that training data provide a fraction of the possible error -correction examples; this is a normal issue when learning from finite, often small training data. The additional complexity derives from two concurrent factors. First, not all the learned error correction rules are universally applicable, and applying them in the wrong context can damage the MT output instead of improving it. Second, once in production, the APE system should be able to process streams of diverse input data presented in random order. Prompt reaction to such variability is therefore crucial.  We define this more complex and realistic scenario as a \textit{multi-domain} translation environment (MDTE), where a domain is made of segments belonging to the same text genre, and the MT outputs are generated by the same MT system. 

This work represents a first step towards MDTE data in real-time or online translation scenario.
Although a full-fledged evaluation centered on human translation in a CAT framework was beyond our reach, we provided a proof of concept in which we simulated the MDTE scenario by running different APE solutions on a stream of data coming from two domains.
An important factor in a real-time translation system is to achieve low latency and high throughput to serve customers better.
This is best achieved by phrase-based technology rather than neural-based solutions.
Therefore, we analyzed alternative solutions within phrase-based APE. In this chapter, we discuss the limitations of batch APE methods (that is, insensitivity to domain shifts) as well as, state-of-the-art online systems evaluated in the APE task in MDTE conditions. 

\section{Related work}
\label{sec:relwork}
Most studies of APE have operated in
a batch framework, in which systems are  evaluated on static test sets that are homogeneous with the training data \cite{simard2007statistical,dugast2007statistical,terumasa2007rule,pilevar2011using,bechara2011statistical,chatterjee-EtAl:2016:WMT}.
These systems, however, cannot leverage the feedback of post-editors in an online translation scenario.
The capacity to evolve by learning from human feedback has been addressed by several 
 online translation systems, mainly focusing on the MT task \cite{hardt2010incremental,bertoldi2013cache,mathur2013online,simard2013pepr,ortiz2014new,denkowski2014learning,wuebker2015hierarchical,ortiz2016online,peris2018active,gonzalez2014cost,alabau2014integrating}.
From among these online MT systems, we discuss the two that were also used for the APE task.
\paragraph{\textbf{PEPr: Post-Edit Propagation.}}\cite{simard2013pepr} proposed a method for PE propagation (PEPr) that, learns post-editors' corrections and applies them on-the-fly to 
an MT output sequence.
To perform PE propagation, the system is trained incrementally using pairs of MT \textit{(mt)} and human PE \textit{(pe)} segments as they were produced. 
When receiving a new \textit{(mt, pe)} pair, 
word alignments are obtained using Levenshtein distance. In the next step, the phrase pairs are extracted and appended to the existing phrase table.
The whole process is assumed to take place within the context of a single document; for every new document, the APE system is initialized with an ``empty'' model.
This represents a possible limitation of the approach. Although document-specific correction rules show relatively high precision, some might be useful in other  contexts too and should be retained.
Our approach avoids this limitation by maintaining a global knowledge base to store all the processed documents. At the same time, it is possible, to retrieve PE rules specific to the document being translated.\footnote{In our experiments we do not  compare against PEPr since, being designed for document-level translation it is unable to operate in the  MDTE scenario.}

\paragraph{\textbf{Thot.}}The Thot toolkit \cite{ortiz2014new} was developed to support automatic and interactive statistical MT.\footnote{\url{https://github.com/daormar/thot}}
It was successfully used to experiment in an online APE setting with several datasets for multiple language pairs, with base MT systems built using different technologies (rule-based MT, statistical MT) \cite{lagarda2015translating}.
To incorporate user feedback in the underlying translation and language models, the system incrementally updates all the required statistics. 
For the language model, it simply updates n-gram counts. In the case of the translation model, the process exploits an incremental version of the expectation maximization algorithm to obtain word alignments and extract the phrase pairs, whose counts are continuously updated.
Other features, like source/target phrase-length models or the distortion model, are extracted considering geometric distributions with fixed parameters.
The feature weights of the log-linear model are static throughout the online learning process, unlike our method that updates the weights on-the-fly.
Our online APE approach is thus independent from any pre-trained model or pre-tuned feature weights.
Moreover, whereas in Thot the correction rules are learned in real-time from all the data processed, our system learns only from the most relevant data samples. Nevertheless, because Thot is state-of-the-art in online APE, we used it as a term of comparison in our experiments.

\section{Online APE system}
\label{sec:system}
The backbone of our online APE system is similar to the statistical batch APE approach \cite{P15-2026}. 
Our migration to the online scenario builds on incrementally extending this backbone with an instance-selection mechanism ($\S$\ref{subsec:is}), a dynamic knowledge base ($\S$\ref{subsec:dkb}), and new features ($\S$\ref{subsec:feat}).

\subsection{Instance selection}
\label{subsec:is}
Current batch and online APE methods estimate the parameters of a model across all available training data. This strategy might not be effective in 
the MDTE scenario,
as the model tends to become more generic by 
incorporating knowledge
from several domains. In the long-run, this can complicate the selection 
of domain-specific correction rules suitable for a particular MT segment. A possible solution is to constrain the 
system to work at document level \cite{simard2013pepr}.  In that approach, however, models are reset to their original state once the entire document is processed, which means that knowledge  gained from the document is lost. 
Our instance selection technique was aimed at overcoming this issue. It allowed the system to preserve all the knowledge acquired during the online learning process while applying specific PE rules when needed.

The instance selection 
mechanism consists of retrieving \textit{ad-hoc} training sentence pairs for each MT output to be post-edited. In practice, the creation of the APE model and estimating its parameters  are performed on-the-fly by processing  relevant instances retrieved from previously processed  data. 
In the MDTE scenario, this data comes from heterogeneous domains.  
The relevance of a training sample is measured as a similarity score, based on  term frequency-inverse document frequency (tf-idf\footnote{In MT,  tf-idf was previously used by \cite{hildebrand2005adaptation} to  create  a  pseudo in-domain  corpus  from  a  large out-of-domain corpus.  Our work is the first to investigate it for the APE task in an online learning scenario.}), computed using the Lucene 
software library.\footnote{\url{https://lucene.apache.org/}}
Indexing and retrieving training triplets (\textit{src, mt, and pe}) in this mechanism is fast, which makes it highly suitable for an online learning scenario.
The cut-off similarity score is empirically estimated over a held-out development set. 
Input segments that lack training samples above the threshold  are left untouched, to avoid possible damage from applying unreliable correction rules learned from 
unrelated contexts. 
This strategy contrasts with that adopted by current APE systems. 
Current systems tend always to ``translate'' the given input segment regardless of the  reliability of the applicable correction rules.

Once the training samples are selected for an input segment, several models are built on-the-fly.
A tri-gram local language model (LM) is built over the target side of the training corpus with the IRSTLM toolkit \cite{federico2008irstlm}. 
Along with the local LM a tri-gram global LM, is also used, which is updated whenever a human PE (\textit{pe}) is received.
Local translation and  reordering models are  built with the Moses toolkit, computing word alignment for each sentence pair  using the incremental GIZA$++$ software.\footnote{\url{https://code.google.com/archive/p/inc-giza-pp/}}

The feature weights of the log-linear model are optimized over a subset of the selected instances. 
The size of this development set is critical: if it is too large, parameter optimization is expensive. 
If it is too small, the tuned weights might not be reliable.
To achieve fast optimization with reliably-tuned weights, multiple instances of MIRA \cite{crammer2003ultraconservative} are run in parallel on 
multiple development sets \cite{Tange2011a}.
For this purpose, the retrieved samples are randomly split three times into training and development.
The tuned weights resulting from the three development runs are then averaged and used to decode the input MT segment.
To summarize, our training or tuning scheme  requires a minimum number of retrieved sentence pairs. 
Following an 80\%-20\% distribution over training and development data, and setting the minimum number of samples needed for tuning to 5, the complete process requires the retrieval of at least 25 samples. If this number is not reached, all the retrieved samples are used for training, 
the optimization step is skipped, and the previously computed weights are used. 
If no sample is selected, the MT output is left untouched.

\subsection{Dynamic knowledge base} 
\label{subsec:dkb}
The APE system described so far was built by  considering only the most similar retrieved sentences, which we hypothesized to be the most useful to learn reliable corrections 
for a given MT output.
On one hand, this strategy avoids creating correction options that are inappropriate to post-editing the MT output.
On the other hand, it computes the statistics of the models (i.e. translation and lexicalized reordering probabilities) using only a few parallel sentences; this limitation can result in potentially unreliable values that would penalize the work of the decoder. To address this issue, we complemented instance selection with a dynamic knowledge base that kept track of all the previous observations relevant for PE. Such a component provides the decoder with all translation options extracted from the retrieved sentences. However, instead of computing the probabilities only on these segments, it takes advantage of all the occurrences of a translation option in the previously processed sentences. This feature allowed our online APE system to use only the most useful translation options, associated with more reliable 
statistics.

The dynamic knowledge base was implemented by a distributed, scalable, and real-time inverted index, which after insertion made all data immediately available for search and update. The ElasticSearch\footnote{\url{http://www.elastic.co/products/elasticsearch}} engine was used for this purpose.
Once the PE was made available to our system, the word alignment between \textit{mt} and \textit{pe} was computed, and the sentence pair was split in phrases and then added to the dynamic model. 
If a translation option was already present, the phrase translation and the orientation counts were updated; if not, the option was inserted for the first time. 
This was run in multi-threading setup by additionally managing possible conflicts in terms of access to the same translation option by different threads. 
Word lexical information and  phrase counts were stored apart. 
At decoding time, 
the IDs of the samples retrieved by the instance selection method and the \textit{mt} sentences were used to query the dynamic knowledge base.  
The translation options that satisfied the query were retrieved and supplied to the decoder in the form of translation and reordering model information. All the feature scores were computed on-the-fly, with four scores for the translation model and six for the reordering model.

Compared to the suffix arrays used to implement MT dynamic models
\cite{AMTA-2014-W1-Germann,denkowski2014learning} in which whole sentence pairs were stored, our technique needed to save more information, in the form of all the translation options. However, the amount of data in APE was far less than that in MT, which mean it could be managed easily by \textit{ad hoc} solutions. In addition, it allowed us to collect global information at the translation option level, which could result in useful additional features for the model. The latter aspect is explored in the next section, which discusses the reliability of translation options as measured by
the behavior of post-editors.

\subsection{Negative feedback in-the-loop} 
\label{subsec:feat}
Similar to the  APE  systems mentioned in Section~\ref{sec:relwork}, the one  described here
stored only PE positive feedback. 
Its knowledge base of correction rules and  the statistics to estimate the model parameters were continuously updated, based only on  alignment information between (\textit{mt, pe}) pairs. PEs, however, can also be used to infer negative feedback and this can be used to penalize unreliable correction options (i.e. those resulting in PEs eventually modified by a human).
The dynamic knowledge base allowed us easily to integrate this kind of information in the form of two additional ``negative feedback'' features:
\begin{itemize}
\item F1. This feature penalizes the correction rules that are selected by the decoder but are eventually modified by the post-editor. 
Faulty selection can result from applying a rule in the wrong context (e.g. in case of domain changes in the input stream of data). 
However, it more likely occurs when the learned rule is wrong (e.g. as the result of ambiguous or incorrect word alignment).
F1 was computed as the ratio of the number of times the post-editors modified a correction made by the APE decoder, relative to the number of times the decoder made that correction.
    The information about which correction rules were applied by the APE system was obtained from  the Moses decoder \texttt{trace} option. The information about which rules were modified was derived by string matching the target side of the rule in the final human PE.
\item F2. This rule penalizes the correction rules that, even if unused, were available to the decoder (i.e. translation options discarded during decoding). Assuming that the application of these options would have been eventually be corrected by the post-editor,  
all rules in the phrase table were scanned to check if their target side (i.e. the correction) was present in the final human PE (again by string matching). If not, the corresponding rule was penalized.
This feature was computed as the ratio of the number of times the correction in the phrase table was modified - or assumed modified - by the post-editor, to the number of times the correction rule was seen in the local phrase table, for all segments processed so far.
Because the decoder operates with a segment-specific local phrase table containing only those options relevant to the segment to be post-edited, computing this feature was not expensive.
    
\end{itemize}
We also evaluated system performance by using  the two features together.
As discussed in Section~\ref{sec:exp}, although our use of negative feedback was still at a preliminary stage,  its integration in our online APE framework yielded some improvements.

\section{Experimental Setting}
\paragraph{Data.}
\label{sec:dataset}
We experimented with 
two EN-DE datasets: \textit{i)} the data released in the APE shared task at WMT 2016
\cite{bojar-EtAl:2016:WMT1}, and \textit{ii)} the data used in our previous study \cite{P15-2026}, which is a subset of the Autodesk Post-Editing
Data corpus.\footnote{\url{https://autodesk.app.box.com/v/autodesk-postediting}} 
Although they come from the same domain, namely IT, they vary in vocabulary, MT engines used for translation, MT errors, and PE style.
According to our broad notion of ``domain'', these variations contribute to making the two datasets different enough to emulate an MDTE scenario for testing online APE capabilities.
The data are pre-processed to obtain the ``context-aware'' representation
that links each MT word with its corresponding source word or words (\textit{mt\#src}). 
This representation was used as a source corpus to train all the APE systems compared in this chapter. It was obtained 
by concatenating words in the source (\textit{src}) and the MT (\textit{mt}) segments after aligning them with MGIZA++ \cite{gao2008parallel}.

The Autodesk training and development sets consists of 12,238, and 1,948 segments, respectively, whereas the WMT 2016 dataset contains 12,000, and 1,000 segments. 
Table \ref{tab:data} provides  additional statistics for the source (\textit{mt\#src}) and target (\textit{pe}) training sets; the repetition rate (RR) to measure the repetitiveness inside a text \cite{bertoldi2013cache}, and the average TER score 
for both the datasets, computed between MT and PE, as an indicator of the original translation quality. 
With regard to  these statistics, several indicators suggest that the WMT 2016 corpus provided a more difficult scenario than the Autodesk corpus for APE. First, the WMT 2016 corpus has on average longer sentences, which increases the complexity of rule extraction and decoding processes. Second, although the two datasets had a similar repetition rate, the  WMT 2016 corpus had more tokens, indicating higher sparsity of the data.  
Finally, the lower  TER  of  WMT 2016 suggested that  there were fewer corrections to perform. In turn, this meant a higher chance of damaging the original MT output if the learned rules were applied in the wrong context.
\begin{table*}[h]
\setlength{\tabcolsep}{5pt}
\centering
\begin{tabular}{|l|c|c|c|c|c|c|c|c|}
\hline
 & \multicolumn{2}{c|}{Tokens} & \multicolumn{2}{c|}{Types} & \multicolumn{2}{c|}{Avg. segment length}  & \multirow{2}{*}{\parbox{1cm}{\centering RR (\textit{mt\#src}) }} & \multirow{2}{*}{TER}\\ \cline{2-7} 
 & \textit{mt\#src}     & \textit{pe}            & \textit{mt\#src}     & \textit{pe} & \textit{mt\#src}     & \textit{pe}  & & \\ \hline
Autodesk & 153,943 & 160,801 & 31,939 & 15,023 & 12.57 & 13.13 & 4.938  & 45.35\\ \hline
WMT16 & 210,573 & 214,720 & 32211 & 16,388 & 17.54 & 17.89 & 4.907  & 26.22\\ \hline
\end{tabular}
\caption{Data statistics}
\label{tab:data}
\end{table*}

To conclude, we measured the diversity of the two datasets by computing the vocabulary overlap between the two joint representations.
Measurements were performed internally for each dataset by splitting the training data into two halves, as well as across the two datasets. 
As expected, in the first case the vocabulary overlap was much larger ($>$ 40\%) than in the second case ($\sim$15\%), indicating that minimal information was shared between the two datasets.
In our MDTE experiments, the training data were first merged, then shuffled, then split in two halves of 12,119 segments each. The same procedure was applied to the development sets.

\paragraph{Terms of comparison.}
We evaluated our online learning approach against: 
\textit{i)} the MT baseline; 
\textit{ii)} the batch phrase-based APE system  
(described in Section~\ref{sec:system}), 
to which we incrementally added our online learning extensions; and \textit{iii)} the Thot toolkit. 

\section{Results}
\label{sec:exp}
The batch APE system was trained on the first half of the shuffled training  set,
tuned with the development set (2,948 segments), and evaluated over the second half of the training data (considered as test set).
Because the batch APE learned only from the training set, we expected its performance to give a lower bound estimate, 
which was expected to be outperformed by the online APE systems that learned from the test data too.
To run the online experiments with Thot, the system first needed to estimate the feature weights of the log-linear model on a development set.
For this purpose, it was trained and tuned off-line like a batch APE system. 
Three online extensions of the batch backbone architecture, described in Section~\ref{sec:system}, were evaluated. These were: \textit{i)} the instance selection system (IS);
\textit{ii)} the dynamic knowledge base system (IS+DynKB); and  
\textit{iii)}  the dynamic knowledge base system enhanced with the negative feedback features, both alone and in combination 
(IS+DynKB+F1, IS+DynKB+F2 and IS+DynKB+F1+F2).
For all of them, the 
cut-off similarity score was  obtained by a grid search and was set to 0.8.
The results achieved by each system are reported in Table \ref{tab:results}.

\begin{table}[h]
\centering
\begin{tabular}{|l|ccc|}\hline
        & \small{BLEU$\uparrow$} & \small{TER$\downarrow$}            & \small{Prec. (\%) }           \\\hline\hline
\small{MT}        & 52.31 & 34.52          & N/A                        \\
\small{Batch APE}   & 52.52 &   34.45    &            42.67           \\
\small{Thot}    & 52.51 & 34.37         & 42.22               \\\hline
\small{IS}  & 53.35$^\dagger$  & 33.36$^\dagger$         &  58.47                   \\
\small{IS+DynKB}  & 53.60$^\dagger$  & 33.23$^\dagger$         & 59.69                    \\ \hline
\small{IS+DynKB+F1}  & 53.56$^\dagger$  & 33.29$^\dagger$         &  58.97                   \\
\small{IS+DynKB+F2}  & 53.21$^\dagger$  & 33.48$^\dagger$         & 54.64                    \\
\small{IS+DynKB+F1+F2}  & \textbf{53.77$^\dagger$}  & \textbf{33.20$^\dagger$}         & \textbf{60.93}                    \\\hline
\end{tabular}
\caption{Results on the mixed data. (``$\dagger$'' indicates statistically significant difference wrt. the MT baseline with p\textless 0.05).}

\label{tab:results}
\end{table}

As shown in Table 8.2, the batch APE system offered a slight improvement over the baseline, even if it damaged most of the translations it modified. Its precision was lower than 45\%.
Although it learned from the test data, Thot achieved 
similar results. 
This was probably due to the system's inability to identify domain-specific correction rules needed to improve the translations, thus damaging most of the modified MT segments.
A significant gain in performance (+1.04 BLEU,  -1.16 TER points) was obtained by our  online IS system, which - by using the instance selection technique - was able to isolate only the most useful training 
samples.
This mechanism also improved the precision by up to 58.4\%  ($\sim$16\% above Thot), indicating that the  applied PE rules were mainly correct. Analysis of the performance of the two online systems revealed that our approach modified fewer segments that Thot did, because it built sentence-specific models only if it found relevant data,  leaving the MT segment untouched otherwise.
In several cases, when modified by the Thot system, these untouched  segments resulted in damaged sentences.

Further performance improvements were yielded by  the dynamic knowledge base (IS+DynKB), which provided the decoder with a better estimation of APE model parameters. 
Although the BLEU and TER gains were minimal, the dynamic knowledge base helped  to significantly increase the precision of the APE system, avoiding unnecessary changes. This result confirmed the effectiveness of keeping track of the  whole past history of each translation option.
Our implementation of the dynamic knowledge base also allowed us to add the two ``negative feedback'' features that modelled the reliability of translation options by looking at the changes made by post-editors.  
When used in combination, the two negative feedback features (IS+DynKB+F1+F2) yielded visible gains in performance over (IS+DynKB), with small but statistically significant improvements in BLEU score and a precision gain of 1.24\%.
These results suggest their possible complementarity with the (default) translation and reordering features.
Overall, our full-fledged system achieved state-of-the-art results, with significant improvement over Thot, by 1.26 BLEU, 1.17 TER, and 18.71\% precision.

\begin{figure*}[h]
\centering
\includegraphics[scale=0.285]{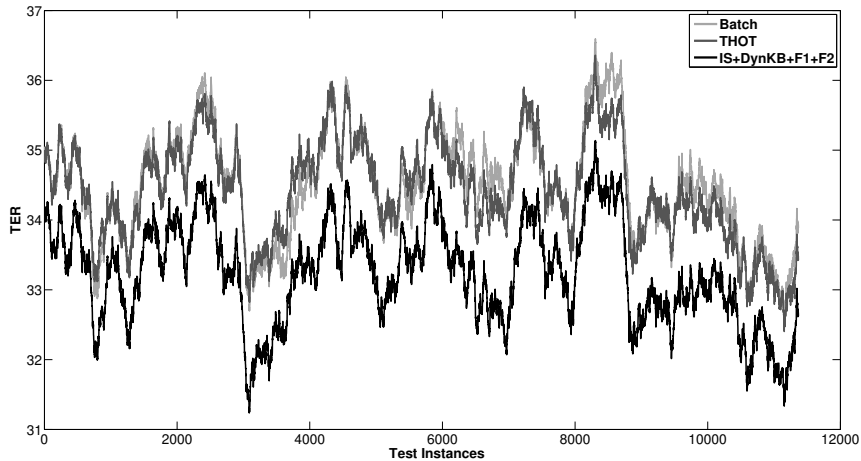}
\caption{Performance comparison of  different APE systems (TER moving average).}
\label{fig:comparison}
\end{figure*}

\begin{table}[h]
\centering
\begin{tabular}{|l|ccc|}\hline
 & \small{BLEU$\uparrow$} & \small{TER$\downarrow$}            & \small{Prec. (\%)}            \\\hline
        & \multicolumn{3}{|c|}{\small{Autodesk}} \\\hline
       \hline
\small{MT}        & 40.01 &     45.42      & N/A                         \\
\small{Batch APE}   & 43.13$^\dagger$ &   43.19$^\dagger$    &            58.86       \\
\small{Thot}    & 43.40$^\dagger$  &  42.96$^\dagger$       & 59.04               \\\hline
\small{IS+DynKB+F1+F2}  & \textbf{44.56$^\dagger$}  & \textbf{41.86$^\dagger$}         & \textbf{73.37}                   \\\hline \hline
        & \multicolumn{3}{|c|}{\small{WMT16}}  \\\hline
\small{MT}    &   61.04 & 26.24          & N/A     \\
\small{Batch APE} &   59.24$^\dagger$ &  27.81$^\dagger$    &            22.18   \\
\small{Thot}&    59.05$^\dagger$ & 27.84$^\dagger$         & 20.06      \\\hline
\small{IS+DynKB+F1+F2}  & \textbf{60.39$^\dagger$}  & \textbf{26.62$^\dagger$}         & \textbf{36.67}                      \\\hline \hline
\end{tabular}
\caption{Performance analysis of each domain.} 
\label{tab:results-domain}
\end{table}

\begin{table*}[h]
\centering
\begin{tabular}{p{1.7cm}p{14cm}}\hline
\hline
SRC     & Specifies the value to define the mid-ordinate distance by which to tessellate baseline alignment curves . \\
MT      & Gibt den Wert f{\"u}r den kürzesten Abstand vom Sekantenmittelpunkt zu Kreisbogen f{\"u}r die Tessellation Basislinienachse Kurven .      \\
MT-Top1 & Gibt den Wert f{\"u}r den kürzesten Abstand vom Sekantenmittelpunkt zu Kreisbogen f{\"u}r die Tessellation Basislinienachse Kurven .          \\
PE-Top1 & Gibt den Wert zum Definieren des kürzesten Abstands vom Sekantenmittelpunkt zum Kreisbogen an , um den Basislinienachsen-Bogen ausgerundet werden sollen . \\
THOT    & Gibt den Wert \textbf{f{\"u}r den Versatzzielbogen Abstand} vom Sekantenmittelpunkt \textbf{zu} Kreisbogen \textbf{f{\"u}r die Tessellation Basislinienachse Kurven} . \\
IS+DynKB +F1+F2     & Gibt den Wert zum Definieren des kürzesten Abstands vom Sekantenmittelpunkt zum Kreisbogen an , um den \textbf{Basislinienachse Versatzzielbogen} ausgerundet werden sollen .      \\
REF     & Gibt den Wert zum Definieren des kürzesten Abstands vom Sekantenmittelpunkt zum Kreisbogen an , um den Basislinienachsen-Bogen ausgerundet werden sollen .    \\                                                         \hline\hline      
\end{tabular}
\caption{Sample outputs where our approach outperform THOT (erroneous words are in bold)}
\label{tab:fbk_better_than_thot}
\end{table*}

\begin{table*}[h]
\centering
\begin{tabular}{p{1.7cm}p{14cm}}\hline\hline
SRC     & Drag to the left and then click to place .                                                       \\
MT      & Ziehen Sie nach links , und klicken Sie dann , um sie zu platzieren .                            \\
MT-Top1 & Ziehen Sie nach links , und klicken Sie dann , um sie zu platzieren .                            \\
PE-Top1 & Ziehen Sie sie nach links , und klicken Sie dann , um sie zu platzieren .                        \\
MT-Top4 & Ziehen Sie den Cursor nach unten und nach rechts , und klicken Sie dann , um sie zu platzieren . \\
PE-Top4 & Ziehen Sie nach unten und nach rechts , und klicken Sie dann zum Platzieren .                    \\
THOT    & Ziehen Sie nach links , und klicken Sie dann , um sie zu platzieren .                            \\
IS+DynKB +F1+F2     & Ziehen Sie nach links , und klicken Sie dann \textbf{zum Platzieren} .                                    \\
REF     & Ziehen Sie sie nach links , und klicken Sie dann , um sie zu platzieren .                       \\
\hline\hline
\end{tabular}
\caption{Sample output where our approach performs poor than THOT (erroneous words are in bold)}
\label{tab:fbk_worse_than_thot}
\end{table*}

\section{Analysis}
We wanted to understand and compare the long run behavior of the batch APE, Thot, and our best online system. The plot in Figure \ref{fig:comparison} shows the TER moving average, with a window of 750 data points, at each segment of the test set (second half of the shuffled training data).
Our approach successfully maintained the best performance across the entire test set.
Moreover, examining the first and last 250 points in the test set shows that the performance gap between our best system and Thot increased from 0.8 to 1.6 TER points on average.
 Hence, during processing, our approach was able to self-adapt  in real-time towards the domain-shifts in the input stream of data.
To better understand their behavior with respect to data coming from the two domains, we evaluated the systems' output separately per domain. The results are reported in Table \ref{tab:results-domain}.
For the Autodesk and the APE shared task domain, there were 6,166 and 5,953 segments respectively.
All the APE systems improved the translations belonging to the Autodesk domain by a large margin, with our approach offering the greatest improvement.
The same did not hold for the other domain, which was challenging due to factors such as longer sentences, data sparsity, and translation quality.
Nonetheless, even in this challenging domain our approach caused less degradation than the other APE methods, which severely damaged the translations.
Overall, compared to other APE approaches, our system showed the best performance in both domains when evaluated in isolation.

To evaluate the efficiency of our approach, we computed the average time in seconds to perform a full online  cycle over the test set, that is, the time needed for  post-editing the MT output and updating the models. 
The analysis included Thot, IS, and IS+DynKB+F1+F2. 
Thot required
 on average 4.75 seconds per cycle. The IS system, which built models on-the-fly by leveraging only the selected data and optimizing weights before PE, was faster than Thot by 1.03 seconds (3.62'' on average). The use of the dynamic model, which was faster in updating and dumping the tables, allowed our system to perform a full online cycle in 3.05''. Hence, our approach was better not only in terms of performance but also in computation time.

Tables \ref{tab:fbk_better_than_thot} and  \ref{tab:fbk_worse_than_thot} respectively show  examples where our approach performed better or worse than Thot.
Both tables report the following: \textit{i)} the source sentence (SRC); \textit{ii)} the MT output to be post-edited (MT); \textit{iii)} the MT and PE segment of the top training samples retrieved, based on cosine similarity (MT-TopX/PE-TopX, where X is the rank of the training sample); \textit{iv)} the output of Thot; \textit{v)} the output of our best system (IS+DynKB+F1+F2); and \textit{vi)} the reference (REF).
Table \ref{tab:fbk_better_than_thot} seems to confirm  our intuition that learning from the most similar examples yielded better translation quality.
An interesting counter example is shown in Table \ref{tab:fbk_worse_than_thot}. Here, despite having access to a training sample (MT-Top1 and PE-Top1) that was identical to the MT segment to be post-edited, our system damaged the translation by selecting a translation option (\textit{``zu platzieren''} $\rightarrow$ \textit{``zum Platzieren''}) learned from a lower ranked training sample (MT-Top4 and PE-Top4). This sample had probably received a higher weight from  the local models.

\section{Summary}
In this chapter, we discussed an online learning scenario, in which streams of data to be processed in real-time may feature high diversity regarding domain, PE style, and MT systems that generate the translations.
We investigated, for the first time, the challenges posed to APE technology by a multi-domain translation environment.
Our study showed that existing online and batch  solutions are not robust enough for this scenario because of their inability to discern which of the learned rules are suitable for a specific context. For example, a correction rule learned from one domain may not be valid for other domains.
We addressed this problem incrementally, first by proposing an instance selection technique that learned rules from contexts that were similar
to the MT segment to be post-edited.
The gains achieved by this solution over  the existing batch phrase-based APE methods were further increased by adding of a dynamic knowledge base that stored more reliable statistics about the learned translation options; this addition also improved the computation time. The adoption of this dynamic knowledge base allowed us to further extend our online approach by including features that capture negative human feedback, giving the system the capacity to  learn from mistakes made in the past.
Our evaluation results generally indicated that our approach improved the state-of- the-art performance for an online learning scenario with an English-German  MDTE dataset.

\chapter{Conclusion}
\label{cha:conclusion}
In this thesis, we addressed the problem of automatic post-editing machine-translated texts to correct recurring errors generated by an MT system.
We explored the potential of APE under several paradigms (starting from the classical phrase-based approach and moving then to 
end-to-end deep learning based solutions that now represent the state of the art) 
with various datasets (featuring different domains, MT systems, and post-edits' origin) under different learning modes (batch and online).
We investigated several problems and challenges posed by the data, technology, and application, and provided solutions to address them. 

\section{Data}
Before the work summarized in this thesis, APE datasets were very limited in terms of quantity, and were available for few languages.
As outcome of initial work, we released a domain-specific dataset - a subset of Autodesk post-editing corpus - that 
consists of the same source sentences machine-translated in several languages and post-edited by professional translators.
As discussed in Chapter \ref{chap:phrase-basedAPE}, this dataset helped us to provide a strong evidence to confirm that APE is quite effective 
when dealing with domain domain-specific scenarios,
 with gains ranging from 3.0 to 5.0 TER points across different language pairs.
We also released several datasets 
for the APE shared task, which helped us to monitor  state-of-the-art 
advancements.
On a domain-specific benchmark released at the 2016 APE shared task, our hybrid APE solution - discussed in Chapter \ref{chap:hybridAPE} - improved the translation quality of a PBSMT system by $\sim$3.0 BLEU points which further increased to $\sim$9.0 BLEU points by applying our neural-based solution - discussed in Chapter \ref{chap:neuralAPE}. 
Nowadays, also thanks to the work discussed in this thesis, the situation has changed: the data we contributed to create represent a solid starting point for future research on the topic.


\section{Technology}
Before the 
research summarized in this thesis, the few prior works in APE were based on the classical phrase-based MT technology
Moreover, in most of the cases, systems were designed to learn correction rules from (\textit{mt, pe}) pairs, disregarding the source context.
Our study in Chapter \ref{chap:phrase-basedAPE} confirmed through  experiments in multiple language directions the findings of \cite{bechara2011statistical} about the importance of modelling source information in order to better learn PE rules with a phrase-based APE system.
In Chapter \ref{chap:hybridAPE}, we extended the notion of using the source context in a more elegant factored machine translation approach (still phrase-based).
This approach allowed us to leverage both ``monolingual'' and ``context-aware'' models at run-time to mutually benefit from each other.
Our findings about the usefulness of source information in the phrase-based paradigm also hold true in the case of the neural paradigm.
In Chapter \ref{chap:neuralAPE}, our proposed multi-source neural APE approach 
leveraging both source and MT achieved the best result at the 2017 APE shared task.
This approach is still used in the current state-of-the-art APE systems.

One of the typical problems that exist in APE - irrespective of the underlying technology - is  over-correction, where the system makes unnecessary changes 
that can deteriorate the overall translation quality.
We addressed this problem in  several chapters, either by designing novel system architectures or by leveraging external resources.
Within the phrase-based paradigm, as discussed in Chapter \ref{chap:phrase-basedAPE}, we designed a similarity feature that penalizes translation containing more edits.
Within the neural paradigm, as discussed in Chapter \ref{chap:neuralAPE}, we introduced a novel task-specific loss function that rewards translation to preserves MT words.
We also addressed the problem of over-correction by leveraging external knowledge in the neural paradigm.
In this case, word-level quality judgments were used to inform the APE system whether an MT word is correct and should be preserved in the final translation, as discussed in Chapter \ref{chap:QEandAPE}.

The limited amount of data seems to have an active role in increasing sparseness, as a result, APE systems tend to learn noisy and unreliable PE rules.
We addressed this issue in Chapter \ref{chap:phrase-basedAPE} by introducing a pruning strategy to filter out unreliable rules from the phrase table of an APE system.
The strategy was more effective for the ``context-aware'' variant than the ``monolingual'' variant, primarily because the former approach learned PE rules from a joint representation that was prone to more data sparseness.
In the neural paradigm, we no longer rely on the joint representation to build APE systems.
Rather, we designed a multi-source neural architecture to process source and MT separately through different encoders which are combined by the decoder at run-time - as discussed in Chapter \ref{chap:neuralAPE}.
The same challenge of limited and sparse data was also addressed for online learning scenario in Chapter \ref{chap:onlineAPE}.
In this case, the system had to cope with a continuous stream of diverse data from different domains or genres.
Our proposed solution which was based on an instance selection mechanism was able to outperform other online learning system in terms of both translation quality and speed.
\section{Applications}
Before the work summarized in this thesis, APE has been only targeted as a mechanism to correct errors in a machine translated text.
Our work showed that it can also be used in an application scenario where a translation must comply with specific terminology - that is translation with constraints.
For example, in the case of MT, the constraints can take the form of a bilingual dictionary containing pre-defined target terms that should be generated in the translation.
Whereas, in the case of APE, the constraints were introduced in terms of words in the MT output that should be preserved or corrected  according to automatic word-level quality predictions.

In the first part of Chapter \ref{chap:QEandAPE}, we proposed a novel 
``guided decoding'' mechanism to leverage external knowledge in neural MT/APE decoding.
We discussed the challenges faced in the neural paradigm and showed that a naive integration of external knowledge based on attention information was insufficient.
A more sophisticated look-ahead mechanism was needed to guide the decoding process.
Specifically, we focused on the following problems: \textit{i)} how to enforce a given suggestion in the decoder's output; \textit{ii)} how to place this suggestion in the right position; and \textit{iii)} how to deal with out-of-vocabulary words.
Our proposed decoding mechanism did not require 
modifications to the training procedures nor 
time-consuming fine-tuning steps, and it was compatible with both word-level and sub-word-level 
input representations.

In the second part of Chapter \ref{chap:QEandAPE}, we investigated different strategies to combine APE and 
automatic quality estimation (QE), ranging from a light integration, in which  QE was used either to trigger APE or to compare the APE with the original MT segment, to tighter integration, in which QE annotations were directly used to guide the inner workings of the APE decoder. 
Our experiments confirmed that QE can provide useful signal to APE to produce better MT outputs.
Among the proposed strategies, the use of word-level QE on both MT and neural APE resulted in the largest gains over the top WMT 2016 APE system.
Nowadays, as a result of this research endeavour, APE can be seen not only as an isolated task, but also as a technology open to useful applications, with close relations with other MT tasks.


\section{Open Problems}
In this thesis we have shown that APE 
is useful to improve translation quality both in batch and in online mode.
However, our investigation considered MT systems based on the phrase-based paradigm that was more popular during our work.
Given the rapid advancement in MT technology, current neural-based solutions can generate much higher quality translations.
In this newer paradigm, checking whether APE can contribute to further improve the higher  quality of neural output remains an open question.
The major challenge to this verification is represented by data availability.
At the moment of writing this thesis, training corpora including post-edits of NMT output are in fact still limited.
Although our shared task on APE aims to address this aspect by releasing new datasets,\footnote{Post-edits of NMT output belonging to IT domain for EN-DE language pair was released in the 2018 APE shared task at WMT.} more efforts are needed to create corpora for different languages and domains to perform sound research. 

Our findings from several works confirm that APE is effective for domain-specific datasets (such as the Autodesk post-editing corpus used in Chapter \ref{chap:phrase-basedAPE} and the WMT 2016/2017 
shared task data used in Chapters \ref{chap:hybridAPE} and \ref{chap:neuralAPE} respectively).
Generic datasets covering different domains increase the level of difficulty.
Our work discussed  in Chapter \ref{chap:phrase-basedAPE}, which
focused on evaluating the effectiveness of APE for generic data, did not achieve positive results.
The main challenge we faced was to deal with data sparsity.
Due to limited and sparse data, the system learned noisy and unreliable PE rules and, as a consequence, application of these rules 
led to overall 
degradation of translation quality.
This problem of dealing with sparse data, especially in the case of generic data, remains open for further investigation.

In several chapters, we have addressed the task-specific problem of over-correction which has received less attention in the literature.
Although, our contributions in terms of introducing a task-specific loss function or leveraging external resource brought limited improvement, positive outcomes encourage 
further investigation.
Specifically, our work on using automatic quality judgments 
to constraint APE decoding resulted in significant improvements using 
an ``oracle'' QE system.
It would be interesting to check how close we can get to the oracle result by using the latest state-of-the-art QE 
technology.


\clearemptydoublepage


\thispagestyle{empty}
\makeatletter
\addcontentsline{toc}{chapter}{Bibliography}
\bibliographystyle{plain}
\bibliography{PhD-Thesis}

\clearemptydoublepage



\end{document}